\documentclass[aos,preprint,reqno]{imsart}
\setattribute{journal}{name}{}

\usepackage[utf8]{inputenc} 
\usepackage[T1]{fontenc}    
\usepackage[margin=1.1in]{geometry}
\usepackage[linkbordercolor={1 0 0}, colorlinks=true, urlcolor=blue, citecolor=blue]{hyperref}
\usepackage{url}            
\usepackage{booktabs}       
\usepackage{amsfonts}       
\usepackage{nicefrac}       
\usepackage{microtype}      
\usepackage{mdframed,calc}
\usepackage{algorithm}
\usepackage{algorithmic}
\usepackage[round,sort&compress]{natbib}
\usepackage{amsmath, amssymb, amsthm, bbm, dsfont, mathrsfs}
\usepackage{xr}
\usepackage{subfiles, subcaption}
\usepackage{color, verbatim, graphicx}
\usepackage{enumitem}
\usepackage{mathtools}
\usepackage{stmaryrd}

\usepackage{cleveref}
\theoremstyle{definition}

\renewcommand{\epsilon}{\varepsilon}

\newcommand{\given}{\,|\,}

\newcommand{\norm}[1]{\left\lVert#1\right\rVert}
\newcommand{\trace}{\text{tr}}

\newcommand{\D}{\mathscr{D}}

\newcommand{\N}{\mathcal{N}}
\renewcommand{\O}{\mathcal{O}}
\newcommand{\Q}{\mathcal{Q}}

\renewcommand{\SS}{\mathbb{S}}

\newcommand{\ZZ}{\reals^D}

\newcommand{\E}{\mathbb{E}}

\newcommand{\Cov}{\text{Cov}}
\newcommand{\KL}{\text{KL}}

\newcommand{\reals}{\mathbb{R}}

\newtheorem*{theorem*}{Theorem}
\newtheorem*{proposition*}{Proposition}
\newtheorem*{lemma*}{Lemma}
\newtheorem*{corollary*}{Corollary}

\usepackage{thmtools}

\definecolor{shade}{rgb}{0.9,0.9,0.9}

\declaretheoremstyle[
spaceabove=6pt, spacebelow=6pt,
postheadspace=1em,
headfont=\normalfont\bfseries,
notefont=\mdseries, notebraces={(}{)},
bodyfont=\normalfont,
]{mystyle}

\declaretheoremstyle[
    spaceabove=-6pt,  spacebelow=6pt,
    headfont=\normalfont\bfseries,
    bodyfont = \normalfont,
    postheadspace=1em,
    qed=$\blacksquare$,
    headpunct={:}]{myproofstyle} 

\declaretheorem[
  style=mystyle,
  name=Theorem,
  numberwithin=section,
  sharenumber=theorem
]{rtheorem}

\declaretheorem[
  shaded={
      rulecolor=black,
      textwidth=0.98\linewidth,
      rulewidth=1pt,
      bgcolor=gray!15,
      margin=5pt,
  },
  style=mystyle,
  name=Theorem,
  numberwithin=section,
  sharenumber=theorem
]{ftheorem}

\declaretheorem[
  shaded={
      rulecolor=black,
      textwidth=0.98\linewidth,
      rulewidth=1pt,
      bgcolor=gray!15,
      margin=5pt,
  },
  style=mystyle,
  name=Lemma,
  sharenumber=theorem
]{flemma}

\declaretheorem[
  shaded={
      rulecolor=black,
      textwidth=0.98\linewidth,
      rulewidth=1pt,
      bgcolor=gray!15,
      margin=5pt,
  },
  style=mystyle,
  name=Definition,
  sharenumber=theorem
]{fdefinition}

\declaretheorem[
  shaded={
      rulecolor=black,
      textwidth=0.98\linewidth,
      rulewidth=1pt,
      bgcolor=gray!10,
      margin=5pt,
  },
  style=mystyle,
  name=Corollary,
  sharenumber=theorem
]{fcorollary}

\declaretheorem[
  shaded={
      rulecolor=black,
      textwidth=0.98\linewidth,
      rulewidth=1pt,
      bgcolor=gray!10,
      margin=5pt,
  },
  style=mystyle,
  name=Proposition,
  sharenumber=theorem
]{fproposition}

\DeclareUnicodeCharacter{03BC}{$\mu$}
\DeclareUnicodeCharacter{03A3}{$\Sigma$}
\DeclareUnicodeCharacter{0393}{$\Gamma$}
\DeclareUnicodeCharacter{03BB}{$\lambda$}

\newcommand{\Sigmap}{\Sigma_*}
\newcommand{\mup}{\mu_*}

\usepackage{eqparbox}

\renewcommand{\vec}[1]{#1} 
\newcommand{\mat}[1]{#1} 

\usepackage[dvipsnames]{xcolor}
\usepackage[most]{tcolorbox}

\counterwithin{figure}{section}

\begin{document}

\begin{frontmatter}

\title{Batch and match:\ black-box variational inference \\ with a score-based
 divergence}
\runtitle{Batch and match black-box variational inference}

\begin{aug}
\author{\fnms{Diana} \snm{Cai}\thanksref{ccm}}
\author{\fnms{Chirag} \snm{Modi}\thanksref{ccm}} 
\\
\author{\fnms{Loucas} \snm{Pillaud-Vivien}\thanksref{ccm}}
\author{\fnms{Charles C.} \snm{Margossian}\thanksref{ccm}}
\\
\author{\fnms{Robert M.} \snm{Gower}\thanksref{ccm}}
\author{\fnms{David M.} \snm{Blei}\thanksref{columbia}}
\author{\fnms{Lawrence K.} \snm{Saul}\thanksref{ccm}}

\affiliation{\thanksmark{ccm}Flatiron Institute, \thanksmark{columbia}Columbia University}

\address[ccm]
{
    Center for Computational Mathematics \\
    Flatiron Institute, New York, NY \\
    Emails:
    \href{mailto:dcai@flatironinstitute.org}{dcai@flatironinstitute.org}, \\
    \href{mailto:cmodi@flatironinstitute.org}{cmodi@flatironinstitute.org}, \\
    \href{mailto:lpillaudvivien@flatironinstitute.org}{lpillaudvivien@flatironinstitute.org}, \\
    \href{mailto:cmargossian@flatironinstitute.org}{cmargossian@flatironinstitute.org}, \\
    \href{mailto:rgower@flatironinstitute.org}{rgower@flatironinstitute.org}, \\
    \href{mailto:lsaul@flatironinstitute.org}{lsaul@flatironinstitute.org}
}
\address[columbia]
{
    Department of Statistics \\ Department of Computer Science \\
    Columbia University, New York, NY \\
    Emails:
    \href{mailto:david.blei@columbia.edu}{david.blei@columbia.edu}
}

\runauthor{Cai, Modi, Pillaud-Vivien, Margossian, Gower, Blei, and Saul}
\end{aug}

\begin{keyword}
\kwd{approximate inference}
\kwd{black-box variational inference}
\kwd{score matching}
\kwd{stochastic proximal point algorithm}
\kwd{score-based divergence}
\kwd{quadratic matrix equations}
\end{keyword}

\begin{abstract}
Most leading implementations of black-box variational inference (BBVI) are based on optimizing a stochastic evidence lower bound (ELBO). But such approaches to BBVI often converge slowly due to the high variance of their gradient estimates and their sensitivity to hyperparameters.  In this work, we propose \emph{batch and match} (BaM), an alternative approach to BBVI based on a score-based divergence.  Notably, this score-based divergence can be optimized by a closed-form proximal update for Gaussian variational families with full covariance matrices. We analyze the convergence of BaM when the target distribution is Gaussian, and we prove that in the limit of infinite batch size the variational parameter updates converge exponentially quickly to the target mean and covariance.  We also evaluate the performance of BaM on Gaussian and non-Gaussian target distributions that arise from posterior inference in hierarchical and deep generative models.  In these experiments, we find that BaM typically converges in fewer (and sometimes significantly fewer) gradient evaluations than leading implementations of BBVI based on ELBO maximization.

\end{abstract}

\end{frontmatter}

\maketitle



\section{Introduction}

Probabilistic modeling plays a fundamental role in many problems of inference and decision-making,
but it can be challenging to develop accurate probabilistic models that remain computationally
tractable.
In {typical} applications, the goal is
to estimate a target distribution that cannot
be evaluated or sampled from exactly, but where an unnormalized form
is  available. A canonical situation is applied Bayesian
statistics, where the target is a posterior distribution of latent
variables given observations, but where only the model's joint
distribution is available in closed form.
Variational inference (VI) has emerged as a
leading method for fast
approximate inference~\citep{jordan1999vi,wainwright2008graphical,blei2017vi}.
The idea behind VI is to posit a parameterized
family of approximating distributions,
and then to find the member of that family
which is closest to the target distribution.

Recently, VI methods have become increasingly
``black box,'' in that they only require calculation of the log of the
unnormalized target and (for some algorithms) its
gradients~\citep{ranganath2014black,kingma2013auto,archer2015black,ryder2018black,locatello2018boosting,burroni2023sample,kim2023convergence,domke2019provable,welandawe2022robust,domke2023provable,Modi2023,giordano2023black}.
Further applications have built on
advances in automatic differentiation,
{and now} black-box variational
inference (BBVI) is widely
deployed in robust software packages for
probabilistic programming
~\citep{salvatier2016probabilistic,kucukelbir2017automatic,bingham2019pyro}.

In general, the ingredients of a BBVI strategy are the form of the
approximating family, the divergence to be minimized, and the
optimization algorithm to minimize it. Most BBVI algorithms work with
a factorized (or mean-field) family, and minimize
the reverse Kullback-Leibler (KL) divergence
via stochastic gradient descent (SGD). But this
approach has its drawbacks. The optimizations can be plagued by
high-variance gradients and sensitivity to hyperparameters of the learning algorithms
\citep{dhaka2020robust,dhaka2021challenges}.
These issues are further exacerbated in high-dimensional problems
and when using richer variational families
{that model the correlations between different latent variables.}
{There has been recent work on BBVI which avoids SGD for Gaussian variational families~\citep{Modi2023},}
but this approach does not minimize an explicit divergence and requires
additional heuristics to converge for non-Gaussian targets.

In this paper, we develop a new approach to BBVI. It is based on a
different divergence, accommodates expressive variational families,
and does not rely on SGD for optimization. In particular, we
introduce a novel \textit{score-based divergence} that measures the
agreement of the scores, or gradients of the log densities, of
the target and variational distributions.
This divergence can be estimated for unnormalized target distributions,
thus making it a natural choice for BBVI. We study the score-based
divergence for Gaussian variational families with full covariance,
rather than the factorized family.
{We also develop an efficient stochastic proximal point algorithm, with closed-form updates,
to optimize this divergence.}

Our algorithm is called \textit{batch and match} (BaM), and it
{alternates} between two types of steps. In the ``batch'' step, we draw a batch of samples
from the current approximation to the target
and use those samples to estimate the divergence;
in the ``match'' step, we estimate a new variational approximation by matching
the scores at these samples.
{By iterating these steps, BaM} finds a variational distribution
that is close in score-based divergence to the target.

Theoretically, we analyze the convergence of BaM when the target
itself is Gaussian. In the limit of an infinite batch size, we prove
that the variational parameters converge exponentially quickly to the target
mean and covariance {at a rate controlled by the quality of initialization
and the amount of regularization. Notably, this convergence result holds for any amount of
regularization; this stability to the ``learning rate'' parameter is characteristic of proximal algorithms,
which are often less brittle than SGD \citep{Asi2019}.}

Empirically, we evaluate BaM on a variety of
Gaussian and non-Gaussian target distributions, including a test suite of Bayesian
hierarchical models and deep generative models.
{On these same problems, we also compare BaM to a leading implementation of BBVI
based on ELBO maximization~\citep{kucukelbir2017automatic}
and a recently proposed algorithm for Gaussian score matching~\citep{Modi2023}.}
By and large, {we find that} BaM
converges faster and to more accurate solutions.

In what follows, we begin by {reviewing}
BBVI and {then developing a}
score-based divergence {for BBVI with several} important properties
(\Cref{sec-divergence}). Next, we propose
BaM, an iterative algorithm
for score-based Gaussian variational inference, and we study its rate of convergence
(\Cref{sec-algorithm}). We then present a discussion of related
methods in the literature (\Cref{sec-related}). Finally, we conclude
with a series of empirical studies on a variety of synthetic and real-data target
distributions (\Cref{sec-experiments}).
A Python implementation of BaM is available at
\href{https://github.com/modichirag/GSM-VI}{github.com/modichirag/GSM-VI/}.




\section{BBVI with the score-based divergence}
\label{sec-divergence}

VI was developed as a way to estimate an unknown \emph{target} distribution
with density $p$; here we assume that the target is a distribution on~$\ZZ$.
The target is estimated by first positing a \emph{variational family} of distributions
$\Q$, then finding the particular $q \in \Q$ that minimizes an objective
$\mathscr{L}(q)$ measuring the difference between $p$ and $q$.

\subsection{From VI to BBVI to score-based BBVI}

In the classical formulation of VI, the objective~$\mathscr{L}(q)$ is
the (reverse) Kullback-Leibler (KL) divergence:
\begin{align}
\KL(q; p) := \int \log\left(\tfrac{q(z)}{ p(z)}\right) \, q(z) \, dz.
\end{align}
For some models the derivatives of $\KL(q;p)$ can be exactly evaluated, but for many others they cannot. In this case a further approximation is needed. This more challenging situation is the typical setting for BBVI.

In BBVI, it is assumed that (a) the target density~$p$ cannot be evaluated pointwise
or sampled from exactly, but that (b) an
unnormalized target density is available.
BBVI algorithms use stochastic gradient descent to minimize
the KL divergence, or equivalently, to maximize the evidence lower bound (ELBO). The necessary gradients in this case can be estimated with access to the unnormalized target density.
But in practice this objective is difficult to optimize:
the optimization can converge slowly due to noisy gradients,
and it can be sensitive to the choice of learning rates.

In this work, we will also assume additionally that
(c) the log target density is differentiable, and its derivatives can be efficiently evaluated.
We define the target density's
\emph{score function} $s: \ZZ \rightarrow \reals^D$ as
\[s(z) := \nabla_z \log p(z).\]
It is often possible to compute these scores even when $p$ is intractable because they only depend on the logarithm of the unormalized target density.
In what follows, we introduce the score-based divergence and study its properties;
in \Cref{sec-algorithm},
we will then propose a BBVI algorithm based on this score-based divergence.

\paragraph{Notation}

For $\Sigma \in \reals^{D \times D}$, let
$\Sigma\! \succ\! 0$
denote that $\Sigma$ is positive definite and
$\Sigma\! \succeq\! 0$  denote that $\Sigma$ is positive semi-definite.
Define the set of symmetric, positive definite matrices as
\mbox{$\SS_{++}^D := \{\Sigma \in \reals^{D \times D}: \Sigma=\Sigma^\top, \Sigma \succ 0 \}$}.
Let $\trace(\Sigma) := \sum_{d=1}^D \Sigma_{dd}$ denote the trace of $\Sigma$
and
let $I \in \reals^{D \times D}$ denote the identity matrix.
We primarily consider two norms throughout the paper: 
first, given $z \in \reals^D$ and $\Sigma\in\reals^{D\times D}$,
we define the $\Sigma$-weighted vector norm,
$\norm{z}_\Sigma := \sqrt{z^\top \Sigma z}$, and second,
given $\Sigma\in\reals^{D\times D}$, we define the matrix norm
$\norm{\Sigma}$ to be the spectral norm.

\subsection{The score-based divergence}

We now introduce the score-based divergence, which will be the basis for a
BBVI objective.
Here we focus on a Gaussian variational family,
i.e.,
\[\Q = \{\N(\mu, \Sigma): \mu \in \reals^D, \Sigma \in \SS^{D}_{++}\},\]
but we generalize the score-based divergence
to non-Gaussian distributions
in \Cref{app:divergence}.

The \emph{score-based divergence}
between densities $q \in \Q$ and $p$~on~$\ZZ$
is defined as
\begin{align}
\label{eq-score-divergence}
    \D(q; p) := \int
\norm{\nabla_z \log\left(\tfrac{q(z)}{p(z)}\right)}^2_{\Cov(q)}
\, q(z) \, dz,
\end{align}
where $\Cov(q) \in \mathbb{S}^D_{++}$ is the covariance matrix of the variational
density $q$.

Importantly, the score-based divergence
can be evaluated when $p$
is only known up to a normalization constant, as it only depends on the target
density through the score $\nabla \log p$.
Thus, not only can this divergence be used as a VI objective,
but
it can also be used for goodness-of-fit evaluations, unlike the KL divergence.

{The divergence in eq.~(\ref{eq-score-divergence}) is
well-defined under mild conditions on $p$ and $q$ (see~\Cref{app:divergence}), and it}
enjoys two {important} properties:

\begin{enumerate}[label=\textbf{Property \arabic*},itemindent=*,topsep=4pt,itemsep=2pt]
\label{prop-1}
\item (Non-negativity \& equality):
    $\D(q; p) \geq 0$ with $\D(q;p) = 0$ iff $p=q$.

\item (Affine invariance):
    Let $h\!:\!\ZZ\!\rightarrow\!\ZZ$
        be an affine transformation,
        and consider the induced densities
        $\tilde q(h(z))\! =\! q(z) |\mathcal{J}(z)|^{-1}$
        and
        $\tilde p(h(z))\! =\! p(z) |\mathcal{J}(z)|^{-1}$,
where $\mathcal{J}$ is the determinant of the Jacobian of $h$.
        Then
        $\D(q; p) = \D(\tilde q; \tilde p)$.
\end{enumerate}

We note that these properties are also satisfied by the KL divergence \citep{qiao2010study}.
The first property shows that $\D(q;p)$ is a proper divergence measuring
the agreement between $p$ and $q$.
The second property states that the score-based divergence
$\D(q,p)$ is invariant under affine transformations; this property is desirable to maintain a consistent measure of similarity
under coordinate transformations of the input.
{This property depends crucially on
the weighted vector norm, mediated by $\Cov(q)$,
in the divergence of eq.~(\ref{eq-score-divergence}).}

There are several related divergences in the research literature.
A generalization of the score-based divergence
is the weighted Fisher divergence \citep{barp2019minimum},
given by
$\E_q[\norm{\nabla\log(q/p)}_M^2]$,
where $M\!\in\!\reals^{D\times D}$;
the score-based divergence is recovered by the choice $M\! =\! \Cov(q)$.
A special case of the score-based divergence is the
Fisher divergence \citep{hyvarinen2005estimation}
given by $\E_q[\norm{\nabla\log(q/p)}_I^2]$,
but this divergence is not affine invariant.  (See the proof of \Cref{theorem-affine} for further
        discussion.)

\section{Score-based Gaussian variational inference}
\label{sec-algorithm}

The score-based divergence has many favorable properties for VI.
We now show that this divergence can also be efficiently optimized by an iterative black-box algorithm.

\subsection{Algorithm}

Our goal is to find some Gaussian distribution $q^*\! \in\! \Q$ that minimizes $\D(q;p)$.
Without additional assumptions on the target $p$,
the score-based divergence ${\D}(q;p)$ is not analytically tractable.
So instead we consider a Monte Carlo estimate of $\D(q;p)$:
given samples $z_1,\ldots,z_B \sim q$, we construct the approximation
\begin{align}
    \D(q; p) \approx
    \frac{1}{B}
    \sum_{b=1}^B
    \norm{\nabla_z \log \left(\tfrac{q(z_b)}{p(z_b)}\right)}_{\Cov(q)}^2.
    \label{eq:DqpApprox}
\end{align}
This estimator is unbiased, but it does not lend itself to optimization:\
we cannot simultaneously sample from $q$ while also optimizing over the family $\Q$ to which it belongs.
There is a generic solution to the above problem:
    the so-called ``reparameterization trick'' (e.g.,
    \citet{kucukelbir2017automatic}) decouples the sampling distribution and optimization variable.
    But this approach leads to a gradient-based algorithm that does not fully capitalize on
    the structure of the Gaussian variational family.

In this paper we take a different approach, one that does capitalize on this structure.
Specifically,
we take an iterative approach whose goal is to
produce a sequence of distributions $\{q_t\}_{t=0}^\infty$ that converges to $q^*$.
At a high level, the approach alternates between two steps---one that constructs a
biased estimate of $\D(q;p)$, and another that updates $q$ based on this biased estimate, but not too aggressively (so as to minimize the effect of the bias). Specifically, at the $t^\textrm{th}$ iteration,
we first estimate $\D(q;p)$ with samples from $q_t$: i.e., given $z_1,\ldots, z_B \sim q_t$, we compute
\begin{equation}
\widehat{\D}_{q_t}(q; p) :=
    \frac{1}{B}
    \sum_{b=1}^B
    \norm{\nabla_z \log \left(\tfrac{q(z_b)}{p(z_b)}\right)}_{\Cov(q)}^2.
  \label{eq:Dhatqt}
\end{equation}
We call eq.~(\ref{eq:Dhatqt}) the \textit{batch} step because it estimates $\D(q,p)$ from the batch of samples $z_1,\ldots,z_B\sim q_t$.

The batch step of the algorithm relies on stochastic sampling, but it alternates with a deterministic step that updates $q$ by minimizing the empirical score-based divergence $\widehat{\D}_{q_t}(q; p)$ in eq.~(\ref{eq:Dhatqt}).
Importantly, this minimization is subject to a {regularizer}:
{we penalize large differences between $q_t$ and $q_{t+1}$ by their KL divergence.}
Intuitively, when $q$ remains close to $q_t$, then $\widehat{\D}_{q_t}(q;p)$
{in eq.~(\ref{eq:Dhatqt})}
remains a good approximation to the unbiased estimate
$\widehat\D_q(q;p)$ {in eq.~(\ref{eq:DqpApprox})}.
{With this in mind, we compute}
 $q_{t+1}$ by minimizing the \textit{regularized} objective function
\begin{equation}
\label{eq:LBaM1}
\mathscr{L}^\text{BaM}(q) := \widehat{\D}_{q_t}(q; p)
    + \tfrac{2}{\lambda_t} \, \KL(q_t; q),
\end{equation}
where $q\in\Q$ and $\lambda_t\!>\!0$ is the inverse regularization parameter. When $\lambda_t$ is
small, the regularizer is large, encouraging the next iterate $q_{t+1}$ to remain close to $q_t$; thus
$\lambda_t$ can also be viewed as a learning rate.

\textit{The objective function in eq.~(\ref{eq:LBaM1}) has the important property that its global minimum can be computed
analytically in closed form.} In particular, {we can optimize eq.~(\ref{eq:LBaM1}) without recourse}
to gradient-based methods 
{that are derived from} a linearization around~$q_t$.
We refer to the minimization of $\mathscr{L}^\text{BaM}(q)$ in  eq.~(\ref{eq:LBaM1}) as the  \textit{match} step because
the updated distribution $q_{t+1}$ always matches the
scores at $z_1,\ldots,z_B$ better than the current one~$q_t$.

Combining these two steps, we arrive at the \textit{batch and match} (BaM) algorithm for BBVI with a
score-based divergence. The intuition behind this iterative approach will be formally justified in
\Cref{sec:proof-converge} by a proof of convergence. We now discuss each step of the algorithm in
greater detail.

\textbf{Batch Step.} {This step begins by sampling $z_1,z_2,\ldots,z_B \sim q_t$ and computing} the scores $g_b
= \nabla\log p(z_b)$ {at each sample. It then calculates}
the means and covariances (over the batch) of these quantities; we
denote these statistics by
\begin{align}
    \overline{z}_{} &= \frac{1}{B} \sum_{b=1}^B z_b,
    \quad
    C_{} = \frac{1}{B} \sum_{b=1}^B (z_b-\overline{z}) (z_b-\overline{z})^\top \\
   \overline{g}_{} &= \frac{1}{B} \sum_{b=1}^B g_b, \quad
   \Gamma_{} = \frac{1}{B} \sum_{b=1}^B (g_b-\overline{g}) (g_b-\overline{g})^\top,
\end{align}
where $\overline{z},\overline{g}\in\mathbb{R}^D$ are the means, respectively, of the samples and the
scores, and $C,\Gamma\in\mathbb{R}^{D\times D}$ are their covariances. In \Cref{app:BaM}, we show
that the empirical score-based divergence $\widehat{\D}_{q_t}(q; p)$
in eq.~(\ref{eq:Dhatqt}) can be written in terms of these statistics as
\begin{equation*}
\widehat{\D}_{q_t}(q; p)= \trace(\Gamma\Sigma) +
         \trace(C\Sigma^{-1}) +
         \big\|\mu\! -\! \overline{z}\!-\!\Sigma\overline{g}\big\|^2_{\Sigma^{-1}} \!+ \mbox{\small
         const.},
\end{equation*}
where {for clarity} we have suppressed additive constants that do not depend on the mean $\mu$ or covariance
$\Sigma$ of $q$. This calculation completes the batch step of BaM.

\textbf{Match Step.} The match step of BaM updates the variational approximation $q$ by setting
\begin{equation}
q_{t+1} = \arg\min_{q\in\Q} \mathscr{L}^\text{BaM}(q),
\label{eq:updateQ}
\end{equation}
where $\mathscr{L}^\text{BaM}(q)$ is given by eq.~(\ref{eq:LBaM1}). This optimization can be solved
in closed form; that is, we can analytically calculate the variational mean $\mu_{t+1}$ and
covariance $\Sigma_{t+1}$ that minimize $\mathscr{L}^\text{BaM}(q)$.

{The details of this calculation are given in \Cref{app:BaM}. There we show}
that the updated covariance~$\Sigma_{t+1}$ satisfies a
quadratic matrix equation,
\begin{equation}
\Sigma_{t+1} U \Sigma_{t+1} + \Sigma_{t+1} = V,
\label{eq:BaM-quadratic}
\end{equation}
where the matrices
$U$ and $V$ in this expression are positive semidefinite and determined by statistics from the batch step of BaM. In particular, these matrices are given by
\begin{align}
   \label{eq:U1}
   U &=  \lambda_t \Gamma + \tfrac{\lambda_t}{1+\lambda_t} \, \overline{g}\,\overline{g}^\top \\
  \label{eq:V1}
  V &=  \Sigma_t + \lambda_t C + \tfrac{\lambda_t}{1+\lambda_t}
            (\mu_t-\overline{z})(\mu_t-\overline{z})^\top.
\end{align}
The quadratic matrix equation in eq.~(\ref{eq:BaM-quadratic}) has a symmetric and positive-definite solution (see \Cref{app:quadratic}), and it is given~by
\begin{equation}
\Sigma_{t+1} =
   2\, V \Big(I + (I + 4\, U V)^\frac{1}{2}\Big)^{-1}.
   \label{eq:update_Sigma1}
\end{equation}
The solution in eq.~(\ref{eq:update_Sigma1}) is the BaM update for the variational covariance.
The update for the variational mean is given~by
\begin{equation}
 \mu_{t+1} = \tfrac{1}{1+\lambda_t} \mu_t + \tfrac{\lambda_t}{1+\lambda_t}\left(\Sigma_{t+1} \, \overline{g} + \overline{z}\right).
 \label{eq:update_mu1}
\end{equation}
Note that the update for $\mu_{t+1}$ depends on $\Sigma_{t+1}$, so these updates must be performed in the order shown above. The updates in eq.~(\ref{eq:update_Sigma1}--\ref{eq:update_mu1}) complete the match step of BaM.


\begin{algorithm}[t]
\begin{algorithmic}[1]
    \small
    \STATE{\textbf{Input:} Iterations $T$, batch size $B$,
    inverse regularization \mbox{$\lambda_t\!>\!0$},
    target score function $s: \ZZ \rightarrow \reals^D$,
        initial variational mean $\mu_0 \in \reals^D$ and
    covariance $\Sigma_0 \in \SS_{++}^{D}$
        }
    \FOR{$t=0,\ldots,T\!-\!1$}
    \STATE Sample batch  $z_b \sim \mathcal{N}(\mu_t, \Sigma_t)$ for $b = 1,\ldots, B$
\STATE Evaluate scores $g_{b} = s(z_b)$ for $b = 1,\ldots, B$
        \STATE Compute statistics $\overline{z}, \overline{g} \in \reals^D$
        and $\Gamma, C \in \reals^{D \times D}$
        \vspace{-8pt}
            \begin{align*}
                \overline{z}_{} &= \tfrac{1}{B} \sum_{b=1}^B z_b,
                \qquad
                C_{} = \tfrac{1}{B} \sum_{b=1}^B (z_b-\overline{z}) (z_b-\overline{z})^\top
                \\
                \overline{g}_{} &= \tfrac{1}{B} \sum_{b=1}^B g_b, \qquad
                \Gamma_{} = \tfrac{1}{B} \sum_{b=1}^B (g_b-\overline{g}) (g_b-\overline{g})^\top
            \end{align*}
          \vspace{-8pt}
         \STATE Compute matrices $U$ and $V$ needed to solve the quadratic matrix equation $\Sigma U \Sigma + \Sigma = V$
         \vspace{-5pt}
        \begin{align*}
            U &=  \lambda_t \Gamma + \tfrac{\lambda_t}{1+\lambda_t} \, \overline{g}\, \overline{g}^\top \\
            V &=  \Sigma_t + \lambda_t C + \tfrac{\lambda_t}{1+\lambda_t}
            (\mu_t-\overline{z})(\mu_t-\overline{z})^\top
        \end{align*}
        \vspace{-12pt}
    \STATE Update variational parameters
       \vspace{-5pt}
        \begin{align*}
            {\Sigma}_{t+1} &=
            2\, V \Big(I + (I + 4\, U V)^\frac{1}{2}\Big)^{-1}
            \\[-1ex] \nonumber
            {\mu}_{t+1} &=
            \tfrac{1}{1+\lambda_t} \mu_t + \tfrac{\lambda_t}{1+\lambda_t}\left(\Sigma_{t+1} \, \overline{g} +
            \overline{z}\right)
        \end{align*}
        \vspace{-12pt}
    \ENDFOR
    \STATE \textbf{Output:} variational parameters $\mu_T, \Sigma_T$
    \end{algorithmic}
    \caption{Batch and match VI}
    \label{alg:LS_GSM_VI}
\end{algorithm}

More intuition for BaM can be obtained by examining certain limiting cases of the batch size and
learning rate. When $\lambda_t\!\rightarrow\! 0$, the updates have no effect, with
$\Sigma_{t+1}\!=\!\Sigma_t$ and $\mu_{t+1}\!=\!\mu_t$. Alternatively, when $B\!=\!1$ and
$\lambda_t\!\rightarrow\!\infty$, the BaM updates reduce to the recently proposed updates for BBVI
by (exact) Gaussian score matching~\citep{Modi2023}; this equivalence is shown in \Cref{app:BaM}.
Finally, when $B\rightarrow\infty$ and $\lambda_0\!\rightarrow\!\infty$ (in that order), BaM converges
to a Gaussian target distribution in one step; see \Cref{cor:one-step} of
\Cref{app:proof-converge}.

We provide pseudocode for BaM in \Cref{alg:LS_GSM_VI}. We note that it costs
$\O(D^3)$ to compute the covariance update as shown in eq.~(\ref{eq:update_Sigma1}), but for small batch sizes, when the
matrix~$U$ is of rank $\O(B)$ with $B\!\ll\!D$, it is possible to compute the
update in {$\O(D^2B + B^3)$}; this update is presented in \Cref{lemma-quadratic-psd-invertible-low} of~\Cref{app:quadratic}.

BaM incorporates many ideas from previous work.
Like the stochastic proximal point (SPP) method~\citep{Asi2019,Davis2019},
it minimizes a Monte Carlo estimate of a divergence subject to a regularization term.
In proximal point methods, the updates are always regularized by squared Euclidean distance, but the
KL divergence has been used elsewhere as a regularizer---for example, in the EM
algorithm~\citep{tseng2004analysis,chretien2000kullback} and for approximate Bayesian
inference~\citep{theis2015trust,khan2015kullback,khan2015faster,dai2016provable}.
KL-based regularizers are also a hallmark of mirror descent methods~\citep{nemirovskii1983problem},
but in these methods the objective function is linearized---a poor approximation for
objective functions with high curvature. Notably, BaM does not introduce any linearizations because
its optimizations in eq.~(\ref{eq:updateQ}) can be solved in closed form.

\subsection{Proof of convergence for Gaussian targets}
\label{sec:proof-converge}

In this section we analyze a concrete setting in which we can rigorously prove the convergence of the updates in~\Cref{alg:LS_GSM_VI}.

Suppose the target distribution is itself a
Gaussian and the updates are computed in the limit of infinite batch size ($B\!\rightarrow\!\infty$). 
{In this setting we show that} BaM converges to the target distribution.
{More precisely,}
we show that the variational parameters converge exponentially quickly to their target values \textit{for all fixed levels of regularization $\lambda\!>\!0$ and no matter how they are initialized}.
{Our proof does not exclude the possibility of convergence in less restrictive settings, and in}
\Cref{sec-experiments}, we
observe empirically
that the updates also converge for non-Gaussian targets and finite batch sizes.
Though the proof here does not cover such cases, it remains instructive in many ways.

To proceed, consider a Gaussian target distribution
\mbox{$p = \N(\vec{\mup},\mat{\Sigmap})$}. At the~$t^{\rm th}$ iteration of Algorithm~\ref{alg:LS_GSM_VI},
we measure the normalized \textit{errors} in the mean and covariance parameters by
\begin{align}
\vec{\varepsilon}_t &:= \mat{\Sigmap}^{-\frac{1}{2}}(\vec{\mu}_t\!-\!\vec{\mup}), \label{eq:epsilon_t} \\
\mat{\Delta}_t &:= \mat{\Sigmap}^{-\frac{1}{2}}(\mat{\Sigma}_t\!-\!\mat{\Sigmap})\,\mat{\Sigmap}^{-\frac{1}{2}}. \label{eq:Delta_t}
\end{align}
The theorem below shows that $\varepsilon_t,\Delta_t\rightarrow 0$ in spectral norm.
Specifically, it shows that this convergence occurs exponentially quickly at a rate controlled by the quality of initialization and amount of regularization. \\

\begin{rtheorem}[\textbf{Exponential convergence}]
Suppose that $p = {\cal N}(\vec{\mup},\mat{\Sigmap})$ in \Cref{alg:LS_GSM_VI},
    and let $\alpha\!>\!0$ denote the minimum eigenvalue of the matrix
    $\mat{\Sigmap}^{-\frac{1}{2}}\mat{\Sigma}_0\mat{\Sigmap}^{-\frac{1}{2}}$.
    For any fixed level of regularization $\lambda\!>\!0$, define
\begin{equation}
\beta := \min\left(\alpha,\frac{1\!+\!\lambda}{1\!+\!\lambda\!+\!\|\varepsilon_0\|^2}\right), \quad
\delta := \frac{\lambda\beta}{1\!+\!\lambda},
\label{eq:beta_delta}
\end{equation}
where $\beta\in(0,1]$ measures the quality of initialization and
    $\delta\in(0,1)$ denotes a rate of decay. Then
with probability 1 in the limit of infinite batch size ($B\!\rightarrow\!\infty$), and for all $t\!\geq 0$, the normalized errors in eqs.~(\ref{eq:epsilon_t}--\ref{eq:Delta_t}) satisfy
\begin{align}
\|\vec{\varepsilon}_t\| &\leq (1\!-\!\delta)^{t} \|\vec{\varepsilon}_0\|, \label{eq:epsilon_converge} \\
\|\mat{\Delta}_t\| &\leq (1\!-\!\delta)^t \|\mat{\Delta}_0\|\,
  +\, t(1\!-\!\delta)^{t-1}  \|\vec{\varepsilon}_0\|^2.\hspace{-2ex} \label{eq:Delta_converge}
\end{align}
\label{thm:converge}
\end{rtheorem}

\vspace{-20pt}
Before sketching the proof we make three remarks. First, these error bounds behave sensibly: they
suggest that the updates converge more slowly when the learning rate is small (with $\lambda\!\ll\!
1$), when the variational mean is poorly initialized (with \mbox{$\|\vec{\varepsilon}_0\|^2\!\gg\!
1$}), and/or when the initial estimate of the covariance is nearly singular (with $\alpha\!\ll\!
1$). Second, the theorem holds under very general conditions---not only for any initialization of
$\mu_0$ and $\Sigma_0\!\succ\!0$, but also for any $\lambda\!>\!0$. This robustness is typical of
\textit{proximal} algorithms, which are well-known for their stability with respect to
hyperparameters~\citep{Asi2019}, but it is uncharacteristic of many \textit{gradient-based} methods,
which only converge when the learning rate varies inversely with the largest eigenvalue of an
underlying Hessian~\citep{handbookgrad}. Third, with more elaborate bookkeeping, we can derive \textit{tighter} bounds both for the above setting and also when different iterations use varying levels of regularization~$\{\lambda_t\}_{t=0}^{\infty}$. We give a full proof with these extensions in Appendix~\ref{app:proof-converge}. 

\begin{proof}[Proof Sketch]
The crux of the proof is to bound the normalized errors in eqs.~(\ref{eq:epsilon_t}--\ref{eq:Delta_t}) from one iteration to the next. Most importantly, we show that
\begin{align}
\|\vec{\varepsilon}_{t+1}\| &\leq (1\!-\!\delta) \|\vec{\varepsilon}_t\|, \label{eq:epsilon_contract} \\
\|\mat{\Delta}_{t+1}\| &\leq (1\!-\!\delta) \|\mat{\Delta}_t\|\,
  +\, \|\vec{\varepsilon}_t\|^2,\hspace{-2ex} \label{eq:Delta_contract}
\end{align}
where $\delta$ is given by eq.~(\ref{eq:beta_delta}), and from these bounds, we use induction to prove the overall rates of decay in \mbox{eqs.~(\ref{eq:epsilon_converge}-\ref{eq:Delta_converge})}.
Here we briefly describe the steps that are needed to derive the bounds in eqs.~(\ref{eq:epsilon_contract}--\ref{eq:Delta_contract}).

The first is to examine the statistics computed at each iteration of the algorithm in the \textit{infinite batch} limit \mbox{($B\!\rightarrow\!\infty$)}.
This limit is simplifying because
by the law of large numbers, we can replace the batched averages over $B$ samples at each iteration by their expected values under the variational distribution $q_t\!=\!{\cal N}(\vec{\mu}_t,\mat{\Sigma}_t)$. The second step of the proof is to analyze the algorithm's convergence in terms of the \textit{normalized} mean~$\vec{\varepsilon}_t$ in eq.~(\ref{eq:epsilon_t}) and the \textit{normalized} covariance~matrix
\begin{equation}
\mat{J}_t = \mat{\Sigmap}^{-\frac{1}{2}}\mat{\Sigma}_t\mat{\Sigmap}^{-\frac{1}{2}}
    = I + \Delta_t,
\end{equation}
where $\mat{I}$ denotes the identity matrix.
In the infinite batch limit, we show that with probability 1 these quantities satisfy 
\begin{align}
\lambda \mat{J}_{t+1}\left(\mat{J}_t\!+\!\tfrac{1}{1+\lambda}\vec{\varepsilon}_t\vec{\varepsilon}_t^\top\right)\mat{J}_{t+1}
 + \mat{J}_{t+1} = (1\!+\!\lambda)\mat{J}_t,  \label{eq:J_update} \\[-1ex]
\vec{\varepsilon}_{t+1} =
    \left(\mat{I}-\tfrac{\lambda}{1+\lambda}\mat{J}_{t+1}\right)\vec{\varepsilon}_t.
\label{eq:epsilon_update}
\end{align}
The third step of the proof is to \textit{sandwich} the matrix~$\mat{J}_{t+1}$ that appears in eq.~(\ref{eq:J_update}) between two other positive-definite matrices whose eigenvalues are more easily bounded. Specifically, at each iteration~$t$, we introduce matrices $\mat{H}_{t+1}$ and~$\mat{K}_{t+1}$ defined by
\begin{align}
\lambda \mat{H}_{t+1}\left(\mat{J}_t\!+\!\tfrac{\|\vec{\varepsilon}_t\|^2}{1+\lambda}\mat{I}\right)\mat{H}_{t+1}
 + \mat{H}_{t+1} &= (1\!+\!\lambda)\mat{J}_t,  \label{eq:H_update} \\[-0.5ex]
\lambda \mat{K}_{t+1}\,\mat{J}_t\,\mat{K}_{t+1}
 + \mat{K}_{t+1} &= (1\!+\!\lambda)\mat{J}_t.  \label{eq:K_update}
\end{align}
It is easier to analyze the solutions to these equations because they replace
the outer-product $\vec{\varepsilon}_t\vec{\varepsilon}_t^\top$
in eq.~(\ref{eq:J_update}) by a multiple of the identity matrix. We show that
for all times $t\geq 0$,
\begin{equation}
\mat{H}_{t+1} \preceq \mat{J}_{t+1} \preceq \mat{K}_{t+1},
\label{eq:sandwich}
\end{equation}
so that we can prove $\|\mat{J_t}\!-\!\mat{I}\|\!\rightarrow\!0$ by showing $\|\mat{H_t}\!-\!\mat{I}\|\!\rightarrow\!0$ and \mbox{$\|\mat{K_t}\!-\!\mat{I}\|\!\rightarrow\!0$}.  Finally, the last (and most technical) step is to derive the bounds in eqs.~(\ref{eq:epsilon_contract}--\ref{eq:Delta_contract}) by combining the sandwich inequality in eq.~(\ref{eq:sandwich}) with a detailed analysis of eqs.~(\ref{eq:J_update}--\ref{eq:K_update}).
\end{proof}



\section{Related work}
\label{sec-related}

{BaM builds on intuitions from earlier work on Gaussian score matching (GSM)~\citep{Modi2023}.
GSM is an iterative algorithm for BBVI that updates a full-covariance Gaussian}
by analytically solving a system of nonlinear equations.
As previously  discussed,
BaM recovers GSM as a special limiting case.
A limitation of GSM is that it aims to match the scores exactly; thus, if the target is not exactly Gaussian,
the updates for GSM attempt to solve an infeasible problem,
{In addition, the batch updates for GSM perform an ad hoc averaging that is not guaranteed to match any scores exactly, even when it is possible to do so.}
BaM {overcomes these limitations by}
optimizing  {a proper} score-based divergence on each batch of samples.
Empirically, with BaM, we observe that
larger batch sizes lead to more stable convergence.
{The score-based divergence behind BaM also lends itself to analysis, and we can provide}
theoretical guarantees on the convergence of BaM
for Gaussian targets.

Proximal point methods have been studied {in several papers} in the context of variational
inference; typically the objective is a stochastic estimate of the ELBO with a
(forward) KL regularization term.
{For example,} \citet{theis2015trust} optimize this objective using alternating coordinate ascent.
{In other work,} \citet{khan2015kullback,khan2015faster}
propose a splitting method {for this objective, and}
{by linearizing the difficult terms, they obtain}
a closed-form solution when the
variational family is Gaussian and additional knowledge is given about the structure of the target.
{By contrast, BaM} does not {resort to}
linearization in order  to obtain an analytical solution, nor {does it}
require additional assumptions on the structure of the target.

Proximal algorithms have also been developed for Gaussian variational families based on the
Wasserstein metric.
\citet{lambert2022variational} consider a KL objective with the Wasserstein
metric as a regularizer; in this case, the proximal step is not solvable in closed form.
On the other hand,
\citet{diao2023forward} consider a proximal-gradient method, and show that the
proximal step admits a closed-form solution.

Several works consider score matching with a Fisher divergence in the context of VI.
For instance,
\citet{yu2023semiimplicit} propose a score-matching approach
for semi-implicit variational families based on
stochastic gradient optimization of the Fisher divergence.
\citet{zhang2018variational} use the Fisher divergence
with an energy-based model as the variational family.
{BaM differs from these approaches by working with a Gaussian variational family and
an affine-invariant score-based divergence.}

Finally, we note that the idea of score matching
\citep{hyvarinen2005estimation} with a (weighted) Fisher divergence
appears in many contexts beyond VI~\citep{song2019generative,barp2019minimum}.
{One such context is generative modeling: here, given a set of training examples,} the goal is 
to {approximate}  an unknown data distribution~$p$
{by a parameterized} model $p_\theta$ with an intractable normalization constant.
{Note that in this setting one can evaluate $\nabla\log p_\theta$ but not $\nabla\log p$.
This setting is quite different from the setting of VI in this paper where}
we do \textit{not} have samples from $p$, where we \textit{can} evaluate $\nabla \log p$,
and {where the approximating distribution $q$ has the much simpler and
more tractable form of a multivariate Gaussian}.





\section{Experiments}
\label{sec-experiments}

\begin{figure*}[t]
    \centering
    \includegraphics[scale=0.49]{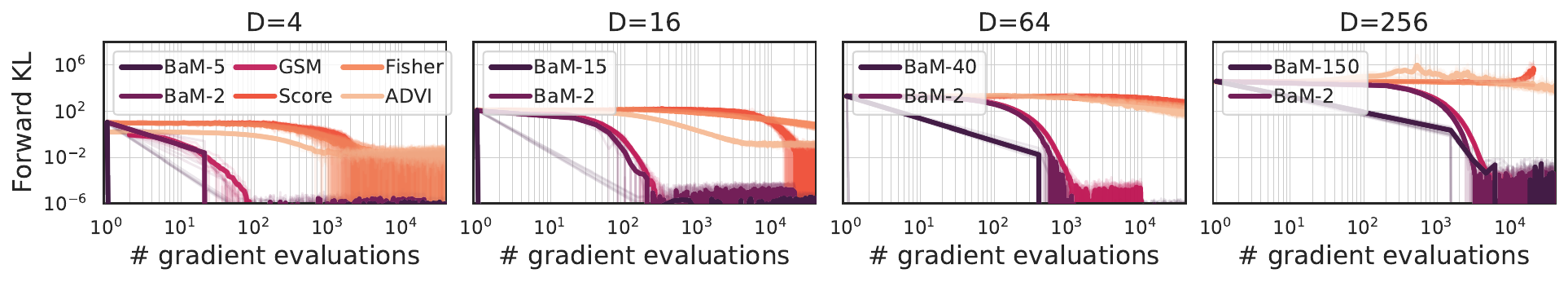}
    \caption{Gaussian targets of increasing dimension.
    Solid curves indicate the mean over 10 runs (transparent curves).
    {ADVI, Score, Fisher,} and GSM use a batch size of $B\!=\!2$.
    The batch size for BaM is given in the legend.
}
    \label{fig-gaussian-target}
\end{figure*}

We evaluate BaM against two other BBVI methods
{for Gaussian variational families with full covariance matrices}.
{The first of these is automatic differentiation VI (ADVI)~\citep{kucukelbir2017automatic}, which is based on ELBO maximization,
and the second is GSM~\citep{Modi2023}, as described in the previous section.}
We implement all algorithms using \texttt{JAX}
\citep{jax2018github},\footnote{Python implementations of BaM and the baselines are available
at: \href{https://github.com/modichirag/GSM-VI}{https://github.com/modichirag/GSM-VI/}.} which supports efficient automatic differentiation both on CPU and GPU.
We provide pseudocode for these methods in  \Cref{appendix-sec-baselines}.

\subsection{Synthetically-constructed target distributions}

We first validate BaM {in two settings}
where we know the true target distribution $p$.
{In the first setting, we} construct Gaussian targets with increasing number of dimensions.
{In the second setting, we study BaM for distributions with increasing (but controlled)}
amounts of non-Gaussianity.
{As evaluation metrics, we use empirical estimates of the KL divergence in both the forward direction, $\KL(p;q)$, and the reverse direction, $\KL(q;p)$.}

\paragraph{Gaussian targets with increasing dimensions}

We construct Gaussian targets
of increasing dimension with $D\!=\!4, 16, 64, 256$.
In \Cref{fig-gaussian-target}, we compare {BaM, ADVI, and GSM}
on each of these target distributions, plotting the
forward KL divergence against the number of gradient evaluations;
{
here we also consider two modified ADVI methods, where instead of the ELBO loss,
we use the score-based divergence (labeled as ``Score'') and the Fisher divergence (labeled as
``Fisher'').
}
Results for the reverse KL divergence and other parameter settings are {provided} in
\Cref{ssec-gaussian-expts}.
In all of these experiments, we {use a constant learning rate $\lambda_t=BD$ for BaM}.
Overall, we find that BaM {converges}  orders of magnitude faster than ADVI.
While GSM is competitive with BAM in some experiments,
{BaM converges more quickly with increasing batch size; this is unlike}
GSM which was observed to have
marginal gains beyond $B\!=\!2$ for Gaussian targets \citep{Modi2023}.

{We also observe that the gradient-based methods (ADVI, Score, Fisher) have similar performance
in terms of convergence,
and the score-based divergence is typically more sensitive to the learning rate.}
{In \Cref{ssec-wallclock}, we present wallclock timings for the methods, which show that
the gradient evaluations dominate the computational cost in lower-dimensional settings.}

\paragraph{Non-Gaussian targets with varying skew and tails}

\begin{figure*}[t]
    \centering
    \includegraphics[scale=0.42]{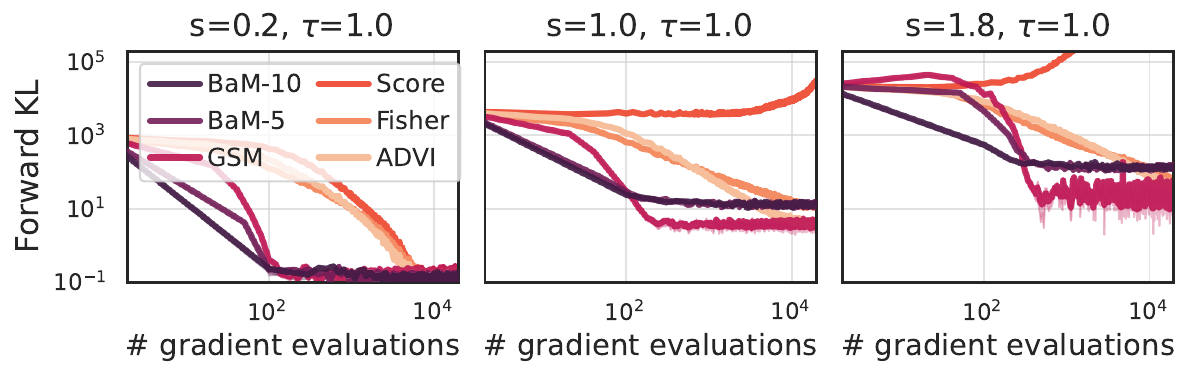}
    \includegraphics[scale=0.42]{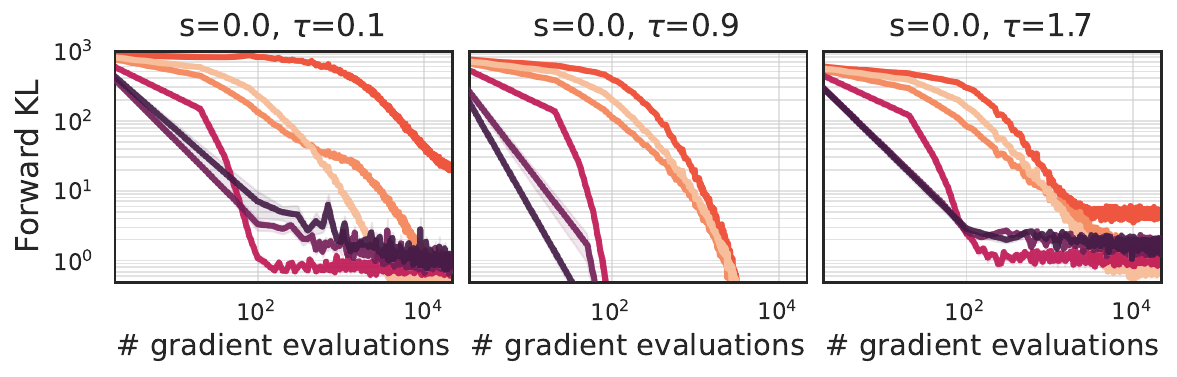}
\caption{Non-Gaussian targets constructed using the sinh-arcsinh
distribution, varying the skew $s$ and the tail weight $t$.
    The curves denote the mean of the forward KL divergence over
    10 runs, and shaded regions denote their standard error.
ADVI, Score, Fisher, and GSM use a batch size of $B\!=\! 5$.
    }
\label{fig-nongaussian-target}
\end{figure*}

The sinh-arcsinh normal distribution
transforms a Gaussian random variable via the hyperbolic sine function
and its inverse  \citep{jones2009sinh,jones2019sinh}.
{If $y\sim\N(\mu,\Sigma)$}, then a sample from the sinh-arcsinh normal distribution is
\begin{align*}
    z = \sinh\left(\tfrac{1}{\tau}\left(\sinh^{-1}(y)+s\right)\right),
\end{align*}
where the
{parameters $s\in\reals$ and $\tau\!>\!0$ control, respectively,
the skew and} the heaviness of the tails.
The Gaussian distribution is recovered when $s\!=\!0$ and $\tau\!=\!1$.

We construct {different non-Gaussian} target distributions
by varying {these parameters}.
The results are presented in \Cref{fig-nongaussian-target}
{and \Cref{fig-nongaussian-target2}}.
Here we {use a decaying learning rate $\lambda_t=BD / (t\!+\!1)$ for BaM,
as some decay is necessary for BaM to converge when the target distribution is non-Gaussian.}

First, we {construct target distributions with normal tails ($t\!=\!1$) but varying skew ($s\!=\!0.2,1.0,1.8$).}
{Here we observe that} BaM converges faster than ADVI.
For large skew ($s=1.0,1.8$),  BaM converges to a higher
{value of the} forward KL divergence but to similar values of the reverse KL divergence.
In these experiments, we see that GSM and ADVI often have similar performance but that
BaM {stabilizes more quickly with larger batch sizes}.
{Notably}, the reverse KL divergence for GSM diverges when the target distribution is highly skewed ($s\!=\!1.8$).
The Score method diverges for highly skewed targets as well, and we found this method to be more
sensitive to the learning rate.

Next we {construct target distributions with no skew ($s\!=\!0$) but tails of
varying heaviness ($t\!=\!0.1,0.9,1.7$).}
Here we find that all methods tend to converge to similar values of the
reverse KL divergence.
In some cases, BaM and ADVI converge to better values than GSM,
and BaM typically converges in fewer gradient evaluations than ADVI.

\subsection{Application:\ hierarchical Bayesian models}

We now consider the application of BaM to posterior inference.
Suppose we have observations $\{x_n\}_{n=1}^N$,
and the target distribution is the posterior density
\begin{align}
    p\left(z \given \{x_n\}_{n=1}^N\right) \propto p(z) \,  p\left(\{x_n\}_{n=1}^N \given z\right),
\end{align}
with prior $p(z)$ and likelihood $p(\{x_n\}_{n=1}^N \given z)$.
We examine three target distributions from \texttt{posteriordb} \citep{magnusson2022posteriordb},
a database of \texttt{Stan}
\citep{carpenter2017stan, roualdes2023bridgestan} models
with reference samples generated using Hamiltonian Monte Carlo (HMC).
The first target is nearly Gaussian (\texttt{arK}, $D\!=\!7$).
The other two targets are non-Gaussian: one is a Gaussian process (GP) Poisson
regression model (\texttt{gp-pois-regr}, $D\!=\!13$),
and the other is the 8-schools hierarchical Bayesian model (\texttt{eight-schools-centered}, $D\!=\!10$).

In these experiments, we evaluate BaM, ADVI, and GSM by computing the relative errors
of the posterior mean and standard deviation (SD) estimates with respect to those from HMC samples~\citep{welandawe2022robust};
we define these quantities and present additional results in \Cref{sec-posterior-db-details}.
We {use a decaying learning rate $\lambda_t=BD/(t\!+\!1)$ for BaM}.

\begin{figure*}[t]
    \centering
    \includegraphics[scale=0.48]{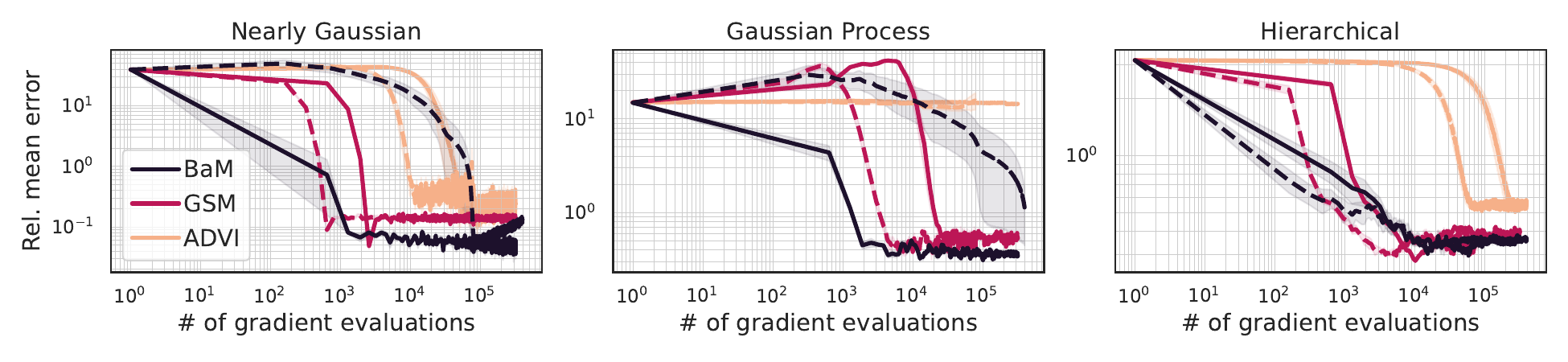}
\caption{Posterior inference in Bayesian models.
    The curves denote the mean over
    5 runs, and shaded regions denote their standard error.
    Solid curves ($B\!=\!32$)
    correspond to larger batch sizes than dashed curves ($B\!=\!8$).
}
\label{fig-pdb-target}
\end{figure*}

\Cref{fig-pdb-target} compares the relative mean errors of BaM, ADVI, and GSM
for batch sizes $B\!=\!8$ and $B\!=\!32$.
We observe that BaM outperforms ADVI.
For smaller batch sizes GSM can converge faster than BaM, but it oscillates around the solution.
{BaM performs better with increasing batch size, converging more quickly and to a more stable result,}
while GSM and ADVI do not benefit from increasing batch size.
In the appendix, we report the relative SD error and find similar results except that in the hierarchical example, BaM converges to a larger relative SD error.

\subsection{Application: deep generative model} \label{sec-deep-learning}

\begin{figure*}[t]
    \centering
    \begin{subfigure}[b]{\linewidth}
    \centering
    \includegraphics[scale=0.50]{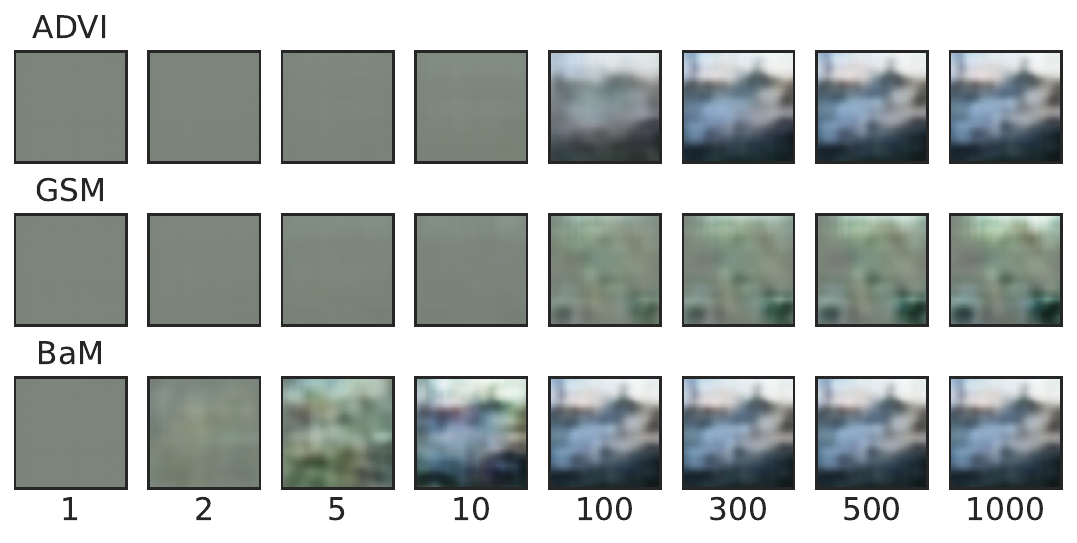}
    \caption{Image reconstruction across iterations with batch size, $B=300$.
    }
    \end{subfigure}
    \begin{subfigure}[b]{\linewidth}
    \centering
      \includegraphics[scale=0.55]{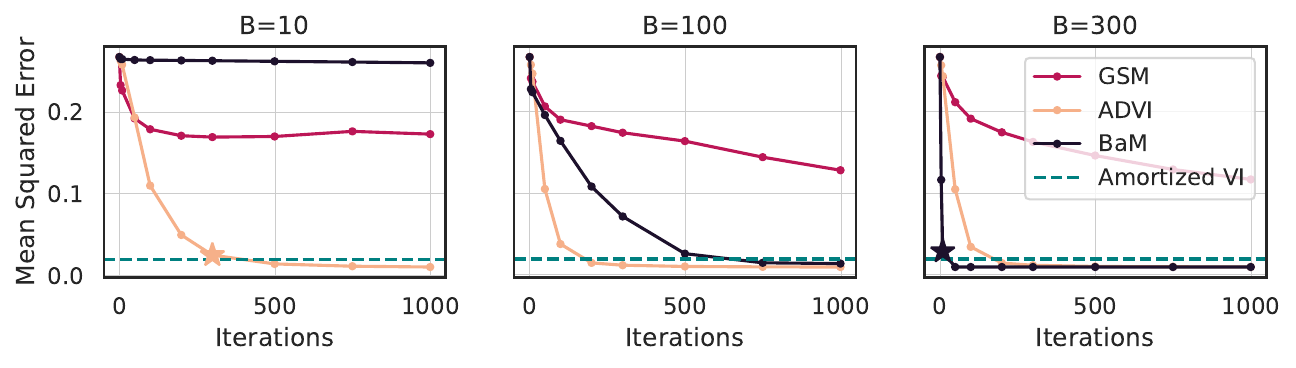}
      \caption{Reconstruction error for varying batch sizes.
}
    \end{subfigure}
    \caption{
    Image reconstruction  and error
        when the posterior mean of $z'$ is fed
        into the generative neural network.
        The beige and purple stars highlight the best outcome for ADVI and BaM, respectively, after 3,000 gradient evaluations.
    }
  \label{fig:image}
\end{figure*}

In a deep generative model, the likelihood is parameterized by the output of a neural network
$\Omega$, e.g.,
\begin{align}
    z_n &\sim \N(0, I)
    \\
    x_n \given z_n &\sim \N(\Omega(z_n, \hat{\theta}), \sigma^2 I),
\end{align}
where $x_n$ corresponds to a high-dimensional object, such as an image, and $z_n$ is a low-dimensional representation of $x_n$.
The neural network $\Omega$ is parameterized by $\hat \theta$ and maps $z_n$ to the mean of the likelihood $p(x_n | z_n)$.
For this example, we set $\sigma^2 = 0.1$.
The above joint distribution underlies many deep learning models
\citep{tomczak2022deep}, including the variational autoencoder \citep{kingma2013auto,rezende2014stochastic}.
We train the neural network on the CIFAR-10 image data set \citep{cifar10}.
We model the images as continuous, with $x_n \in \mathbb R^{3072}$,
and learn a latent representation $z_n \in \mathbb R^{256}$;
see \Cref{ssec-deep-learning} for details.

Given a new observation $x'$, we wish to approximate the posterior $p(z' | x')$.
As an evaluation metric, we examine how well $x'$ is reconstructed by feeding the
posterior expectation $\E[z' | x']$ into the neural network $\Omega(\cdot, \hat \theta)$.
The quality of the reconstruction is assessed visually and using the mean squared error (MSE,
    \Cref{fig:image});
{we present the MSE plotted against wallclock time in  \Cref{fig:image:time}.}
For ADVI and BaM, we use a pilot run of $T\!=\!100$ iterations to find a suitable learning rate; we then run the algorithms for $T\!=\!1000$ iterations.
(GSM does not require this tuning step.)
BaM performs poorly
when the batch size is very small ($B\!=\!10$) relative to the dimension of the latent
        variable~$z'$, but it becomes competitive as the batch size is increased. When the batch
        size is comparable to the dimension of $z_n$ (i.e. $B\!=\!300$), BaM converges an order of
        magnitude (or more) faster than ADVI and GSM.

To refine our comparison, suppose we have a computational budget of 3000 gradient evaluations.
Under this budget, ADVI achieves its lowest MSE for $B\!=\!10$ and $T\!=\!300$, while BaM produces a comparable result for $B\!=\!300$ and $T\!=\!10$.
Hence, the gradient evaluations for BaM can be largely parallelized.
By contrast, most gradients for ADVI must be evaluated sequentially.
{Notably, \Cref{fig:image:time} shows that BaM with $B=300$ converges faster in  wallclock time.}

Depending on how the parameter $\hat \theta$ of the neural network is estimated, it possible to learn an encoder and perform amortized variational inference (AVI) on a new observation $x'$.
When such an encoder is available, estimations of $p(z' | x')$ can be obtained essentially for free.
In our experiment, both BaM and ADVI eventually achieve a lower reconstruction error than AVI.
{This result is expected because our AVI implementation uses a factorized} Gaussian approximation,
whereas BaM and ADVI use a
{full-covariance} approximation, and the latter provides better compression of
$x'$ even though the dimension of $z'$ and the weights of the neural network
remain unchanged.



\section{Discussion and future work}

In this paper, we introduce  a score-based divergence
that {is especially well-suited to BBVI with Gaussian variational families}.
We show that the score-based divergence has a number of desirable properties.
We then propose  {a regularized optimization based on this divergence,}
and we show that it admits a closed-form solution, leading to a fast iterative
algorithm for score-based BBVI.
{We analyze the convergence of score-based BBVI}
when the target is Gaussian, and in the limit of an infinite batch size,
we show that the updates
converge exponentially quickly to the target mean and covariance.
Finally, we demonstrate the effectiveness of BaM in a number of empirical studies
involving both Gaussian and non-Gaussian targets;
here we observe that {for sufficiently large batch sizes},
our method converges much faster than other BBVI algorithms.

There are a number of fruitful directions for future work.
First, {it remains to analyze the convergence of BaM}
in the finite-batch case and for a larger class of target distributions.
Second, {it seems promising to develop score-based BBVI for other (non-Gaussian)
variational families, and
more generally, to study what divergences lend themselves to stochastic proximal point algorithms.}
Third, the BaM approach can be modified to utilize data subsampling (potentially with  control
    variates \citep{wang2022dual}) for
    large-scale Bayesian inference problems, where a noisy estimate of the target density's
    score is used in place of its exact score.

Finally, we note that the score-based divergence, which is computable for unnormalized models,
has useful applications beyond VI~\citep{hyvarinen2005estimation}; e.g., the
affine invariance property makes it attractive as
a goodness-of-fit diagnostic for inference methods.
Further study remains to characterize the relationship of the score-based divergence
to other such diagnostics
\citep{gorham2015measuring,liu2016kernelized,barp2019minimum,welandawe2022robust}.
%


\section*{Acknowledgements}

We thank Bob Carpenter, Ryan Giordano, and Yuling Yao for helpful discussions
and anonymous reviewers for their feedback on the paper.
This work was supported in part by
NSF IIS-2127869, NSF DMS-2311108, NSF/DoD PHY-2229929, ONR N00014-17-1-2131,
ONR N00014-15-1-2209, the Simons Foundation, and Open Philanthropy.

\bibliographystyle{plainnat-mod}
\bibliography{main}

\appendix





\section{Score-based divergence}
\label{app:divergence}

In \Cref{sec-divergence} we introduced a score-based divergence between two distributions, $p$ and $q$, over~$\mathbb{R}^D$, and specifically we considered the case where $q$ was Gaussian. In this section, we define this score-based divergence more generally. In particular, here we assume only that these distributions satisfy the following properties:
\begin{itemize}
\setlength\itemsep{-0.3ex}
\item[(i)] $p(\vec{z})\!>\!0$ and $q(\vec{z})\!>\!0$ for all $\vec{z}\in\mathbb{R}^D$.
\item[(ii)] $\vec{\nabla} p$ and $\vec{\nabla} q$ exist and are continuous everywhere in $\mathbb{R}^D$.
\item[(iii)] $\E_q\left[\|\vec{\nabla} \log q\|^2\right] < \infty$.
\end{itemize}
There may be weaker properties than these that also yield the following results
(or various generalizations thereof), but the above will suffice for our purposes.

This appendix is organized as follows. We begin with a lemma that is needed to define a score-based divergence for distributions (not necessarily Gaussian) satisfying the above properties. We then show that this score-based divergence has several appealing properties in its own right: it is nonnegative and invariant under affine reparameterizations, it takes a simple and intuitive form for distributions that are related by annealing or exponential tilting, and it reduces to the KL divergence in certain special cases.

\begin{flemma}
\noindent
The matrix defined by $\mat{\Gamma}_q = \E_q\left[(\vec{\nabla} \log q)(\vec{\nabla} \log q)^\top\right]$ exists in $\mathbb{R}^{D\times D}$ and is positive definite.
\label{lemma:Gamma}
\end{flemma}

\begin{proof}
Let $\vec{u}$ be any unit vector in $\mathbb{R}^D$. We shall prove the theorem by showing that \mbox{$0\!<\!\vec{u}^\top \mat{\Gamma}_q\vec{u}\!<\!\infty$}, or equivalently that all of the eigenvalues of $\mat{\Gamma}_q$ are finite and positive. The boundedness follows easily from property~(iii) since
\begin{equation}
\vec{u}^\top \mat{\Gamma}_q\vec{u}\
  =\ \E_q\left[(\vec{\nabla} \log q\cdot\vec{u})^2\right]\
  \leq\ \E_q\left[\|\vec{\nabla} \log q\|^2\right]\
  < \infty.
\end{equation}
To show positivity, we appeal to property~(ii) that $q$ is differentiable; hence for all $t>0$ we can write
\begin{equation}
q(t\vec{u}) = q(\vec{0}) + \int_0^t\!\! d\tau\ \vec{u}^\top \vec{\nabla} q(\tau\vec{u}) = q(\vec{0}) + \int_0^t\!\! d\tau\ q(\tau\vec{u})\, \vec{\nabla} \log q(\tau\vec{u})\cdot\vec{u}.
\end{equation}
To proceed, we take the limit $t\rightarrow\infty$ on both sides of this equation, and we appeal to property~(i) that $q(0)\!>\!0$. Moreover, since $\lim_{t\rightarrow\infty} q(t\vec{u})\! =\! 0$ for all normalizable distributions $q$, we see that
\begin{equation}
\int_0^\infty \!\! d\tau\ q(\tau\vec{u})\,\vec{\nabla} \log q(\tau\vec{u})\cdot\vec{u} < 0.
\label{eq:int}
\end{equation}
For this inequality to be satisfied, there must exist some $t_0\geq 0$ such that $\vec{\nabla} \log q(t_0\vec{u})\cdot\vec{u}<0$. Let $\vec{z}_0=t_0\vec{u}$, and let $\delta = -\vec{\nabla} \log q(\vec{z}_0)\cdot\vec{u}$. Since $q$ and $\vec{\nabla} q$ are continuous by properties (iii-iv), there must exist some finite ball~${\cal B}$ around $\vec{z}_0$ such that $\vec{\nabla} \log q(\vec{z})\cdot\vec{u}<-\frac{\delta}{2}$ for all $\vec{z}\in{\cal B}$. Let $q_{\cal B} = \min_{\vec{z}\in {\cal B}} q(\vec{z})$, and note that $q_{\cal B}>0$ since it is the minimum of a positive-valued function on a compact set. It follows that
\begin{equation}
\vec{u}^\top \mat{\Gamma}_q\vec{u}
  = \E_q\left[(\vec{\nabla} \log q\cdot\vec{u})^2\right]
   > q_{\cal B}\cdot\mbox{vol}({\cal B})\cdot \left(\tfrac{\delta}{2}\right)^2
  > 0,
\end{equation}
where the inequality is obtained by considering only those contributions to the expected value from within the volume of the ball~${\cal B}$ around $\vec{z}_0$.
This proves the lemma.
\end{proof}


The lemma is needed for the following definition of the score-based divergence. Notably, the definition assumes that the matrix $\E_q\left[(\vec{\nabla} \log q)(\vec{\nabla} \log q)^\top\right]$ is invertible.

\begin{fdefinition}[\textbf{Score-based divergence}]
Let $p$ and $q$ satisfy the properties listed above, and let $\mat{\Gamma}_q$ be defined as in \Cref{lemma:Gamma}.
    Then we define the \textit{score-based divergence} between $q$ and $p$ as
\begin{equation}
{\D}(q;p) = \E_q\left[ \left(\vec{\nabla}\log\tfrac{q}{p}\right)^\top \mat{\Gamma}_q^{-1}\left(\vec{\nabla}\log\tfrac{q}{p}\right)\right].
\label{eq:Dqp}
\end{equation}
\end{fdefinition}


Let us quickly verify that this definition reduces to the previous one in \Cref{sec-divergence} where $q$ is assumed to be Gaussian. In particular, suppose that $q={\cal N}(\nu,\Psi)$. In this case
\begin{equation}
\Gamma_q = \E_q[(\nabla\log q)(\nabla\log q)^\top] =
  \E_q[\Psi^{-1}(z\!-\!\nu)(z\!-\!\nu)^\top\Psi^{-1}] = \Psi^{-1}\Psi\Psi^{-1} = \Psi^{-1} = [\text{Cov}(q)]^{-1}.
\label{eq:GaussianGamma}
\end{equation}
Substituting this result into eq.~(\ref{eq:Dqp}), we recover the more specialized definition of the score-based divergence in \Cref{sec-divergence}.

We now return to the more general definition in eq.~(\ref{eq:Dqp}). Next we show this score-based divergence shares many desirable properties with the Kullback-Leibler divergence; indeed, in certain special cases of interest, these two divergences, ${\D}(q;p)$ and $\mbox{KL}(q;p)$, are equivalent. These properties are demonstrated in the following theorems. \\


\begin{ftheorem}[\textbf{Nonnegativity}]
${\D}(q;p) \geq 0$ with equality if and only if
    $p(\vec{z})=q(\vec{z})$ for all $\vec{z}\in\mathbb{R}^D$.
\end{ftheorem}

\begin{proof}
Nonnegativity follows from the previous lemma, and it is clear that the divergence vanishes if $p=q$. To prove the converse, we note that for any $\vec{z}\in\mathbb{R}^D$, we can write
\begin{equation}
\log \frac{p(\vec{z})}{q(\vec{z})}\ =\ \log\frac{p(\vec{0})}{q(\vec{0})} + \int_0^1\! dt\ \vec{\nabla}\log\left[\frac{p(t\vec{z})}{q(t\vec{z})}\right]\cdot\vec{z}.
\label{eq:lineInt}
\end{equation}
Now suppose that ${\D}(q;p)=0$. Then it must be the case that $\vec{\nabla}\log p = \vec{\nabla}\log q$ everywhere in $\mathbb{R}^D$. (If it were the case that $\vec{\nabla} \log p(\vec{z}_0)\neq\vec{\nabla}\log q(\vec{z}_0)$ for some $\vec{z}_0\in\mathbb{R}^D$, then by continuity, there would also exist some ball around $\vec{z}_0$ where these gradients were not equal; furthermore, in this case, the value inside the expectation of eq.~(\ref{eq:Dqp}) would be positive everywhere inside this ball, yielding a positive value for the divergence.) Since the gradients of $\log p$ and $\log q$ are everywhere equal, it follows from eq.~(\ref{eq:lineInt}) that
\begin{equation}
\log \frac{p(\vec{z})}{q(\vec{z})}\ =\ \log\frac{p(\vec{0})}{q(\vec{0})},
\end{equation}
or equivalently, that $p(\vec{z})$ and $q(\vec{z})$ have some constant ratio independent of $\vec{z}$. But this constant ratio must be equal to one because both distributions yield the same value when they are integrated over $\mathbb{R}^D$.
\end{proof}


\begin{ftheorem}[\textbf{Affine invariance}]
\label{theorem-affine}
 Let $f:\mathbb{R}^D\!\rightarrow\!\mathbb{R}^D$ be an affine transformation,
  and consider the induced densities
        $\tilde q(f(z)) \!=\! q(z) |\mathcal{J}(z)|^{-1}$
        and
        $\tilde p(f(z)) \!=\! p(z) |\mathcal{J}(z)|^{-1}$,
where $\mathcal{J}(z)$ is the determinant of the Jacobian of $f$.
        Then
        $\D(q; p) = \D(\tilde q; \tilde p)$.
\end{ftheorem}

\begin{proof}
Denote the affine transformation by $\tilde{\vec{z}} = \mat{A}\vec{z} + \vec{b}$ where $\mat{A}\in\mathbb{R}^{D\times D}$ and $\vec{b}\in\mathbb{R}^D$.  Then we have
\begin{equation}
\vec{\nabla}_{\vec{z}}\big[\log p(\vec{z})\big]
  = \vec{\nabla}_{\vec{z}}\!\left[\log\left(\tilde{p}(\tilde{\vec{z}})\left|\frac{d\tilde{\vec{z}}}{d\vec{z}}\right|\right)\right]
  = \vec{\nabla}_{\vec{z}}\!\left[\log\left(\tilde{p}(\tilde{\vec{z}})\,|\mat{A}|\right)\right]
  = \left(\frac{d\tilde{\vec{z}}}{d\vec{z}}\right)^{\!\!\top}\!\!\vec{\nabla}_{\tilde{\vec{z}}}\!\left[\log\tilde{p}(\tilde{\vec{z}})\right]
  = \mat{A}^{\!\top}\vec{\nabla}_{\tilde{\vec{z}}}\big[\log\tilde{p}(\tilde{\vec{z}})\big],\end{equation}
and a similar relation holds for $\vec{\nabla}_x \log q(z)$. It follows that
\begin{align}
{\D}(q;p)
  &= \E_q\!\left[ \left(\vec{\nabla}\log p \!-\! \vec{\nabla}\log q\right)^\top
  \left(\E_q\!\left[(\vec{\nabla} \log q)(\vec{\nabla} \log q)^\top\right]\right)^{-1} \left(\vec{\nabla}\log p\! -\! \vec{\nabla}\log q\right) \right]\\
  &= \E_{\tilde{q}}\!\left[ \left(\vec{\nabla}\log \tilde{p} \!-\! \vec{\nabla}\log \tilde{q}\right)^{\!\top}\!\!\! \mat{A}\!
  \left(\mat{A}^{\!\top}\E_{\tilde{q}}\!\left[(\vec{\nabla} \log \tilde{q})(\vec{\nabla} \log \tilde{q})^\top\right]\!\mat{A}\right)^{-1}\!\!
  \mat{A}^{\!\top}\!\!\left(\vec{\nabla}\log \tilde{p} \!-\! \vec{\nabla}\log \tilde{q}\right) \right]\!\\
 &= \E_{\tilde{q}}\!\left[ \left(\vec{\nabla}\log \tilde{p} \!-\! \vec{\nabla}\log \tilde{q}\right)^{\!\top}\!
  \left(\E_{\tilde{q}}\!\left[(\vec{\nabla} \log \tilde{q})(\vec{\nabla} \log \tilde{q})^\top\right]\right)^{-1}
 \left(\vec{\nabla}\log \tilde{p} \!-\! \vec{\nabla}\log \tilde{q}\right) \right]\\
  &= {\D}\big(\tilde{q},\tilde{p}\big).
\end{align}
Note the important role played by the matrix $\Gamma_q = \E_q\!\left[(\vec{\nabla} \log q)(\vec{\nabla} \log q)^\top\right]$ in this calculation. In particular, the unscaled quantity $\E_q[\|\vec{\nabla}\log p\!-\!\vec{\nabla}\log q\|^2]$ is not invariant under affine reparameterizations of $\mathbb{R}^D$.
\end{proof}


\begin{ftheorem}[\textbf{Annealing}]
If $p$ is an annealing of $q$, with $p\propto q^\beta$, then ${\D}(q;p) = D(\beta\!-\!1)^2$.
\end{ftheorem}
\begin{proof}
In this case $\vec{\nabla}\log p=\beta\vec{\nabla}\log q$. Thus, with $\mat{\Gamma}_q$ defined as in Lemma~\ref{lemma:Gamma}, we have
\begin{equation}
{\D}(q;p)
  = (\beta\!-\!1)^2\, \E_{q}\left[\left(\vec{\nabla}\log q\right)^\top\mat{\Gamma}_q^{-1}\left(\vec{\nabla}\log q\right)\right]
  = (\beta\!-\!1)^2\, \trace\left(\mat{\Gamma}_q^{-1}\mat{\Gamma}_q\right) = D(\beta\!-\!1)^2.
\label{eq:anneal}
\end{equation}
Here we see that ${\D}(q;p)$ measures the \textit{difference in inverse temperature} from the
    annealing. Note that in the limit $\beta\rightarrow 0$ of a uniform distribution,
    eq.~(\ref{eq:anneal}) yields a divergence of $D$ that is independent of the base distribution~$q$.
\end{proof}


\begin{ftheorem}[\textbf{Exponential tilting}]
If $p$ is an exponential tilting of $q$, with $p(\vec{z})\propto q(\vec{z})\,e^{\vec{\theta}^{\!\top}\!\vec{z}}$, then
${\D}(q;p) = \vec{\theta}^\top\mat{\Gamma}_q^{-1}\vec{\theta}$ where~$\mat{\Gamma}_q$ is defined as in Lemma~\ref{lemma:Gamma}.
\end{ftheorem}
\begin{proof}
In this case $\vec{\nabla}\log p\!-\!\vec{\nabla}\log q = \vec{\theta}$, and the result follows at once from substitution into eq.~(\ref{eq:Dqp}).
\end{proof}


\begin{fproposition}[\textbf{Gaussian score-based divergences}]
Suppose that $p$ is multivariate Gaussian with mean $\vec{\mu}$ and covariance $\mat{\Sigma}$ and that $q$ is multivariate Gaussian with mean $\vec{\nu}$ and covariance $\mat{\Psi}$, respectively.  Then
\begin{equation}
{\D}(q;p) =
    \trace\left[\left(\mat{I}-\mat{\Psi}\mat{\Sigma}^{-1}\right)^2\right] +
  (\vec{\nu}\!-\!\vec{\mu})^\top\mat{\Sigma}^{-1}\mat{\Psi}\mat{\Sigma}^{-1}(\vec{\nu}\!-\!\vec{\mu}).
\end{equation}
\end{fproposition}

\begin{proof}
We use the previous result in eq.~(\ref{eq:GaussianGamma}) that $\mat{\Gamma}_q\! =\! \Psi^{-1}$ when $q$ is Gaussian with covariance~$\Psi$. Then from eq.~(\ref{eq:Dqp}) the score-based divergence is given by
\begin{align}
{\D}(q;p)
  &= \E_q\left[ \left(\vec{\nabla}\log p\!-\!\vec{\nabla}\log q\right)^\top
  \mat{\Gamma}_q^{-1} \left(\vec{\nabla}\log p\!-\!\vec{\nabla}\log q\right) \right], \\
  &= \E_q\left[\left(\mat{\Sigma}^{-1}(\vec{z}\!-\!\vec{\mu})-\mat{\Psi}^{-1}(\vec{z}\!-\!\vec{\nu})\right)^\top\mat{\Psi}
    \left(\mat{\Sigma}^{-1}(\vec{z}\!-\!\vec{\mu})-\mat{\Psi}^{-1}(\vec{z}\!-\!\vec{\nu})\right)\right], \\
 &= \E_q\left[
   \left(\left(\mat{\Sigma}^{-1}\!-\!\mat{\Psi}^{-1}\right)(\vec{z}\!-\!\vec{\nu})-\mat{\Sigma}^{-1}(\vec{\mu}\!-\!\vec{\nu})\right)^\top\mat{\Psi}
     \left(\left(\mat{\Sigma}^{-1}\!-\!\mat{\Psi}^{-1}\right)(\vec{z}\!-\!\vec{\nu})-\mat{\Sigma}^{-1}(\vec{\mu}\!-\!\vec{\nu})\right)\right], \\
 &= \trace\left[\mat{\Psi}(\mat{\Sigma}^{-1}\!-\!\mat{\Psi}^{-1})\mat{\Psi}(\mat{\Sigma}^{-1}\!-\!\mat{\Psi}^{-1})\right]\, +\,
   (\vec{\nu}\!-\!\vec{\mu})^\top\mat{\Sigma}^{-1}\mat{\Psi}\mat{\Sigma}^{-1}(\vec{\nu}\!-\!\vec{\mu}), \\
   &=  \trace\left[\left(\mat{I}-\mat{\Psi}\mat{\Sigma}^{-1}\right)^2\right]\, +\, (\vec{\nu}\!-\!\vec{\mu})^\top\mat{\Sigma}^{-1}\mat{\Psi}\mat{\Sigma}^{-1}(\vec{\nu}\!-\!\vec{\mu}).
\end{align}
\end{proof}


\begin{fcorollary}[\textbf{Relation to KL divergence}]
Let $p$ and $q$ be multivariate Gaussian distributions with different means but the same covariance matrix. Then \mbox{$\frac{1}{2}{\D}(q;p) = {\rm KL}(q;p) = {\rm KL}(p;q)$}.
\end{fcorollary}
\begin{proof}
Let $\vec{\mu}$ and $\vec{\nu}$ denote, respectively, the means of $p$ and $q$, and let $\mat{\Sigma}$ denote their shared covariance. From the previous result, we find
\begin{equation}
{\D}(q;p) = (\vec{\nu}\!-\!\vec{\mu})^\top\mat{\Sigma}^{-1}(\vec{\nu}\!-\!\vec{\mu}).
\end{equation}
Finally, we recall the standard derivation for these distributions that
\begin{align}
{\rm KL}(q;p)
  &= \E_q\left[\log\tfrac{q}{p}\right] \\
  &= \tfrac{1}{2}\E_q\left[(\vec{z}\!-\!\vec{\nu})^\top\mat{\Sigma}^{-1}(\vec{z}\!-\!\vec{\nu}) - (\vec{z}\!-\!\vec{\mu})^\top\mat{\Sigma}^{-1}(\vec{z}\!-\!\vec{\mu}) \right] \\
  &= \tfrac{1}{2}\E_q\left[((\vec{z}\!-\!\vec{\mu})-(\vec{\nu}\!-\!\vec{\mu}))^\top\mat{\Sigma}^{-1}((\vec{z}\!-\!\vec{\mu})-(\vec{\nu}\!-\!\vec{\mu})) - (\vec{z}\!-\!\vec{\mu})^\top\mat{\Sigma}^{-1}(\vec{z}\!-\!\vec{\mu}) \right] \\
  &= \tfrac{1}{2}(\vec{\nu}\!-\!\vec{\mu})^\top\mat{\Sigma}^{-1}(\vec{\nu}\!-\!\vec{\mu}),
\end{align}
thus matching the result for $\frac{1}{2}{\D}(q;p)$. Moreover, we obtain the same result for ${\rm KL}(p;q)$ by noting that the above expression is symmetric with respect to the means $\vec{\mu}$ and $\vec{\nu}$.
\end{proof}

In sum, the score-based divergence $\D(q;p)$ in eq.~(\ref{eq:Dqp}) has several attractive properties as a measure of difference between most smooth distributions $p$ and $q$ with support on all of~$\mathbb{R}^D$.  First, it is nonnegative and equal to zero if and only if $p\!=\!q$. Second, it is invariant to affine reparameterizations of the underlying domain. Third, it behaves intuitively for simple transformations such as exponential tilting and annealing. Fourth, it is normalized such that every base distribution $q$ has the same divergence to (the limiting case of) a uniform distribution. Finally, it reduces to a constant factor of the KL divergence for the special case of two multivariate Gaussians with the same covariance matrix but different means.






\section{Quadratic matrix equations}
\label{app:quadratic}

In this appendix we show how to solve the quadratic matrix equation $XUX\!+\!X\!=\!V$ where $U$ and~$V$
are positive semidefinite matrices in $\mathbb{R}^{D\times D}$. We also verify certain
properties of these solutions that are needed elsewhere in the paper but that are not immediately
obvious. Quadratic matrix equations of this type (and of many generalizations thereof) have been
studied for decades~\citep{potter1966quadratic,kucera1972quadratic,kucera1972quadratic2,shurbet1974quadratic,yuan2021quadratic}, and our main goal here is to collect the results that we need in their simplest forms. These results are contained in the following four lemmas. \\


\begin{flemma}
\label{lemma-quadratic-psd-invertible}
Let $U\!\succeq\! 0$ and $V\!\succ\! 0$,
and suppose that $XUX\! +\! X\! =\! V$. Then a solution to this equation is given by
\begin{equation}
X = 2V\left[I + (I+4UV)^\frac{1}{2}\right]^{-1}.
\label{eq:qmsol}
\end{equation}
\end{flemma}

\begin{proof}
We start by turning the left side of the equation $XUX\!+\!X\!=\!V$ into a form that can be easily factored. Multiplying both sides by $U$, we see that
 \begin{equation}
 UXUX + UX = UV.
 \end{equation}
The next step is to complete the square by adding $\frac{1}{4} I$ to both sides; in this way, we find that
\begin{equation}
    \left(UX + \tfrac{1}{2} I \right)^2 = UV + \tfrac{1}{4} I.
 \label{eq:completeSquare}
\end{equation}
Next we claim that the matrix $UV\! +\! \tfrac{1}{4} I$ on the right side of eq.~(\ref{eq:completeSquare}) has all positive eigenvalues. To verify this claim, we note that
\begin{equation}
 \label{eq:allPos}
  UV +\tfrac{1}{4}I= V^{-\frac{1}{2}}\left(V^{\frac{1}{2}}UV^{\frac{1}{2}} + \tfrac{1}{4}I\right)V^{\frac{1}{2}}.
 \end{equation}
Thus we see that this matrix is similar to (and thus shares all the same eigenvalues as) the positive definite matrix $U^{\frac{1}{2}}VU^{\frac{1}{2}} + \tfrac{1}{4}I$ in parentheses on the right side of eq.~(\ref{eq:allPos}). Since the matrix has all positive eigenvalues, it has a unique principal square root, and from eq.~(\ref{eq:completeSquare}) it follows that
 \begin{equation}
     UX  = (UV + \tfrac{1}{4} I)^\frac{1}{2} - \tfrac{1}{2} I.
  \label{eq:UX}
 \end{equation}
If the matrix $U$ were of full rank, then we could solve for $X$ by left-multiplying both sides of eq.~(\ref{eq:UX}) by its inverse; however, we desire a general solution even in the case that $U$ is not full rank. Thus we proceed in a different way. In particular, we substitute the solution for $UX$ in eq.~(\ref{eq:UX}) into the original form of the quadratic matrix equation. In this way we find~that
 \begin{align}
  V &= XUX + X, \\[1ex]
     &= X(UX+I), \\
     &= X\left[\left(\left(UV + \tfrac{1}{4} I\right)^\frac{1}{2} - \tfrac{1}{2} I\right) + I\right], \\
     &= X\left[(UV + \tfrac{1}{4}I)^\frac{1}{2} + \tfrac{1}{2} I\right], \\
     &= \tfrac{1}{2} X\left[\left(4UV +  I\right)^\frac{1}{2} +  I\right].
     \label{eq:VX}
\end{align}
Finally we note that the matrix in brackets on the right side of eq.~(\ref{eq:VX}) has all positive eigenvalues; hence it is invertible, and after right-multiplying eq.~(\ref{eq:VX}) by its inverse we obtain the desired solution in eq.~(\ref{eq:qmsol}).
\end{proof}


\begin{flemma}
\label{lemma-qsol-psd}
The solution to $XUX\! +\! X\! =\! V$ in eq.~(\ref{eq:qmsol}) is symmetric and positive definite.
\end{flemma}

\begin{proof}
The key idea of the proof is to simultaneously diagonalize the matrices $U$ and $V^{-1}$ \textit{by
    congruence}. In particular, let~$\Lambda$ and $E$ be, respectively, the diagonal and orthogonal matrices satisfying
\begin{equation}
V^\frac{1}{2}UV^\frac{1}{2} = E\Lambda E^\top,
\end{equation}
where $\Lambda\!\succeq\! 0$. Now define $C\!=\!V^\frac{1}{2}E$.
It follows that $C^\top V^{-1}C\! =\! I$ and $C^\top U C\! =\! \Lambda$, showing that~$C$ simultaneously
    diagonalizes $V^{-1}$ and $U$ by congruence. Alternatively, we may use these relations to express $U$ and $V$
    in terms of $C$ and $\Lambda$~as
\begin{align}
V &= CC^\top, \label{eq:VCC} \\
U &= C^{-\!\top}\Lambda C^{-1}. \label{eq:UCDC}
\end{align}
We now substitute these expressions for $U$ and $V$ into the solution from eq.~(\ref{eq:qmsol}). The following calculation then gives the desired result:
\begin{align}
X &= 2V\left[I+(I+4UV)^{-\frac{1}{2}}\right]^{-1}, \\
  &= 2CC^\top\left[I + \left(I+4C^{-\!\top}\Lambda C^\top\right)^\frac{1}{2}\right]^{-1}, \\
  &= 2CC^\top\left[I + \left(C^{-\!\top}(I + 4\Lambda)C^\top\right)^\frac{1}{2}\right]^{-1}, \\
  &= 2CC^\top\left[I + C^{-\!\top}(I + 4\Lambda)^\frac{1}{2}C^\top\right]^{-1}, \\
  &= 2CC^\top\left[C^{-\!\top}\left(I + (I + 4\Lambda)^\frac{1}{2}\right)C^\top\right]^{-1}, \\
  &= 2CC^\top C^{-\top}\left[I + (I + 4\Lambda)^\frac{1}{2}\right]^{-1}C^\top, \\
  &= 2C\left[I + (I + 4\Lambda)^\frac{1}{2}\right]^{-1}C^\top. \label{eq:qmsol2}
\end{align}
Recalling that $\Lambda\!\succeq\! 0$, we see that the above expression for $X$ is manifestly symmetric and positive definite.
\end{proof}


Next we consider the cost of computing the solution to $XUX\!+\!X\!=\!V$ in eq.~(\ref{eq:qmsol}). On the right side of eq.~(\ref{eq:qmsol}) there appear both a matrix square root and a matrix inverse. As written, it therefore costs $O(D^3)$ to compute this solution when $U$ and $V$ are $D\!\times\!D$ matrices. However, if $U$ is of very low rank, there is a way to compute this solution much more efficiently. This possibility is demonstrated by the following lemma. \\

\begin{flemma}[\textbf{Low rank solver}]
\label{lemma-quadratic-psd-invertible-low}
Let $U\! =\! QQ^\top$ where $Q\!\in\! \mathbb{R}^{D\times K}$.
Then the solution in eq.~(\ref{eq:qmsol}), or equivalently in eq.~(\ref{eq:qmsol2}), can also be computed as
\begin{equation}
  \label{eq:quadmateq-update}
   X = V-   V^\top Q\left[\tfrac{1}{2} I + \left(Q^\top VQ + \tfrac{1}{4} I\right)^\frac{1}{2}\right]^{-2} Q^\top V.
\end{equation}
\end{flemma}

\vspace{1ex}
Before proving the lemma, we analyze the computational cost to evaluate eq.~(\ref{eq:quadmateq-update}).
Note that it costs $\mathcal{O}(KD^2)$ to compute the decomposition $U = QQ^\top$
as well as to form the product $Q^\top V$, while it costs $\mathcal{O}(K^3)$ to invert and take
square roots of $K\!\times\!K$ matrices.
Thus the total cost of eq.~(\ref{eq:quadmateq-update}) is $\mathcal{O}(KD^2\! +\! K^3)$,
in comparison to the $\mathcal{O}(D^3)$ cost of eq.~(\ref{eq:qmsol}).
This computational cost results in a potentially large savings if $K\!\ll\! D$. We now prove the lemma.

\begin{proof}
We will show that eq.~(\ref{eq:quadmateq-update}) is equivalent to eq.~(\ref{eq:qmsol2}) in the previous lemma. Again we appeal to the existence of an invertible matrix $C$ that simultaneously diagonalizes $V^{-1}$ and $U$ as in eqs.~(\ref{eq:VCC}--\ref{eq:UCDC}). If $U\!=\!QQ^\top$, then it follows from eq.~(\ref{eq:UCDC}) that
\begin{equation}
Q = C^{-\top}\Lambda^\frac{1}{2} R
\label{eq:QCDR}
\end{equation}
for some orthogonal matrix $R$. Next we substitute $V\!=\!CC^\top$ from eq.~(\ref{eq:VCC}) and $Q\!
    =\! C^{-\top}\Lambda^\frac{1}{2} R$ from eq.~(\ref{eq:QCDR}) in place of each appearance of $V$ and $Q$ in eq.~(\ref{eq:quadmateq-update}). In this way we find that
\begin{align}
X &= V - V^\top Q\left[\tfrac{1}{2} I + \left(Q^\top VQ + \tfrac{1}{4} I\right)^\frac{1}{2}
  \right]^{-2} Q^\top V, \\
  &= CC^\top - C\Lambda^\frac{1}{2} R
      \left[\tfrac{1}{2}I +
      \left((R^\top \Lambda^\frac{1}{2}C^{-1})(CC^\top)(C^{-\top}\Lambda^\frac{1}{2} R) +
       \tfrac{1}{4} I\right)^\frac{1}{2}
      \right]^{-2}\!\! R^\top \Lambda^\frac{1}{2}C^\top, \\
   &= C\left[I - {\Lambda}^\frac{1}{2}R\left[ \tfrac{1}{2}I + \left(R^\top \Lambda R +
    \tfrac{1}{4}I\right)^\frac{1}{2}\right]^{-2}R^\top \Lambda^\frac{1}{2}\right]C^\top, \\
   &= C\left[I - {\Lambda}^\frac{1}{2}R\left[ \tfrac{1}{2}I + R^\top\left(\Lambda +
    \tfrac{1}{4}I\right)^\frac{1}{2}R\right]^{-2}R^\top \Lambda^\frac{1}{2}\right]C^\top, \\
   &= C\left[I - {\Lambda}^\frac{1}{2}R\left[
       R^\top\left( \tfrac{1}{2}I + \left(\Lambda +
       \tfrac{1}{4}I\right)^\frac{1}{2}\right)R\right]^{-2}R^\top \Lambda^\frac{1}{2}\right]C^\top, \\
   &= C\left[I - {\Lambda}^\frac{1}{2}R\left[
       R^\top\left(\tfrac{1}{2}I + \left(\Lambda + \tfrac{1}{4}I\right)^\frac{1}{2}
       \right)^{2}R\right]^{-1}R^\top \Lambda^\frac{1}{2}\right]C^\top, \\
  &= C\left[I - {\Lambda}^\frac{1}{2}R\left[
       R^\top\left(\tfrac{1}{2}I + \left(\Lambda +
       \tfrac{1}{4}I\right)^\frac{1}{2}\right)^{-2}R\right]R^\top \Lambda^\frac{1}{2}\right]C^\top, \\
  &= C\left[I - \Lambda^\frac{1}{2}\left(\tfrac{1}{2}I + \left(\Lambda +
    \tfrac{1}{4}I\right)^\frac{1}{2}\right)^{-2}\Lambda^\frac{1}{2}\right]C^\top.
  \label{eq:qmsol3}
\end{align}
We now compare the matrices sandwiched between $C$ and $C^\top$ in eqs.~(\ref{eq:qmsol2}) and~(\ref{eq:qmsol3}). Both of these sandwiched matrices are diagonal, so it is enough to compare their corresponding diagonal elements.
    Let $\nu$ denote one element along the diagonal of~$\Lambda$.
    Then starting from eq.~(\ref{eq:qmsol3}), we see that
\begin{equation}
  1 - \frac{\nu}{\left(\tfrac{1}{2}+\sqrt{\nu+\tfrac{1}{4}}\right)^2}\
   =\  1 - \frac{4\nu}{(1+\sqrt{4\nu+1})^2}\
   =\ \frac{(1+\sqrt{4\nu+1})^2-4\nu}{(1+\sqrt{4\nu+1})^2}\
   =\ \frac{2}{1+\sqrt{4\nu+1}}.
  \label{eq:diags}
\end{equation}
Comparing the left and right terms in eq.~(\ref{eq:diags}), we see that the corresponding  elements of diagonal matrices in eqs.~(\ref{eq:qmsol2}) and (\ref{eq:qmsol3}) are equal, and we conclude that eqs.~(\ref{eq:qmsol}) and (\ref{eq:quadmateq-update}) yield the same solution.
\end{proof}


The last lemma in this appendix is one that we will need for the proof of
convergence of \Cref{alg:LS_GSM_VI} in the limit of infinite batch size.
In particular, it is needed to prove the sandwiching inequality in eq.~(\ref{eq:sandwich}).  \\

\begin{flemma} [\textbf{Monotonicity}]
\label{lemma-ABCXY}
Let $X$, $Y$, and $V$ be positive-definite matrices satisfying
$XTX + X = YUY + Y = V$, where $T\succeq U\succeq 0$. Then $X\preceq Y$.
\end{flemma}

\begin{proof}
The result follows from examining the solutions for $X$ and $Y$ directly. As shorthand, let $S=V^\frac{1}{2}$. By \Cref{lemma-quadratic-psd-invertible}, we have the solutions
\begin{align}
X &= 2S\left[I + \left(S + 4STS\right)^{\frac{1}{2}}\right]^{-1} S, \label{eq:solX}\\
Y &= 2S\left[I + \left(S + 4SUS\right)^{\frac{1}{2}}\right]^{-1} S. \label{eq:solY}
\end{align}
If $T\succeq U$, then the positive semi-definite ordering is preserved by the following chain of implications:
\begin{align}
STS &\succeq SUS, \\[0.5ex]
S + 4STS &\succeq S + 4SUS, \\
(S + 4STS)^\frac{1}{2} &\succeq (S+4SUS)^\frac{1}{2}, \label{eq:sqrt} \\
I + (S + 4STS)^\frac{1}{2} &\succeq I + (S+4SUS)^\frac{1}{2},
\end{align}
where in eq.~(\ref{eq:sqrt}) we have used the fact that positive semi-definite orderings are preserved by matrix square roots. Finally, these orderings are reversed by inverse operations, so that
\begin{equation}
\left[I + (S + 4STS)^\frac{1}{2}\right]^{-1} \preceq  \left[I + (S+4SUS)^\frac{1}{2}\right]^{-1}.
\label{eq:inverse_order}
\end{equation}
It follows from eq.~(\ref{eq:inverse_order}) and the solutions in eqs.~(\ref{eq:solX}--\ref{eq:solY}) that $X\preceq Y$, thus proving the lemma.
\end{proof}


\section{Derivation of batch and match updates}
\label{app:BaM}

In this appendix we derive the updates in \Cref{alg:LS_GSM_VI} for score-based variational
inference. The algorithm alternates between two steps---a \textsc{batch} step that draws samples
from an approximating Gaussian distribution and computes various statistics of these samples, and a
\textsc{match} step that uses these statistics to derive an updated Gaussian approximation, one that
better matches the scores of the target distribution. We explain each of these steps in turn, and
then we review the special case in which they reduce to the previously published updates~\citep{Modi2023} for Gaussian Score Matching (GSM).

\subsection{Batch step}

At each iteration, \Cref{alg:LS_GSM_VI} solves an optimization based on samples drawn from its current Gaussian approximation to the target distribution. Let $q_t$ denote this approximation at the $t^{\textrm{th}}$ iteration, with mean $\mu_t$ and covariance $\Sigma_t$, and let $z_1,z_2,\ldots,z_B$ denote the $B$ samples that are drawn from this distribution. The algorithm uses these samples to compute a (biased) empirical estimate of the score-based divergence between the target distribution, $p$, and another Gaussian approximation $q$ with mean~$\mu$ and covariance $\Sigma$. We denote this empirical estimate by
\begin{equation}
\widehat{\D}_{q_t}(q;p) =
   \frac{1}{B}\sum_{b=1}^B \Big\|\nabla \log q(z_b) - \nabla \log p(z_b)\Big\|^2_\Sigma.
    \label{eq-empirical-gsm}
\end{equation}
To optimize the Gaussian approximation $q$ that appears in this divergence, it is first necessary to evaluate the sum in eq.~(\ref{eq-empirical-gsm}) over the batch of samples $z_1,z_2,\ldots,z_B$ that have been drawn from~$q_t$.

The \textsc{batch} step of \Cref{alg:LS_GSM_VI} computes the statistics of these samples that enter into this calculation. Since $q$ is Gaussian, its score at the $b^{\textrm{th}}$ sample is given by $\nabla\log q(z_b)=-\Sigma^{-1}(z_b-\mu)$. As shorthand, let $g_b=\nabla\log p(z_b)$ denote the score of the target distribution at the $b^\textrm{th}$ sample. In terms of these scores, the sum in eq.~(\ref{eq-empirical-gsm}) is given by
\begin{equation}
\widehat{\D}_{q_t}(q;p)\
  =\ \frac{1}{B}\sum_{b=1}^B \Big\|-\Sigma^{-1}(z_b-\mu) - g_b\Big\|^2_\Sigma.
\label{eq:DBqp1}
\end{equation}
Next we show that $\widehat{\D}_{q_t}(q,p)$ depends in a simple way on certain first-order and second-order statistics of the samples, and it is precisely these statistics that are computed in the \textsc{batch} step. In particular,
we compute the following:

\begin{equation}
    \overline{z} = \frac{1}{B}\sum_{b=1}^B z_b,\quad
    \overline{g} = \frac{1}{B}\sum_{b=1}^B g_b,\quad
   C = \frac{1}{B}\sum_{b=1}^B (z_b - \overline{z})(z_b - \overline{z})^\top,\quad
  \Gamma = \frac{1}{B}\sum_{n=1}^N (g_b - \overline{g})(g_b - \overline{g})^\top.
 \label{eq:batch-stats}
\end{equation}

\vspace{1ex}
Note that the first two of these statistics compute the \textit{means} of the samples and scores in the current iteration of the algorithm, while the remaining two compute their \textit{covariance matrices}.
With these definitions, we can now express $\widehat{\D}_{q_t}(q,p)$ in an especially revealing form. Proceeding from eq.~(\ref{eq:DBqp1}), we have
\begin{align}
\widehat{\D}_{q_t}(q;p)
  &= \frac{1}{B}\sum_{b=1}^B \Big\|(\overline{g}-g_b) + \Sigma^{-1}(\overline{z} - z_b) + \Sigma^{-1}(\mu - \overline{z}-\Sigma\overline{g})\Big\|^2_\Sigma, \\
  &= \frac{1}{B}\sum_{b=1}^B \left[
            \big\|g_b-\overline{g}\big\|^2_\Sigma
         + \big\|z_b-\overline{z}\big\|^2_{\Sigma^{-1}}
         + \big\|\mu - \overline{z}-\Sigma\overline{g}\big\|^2_{\Sigma^{-1}}
         + 2(g_b-\overline{g})(z_b-\overline{z})\right], \\[0.5ex]
   &= \trace(\Gamma\Sigma)\, +\,
         \trace(C\Sigma^{-1})\, +\,
         \big\|\mu - \overline{z}-\Sigma\overline{g}\big\|^2_{\Sigma^{-1}}\, +\,
         \textrm{constant},
   \label{eq:DBqp2}
 \end{align}
 where in the second line we have exploited that many cross-terms vanish, and in the third line we have appealed to the definitions of $C$ and $\Gamma$ in  eqs.~(\ref{eq:batch-stats}). We have also indicated explicitly that the last term in eq.~(\ref{eq:DBqp2}) has no dependence on~$\mu$ and~$\Sigma$; it is a constant with respect to the approximating distribution $q$ that the algorithm seeks to optimize. This optimization is performed by the \textsc{match} step, to which we turn our attention next.

\subsection{Match step}
The \textsc{match} step of the algorithm updates the Gaussian approximation of VI to better match the recently sampled scores of the target distribution. The update at the $t^{\textrm{th}}$ iteration is computed as
\begin{equation}
q_{t+1} = \arg\!\min_{\!\!\!\!q\in\mathcal{Q}}\Big[ \mathscr{L}^\textrm{BaM}(q)\Big],
\label{eq:argminL}
\end{equation}
where $\mathcal{Q}$ is the Gaussian variational family of \Cref{sec-divergence} and $\mathscr{L}^\textrm{BaM}(q)$ is an objective function that balances the empirical estimate of the score based divergence in in eq.~(\ref{eq:DBqp2}) against a regularizer that controls how far $q_{t+1}$ can move away from~$q_t$.
Specifically, the objective function takes the form
\begin{equation}
\mathscr{L}^\textrm{BaM}(q) = \widehat{\D}_{q_t}(q;p) + \tfrac{2}{\lambda_t}\textrm{KL}(q_t;q),
\label{eq:LBaM}
\end{equation}
where the regularizing term is proportional to the KL divergence between the Gaussian distributions $q_t$ and $q$. This KL divergence is in turn given by the standard result
\begin{equation}
\textrm{KL}(q_t;q) = \tfrac{1}{2} \left[\trace(\Sigma^{-1}\Sigma_t) -
    \log\frac{|\Sigma_t|}{|\Sigma|} + \norm{\mu-\mu_t}_{\Sigma^{-1}}^2 - D
    \right].
 \label{eq:KLmatch}
\end{equation}
From eqs.~(\ref{eq:DBqp2}) and~(\ref{eq:KLmatch}), we see that this objective function has a complicated coupled dependence on $\mu$ and $\Sigma$; nevertheless, the optimal values of $\mu$ and $\Sigma$ can be computed in closed form. The rest of this section is devoted to performing this optimization.

First we perform the optimization with respect to the mean $\mu$, which appears quadratically in the objective $\mathscr{L}^\textrm{BaM}$ through the third terms in (\ref{eq:DBqp2}) and~(\ref{eq:KLmatch}). Thus we find
\begin{equation}
\frac{\partial\mathscr{L}^\textrm{BaM}}{\partial\mu}\,
  =\, \frac{\partial}{\partial\mu}\left\{
     \big\|\mu - \overline{z}-\Sigma\overline{g}\big\|^2_{\Sigma^{-1}} +
     \frac{1}{\lambda_t}  \norm{\mu-\mu_t}_{\Sigma^{-1}}^2\right\}\,
  =\ 2\Sigma^{-1}\!\left[\mu - \overline{z}-\Sigma\overline{g} + \tfrac{1}{\lambda_t}(\mu-\mu_t)\right].
\end{equation}
Setting this gradient to zero, we obtain a linear system which can be solved for the updated mean~$\mu_{t+1}$ in terms of the updated covariance~$\Sigma_{t+1}$. Specifically we find

\begin{equation}
\boxed{
\mu_{t+1}\ =\ \dfrac{\lambda_t}{1\!+\!\lambda_t}\left(\overline{z}+\Sigma_{t+1}\overline{g}\right)\,
    +\, \dfrac{1}{1\!+\!\lambda_t}\mu_t,
}
\label{eq:app_mu_update}
\end{equation}
matching eq.~(\ref{eq:update_mu1}) in \Cref{sec-algorithm} of the paper.
As a sanity check, we observe that in the limit of infinite regularization
($\lambda_t\!\rightarrow\!0$), the updated mean is equal to the previous mean (with $\mu_{t+1}\!=\!\mu_t$),
while in the limit of zero regularization $(\lambda_t\!\rightarrow\!\infty)$,
the updated mean is equal to precisely the value that zeros its contribution
to~$\widehat{\D}_{q_t}(q,p)$ in eq.~(\ref{eq:DBqp2}).

Next we perform this optimization with respect to the covariance $\Sigma$. To simplify our work, we first eliminate the mean $\mu$ from the optimization via eq.~(\ref{eq:app_mu_update}). When the mean is eliminated in this way from eqs.~(\ref{eq:DBqp2}) and~(\ref{eq:KLmatch}), we find that
\begin{align}
\widehat{\D}_{q_t}(q;p)
    &= \trace(\Gamma\Sigma)\, +\,
         \trace(C\Sigma^{-1})\, +\,
         \frac{1}{(1+\lambda_t)^2}
           \big\|\mu_t\! -\! \overline{z}\!-\!\Sigma\overline{g}\big\|^2_{\Sigma^{-1}}\, +\,
         \textrm{constant}, \\
 \textrm{KL}(q_t;q)
   &= \tfrac{1}{2} \left[\trace(\Sigma^{-1}\Sigma_t) - \log\frac{|\Sigma_t|}{|\Sigma|} +
        \frac{\lambda_t^2}{(1+\lambda_t)^2}
           \big\|\mu_t\! -\! \overline{z}\!-\!\Sigma\overline{g}\big\|^2_{\Sigma^{-1}} - D \right].
  \end{align}
Combining these terms via eq.~(\ref{eq:LBaM}), and dropping additive constants, we obtain an objective function of the covariance matrix $\Sigma$ alone. We denote this objective function by $\mathscr{M}(\Sigma)$, and it is given~by
\begin{equation}
\mathscr{M}(\Sigma) =
\trace(\Gamma\Sigma)\, +\,
         \trace\left(\left[C\!+\!\frac{1}{\lambda_t}\Sigma_t\right]\Sigma^{-1}\right)\, +\,
         \frac{1}{1+\lambda_t}\left(
           \big\|\mu_t\! -\! \overline{z}\big\|^2_{\Sigma^{-1}} +
           \big\|\overline{g}\big\|^2_{\Sigma}\right)\, +\,
          \frac{1}{\lambda_t}\log|\Sigma|.
\end{equation}
All the terms in this objective function can be differentiated with respect to $\Sigma$.  To minimize~$\mathscr{M}(\Sigma)$, we set its total derivative to zero. Doing this, we find that
\begin{equation}
0 = \Gamma + \frac{1}{1\!+\!\lambda_t}\overline{g}\,\overline{g}^\top - \Sigma^{-1}\left[C\!+\!\frac{1}{\lambda_t}\Sigma_t\!+\!\frac{1}{1\!+\!\lambda_t}(\mu_t\!-\!\overline{z})(\mu_t\!-\!\overline{z})^\top\right]\Sigma^{-1} + \frac{1}{\lambda_t}\Sigma^{-1}.
\end{equation}
The above is a quadratic matrix equation for the inverse covariance matrix $\Sigma^{-1}$; multiplying on the left and right by $\Sigma$, we can rewrite it as a quadratic matrix equation for $\Sigma$.
In this way we find that
\begin{equation}
\boxed{
\Sigma U\Sigma + \Sigma = V\quad\mbox{where}\quad\left\{
\begin{array}{l}
 U = \lambda_t\Gamma + \dfrac{\lambda_t}{1\!+\!\lambda_t}\overline{g}\,\overline{g}^\top,\\[2ex]
 V =   \Sigma_t + \lambda_t C +
    \dfrac{\lambda_t}{1\!+\!\lambda_t}(\mu_t\!-\!\overline{z})(\mu_t\!-\!\overline{z})^\top,
 \end{array}
 \right.
 \label{eq:app_Sigma_update}
}
\end{equation}
matching eq.~(\ref{eq:BaM-quadratic}) in \Cref{sec-algorithm} of the paper.
The solution to this quadratic matrix equation is given by \Cref{lemma-quadratic-psd-invertible}, yielding the update rule
\begin{equation}
\boxed{
\Sigma_{t+1} = 2V\left[I+(I+4UV)^\frac{1}{2}\right]^{-1}
}
\end{equation}
and matching eq.~(\ref{eq:update_Sigma1}) in \Cref{sec-algorithm} of the paper.
Moreover, this solution is guaranteed to be symmetric and positive definite by \Cref{lemma-qsol-psd}.


\subsection{Gaussian score matching as a special case}

In this section, we show that the updates for BaM include the
updates for GSM~\citep{Modi2023} as a limiting case. In BaM, this limiting case occurs when there is
no regularization ($\lambda\!\rightarrow\!\infty$) and when the batch size is equal to one ($B\!=\!1$). In this case,  we show that the updates in
eqs.~(\ref{eq:app_mu_update}) and (\ref{eq:app_Sigma_update}) coincide with those of GSM.

To see this equivalence, we set $B\!=\!1$, and we use $z_t$ and $g_t$ to denote, respectively, the single sample from $q_t$ and its score under $p$ at the $t^\textrm{th}$ iteration of BaM. The equivalence arises from a simple intuition: as $\lambda\!\rightarrow\!\infty$, all the weight in the loss shifts
to minimizing the divergence~$\widehat{\D}_{q_t}(q;p),$ which is then minimized
exactly so that
$\widehat{\D}_{q_t}(q;p)\! =\!0.$ More formally, in this limit the batch step can be written as
\begin{equation}
\lim_{\lambda\rightarrow\infty}\min_{q \in \mathcal{Q}}\left[ \widehat{\D}_{q_t}(q;p) + \frac{2}{\lambda_t}\textrm{KL}(q_t;q)\right]\, =\,
\min_{q \in \mathcal{Q}}\big[ \textrm{KL}(q_t;q)\big]\  \mbox{such that}\ \widehat{\D}_{q_t}(q;p)\! =\!0.
\end{equation}
The divergence term $\widehat{\D}_{q_t}(q;p)$ only vanishes when the scores match exactly; thus the above can be re-written as
\begin{equation}
\min_{q \in \mathcal{Q}} \big[ \textrm{KL}(q_t;q)\big]\ \mbox{such that}\ \nabla \log q(z_t)\! =\! \nabla \log p(z_t),
\end{equation}
which is exactly the variational formulation of the GSM method~\citep{Modi2023}

We can also make this equivalence more precise by studying the resulting update. Indeed,
the batch statistics in eq.~(\ref{eq:batch-stats}) simplify in this setting: namely, we have $\overline{z}=z_t$ and $\overline{g}=g_t$ (because there is only one sample) and $C\!=\!\Gamma\!=\!0$ (because the batch has no variance). Next we take the limit $\lambda_t\!\rightarrow\!\infty$ in eq.~(\ref{eq:app_Sigma_update}).
In this limit we find that
\begin{align}
U &= g_t g_t^\top,\\
V &= \Sigma_t + (\mu_t\!-\!z_t)(\mu_t\!-\!z_t)^\top,
\end{align}
so that the covariance is updated by solving the quadratic matrix equation
\begin{equation}
\Sigma_{t+1}g_t g_t^\top\Sigma_{t+1} + \Sigma_{t+1} = \Sigma_t + (\mu_t\!-\!z_t)(\mu_t\!-\!z_t)^\top.
\label{eq:gsm_cov}
\end{equation}
Similarly, taking the limit $\lambda_t\!\rightarrow\!\infty$ in eq.~(\ref{eq:app_mu_update}), we see that the mean is updated as
\begin{equation}
\mu_{t+1} = \Sigma_{t+1}g_t + z_t.
\label{eq:gsm_mean}
\end{equation}
These BaM updates coincide exactly with the updates for GSM: specifically, eqs.~(\ref{eq:gsm_cov})
and~(\ref{eq:gsm_mean}) here are identical to eqs.~(42) and (23) in \citet{Modi2023}.







\section{Proof of convergence}
\label{app:proof-converge}

In this appendix we provide full details for the proof of convergence in
\Cref{thm:converge}. We repeat equations freely from earlier parts of the paper
when it helps to make the appendix more self-contained. Recall that the target
distribution in this setting is assumed to be Gaussian with mean $\mup$ and
covariance~$\Sigmap$; in addition, we measure the normalized errors at the $t^{\rm th}$ iteration by
\begin{align}
    \varepsilon_{t} &= \Sigmap^{-\frac{1}{2}} (\mu_t - \mup), \label{eq:epsilon_t_2} \\
    \Delta_{t} &= \Sigma^{-\frac{1}{2}} \Sigma_t \Sigma^{-\frac{1}{2}} - I. \label{eq:Delta_t_2}
\end{align}
If the mean and covariance iterates of \Cref{alg:LS_GSM_VI} converge to those of the target distribution, then equivalently the norms of these errors must converge to zero. Many of our intermediate results are expressed in terms of the matrices
\begin{equation}
J_t = \Sigmap^{-\frac{1}{2}} \Sigma_t \Sigmap^{-\frac{1}{2}},\label{eq:J_t_2}
\end{equation}
which from eq.~(\ref{eq:Delta_t_2}) we can also write as $J_t = I+\Delta_t$.
For convenience we restate the theorem in section \ref{ssec:thm-restate}; our main result is that in the limit of an infinite batch size, the norms of the errors in eqs.~(\ref{eq:epsilon_t_2}--\ref{eq:Delta_t_2}) decay exponentially to zero with rates that we can bound from below.

The rest of the appendix is organized according to the major steps of the proof as sketched in section~\ref{sec:proof-converge}. In section~\ref{ssec-infinite-batch}, we examine the statistics that are computed by \Cref{alg:LS_GSM_VI} when the target distribution is Gaussian and the number of batch samples goes to infinity. In section~\ref{ssec-recursions-infinite}, we derive the recursions that are satisfied for the normalized mean $\varepsilon_t$ and covariance~$J_t$ in this limit. In section~\ref{ssec:sandwich}, we derive a sandwiching inequality for positive-definite matrices that arise in the analysis of these recursions. In section~\ref{ssec:eig-bounds}, we use the sandwiching inequality to derive upper and lower bounds on the eigenvalues of~$J_t$. In section~\ref{ssec:errors-recurse}, we use these eigenvalue bounds to derive how the normalized errors~$\varepsilon_t$ and~$\Delta_t$ decay from one iteration to the next. In section~\ref{ssec:induction}, we use induction on these results to derive the final bounds on the errors in eqs.~(\ref{eq:epsilon_converge_2}--\ref{eq:Delta_converge_2}), thus proving the theorem. In the more technical sections of the appendix, we sometimes require intermediate results that digress from the main flow of the argument; to avoid too many digressions, we collect the proofs for all of these intermediate results in section~\ref{ssec:supporting-lemmas}.


\subsection{Main result}
\label{ssec:thm-restate}

Recall that our main result is that as $B\rightarrow \infty$, the spectral norms of the
normalized mean and covariance errors in
decay exponentially to zero with rates that we can bound from below.

\begin{ftheorem}[\textbf{Restatement of~\Cref{thm:converge}}]
Suppose that $p = \N(\mup,\Sigmap)$ in \Cref{alg:LS_GSM_VI},
    and let $\alpha\!>\!0$ denote the minimum eigenvalue of the matrix
    $\Sigmap^{-\frac{1}{2}}\mat{\Sigma}_0 \Sigmap^{-\frac{1}{2}}$.
    For any fixed level of regularization $\lambda\!>\!0$, define
\begin{align}
\beta &:= \min\left(\alpha,\frac{1\!+\!\lambda}{1\!+\!\lambda\!+\!\|\varepsilon_0\|^2}\right), \\
\delta &:= \frac{\lambda\beta}{1\!+\!\lambda},
\end{align}
where $\beta\in(0,1]$ measures the quality of initialization and
    $\delta\in(0,1)$ denotes a rate of decay.
    Then with probability 1 in the limit of infinite batch size ($B\!\rightarrow\!\infty$),
    and for all $t\!\geq 0$, the normalized errors in eqs.~(\ref{eq:epsilon_t_2}--\ref{eq:Delta_t_2}) satisfy
\begin{align}
\|\vec{\varepsilon}_t\| &\leq (1\!-\!\delta)^{t} \|\vec{\varepsilon}_0\|, \label{eq:epsilon_converge_2} \\
\|\mat{\Delta}_t\| &\leq (1\!-\!\delta)^t \|\mat{\Delta}_0\|\,
  +\, t(1\!-\!\delta)^{t-1}  \|\vec{\varepsilon}_0\|^2.\hspace{-2ex} \label{eq:Delta_converge_2}
\end{align}
\label{thm:converge2}
\end{ftheorem}

We emphasize that the theorem holds under very general conditions: it is true no matter how the variational parameters are initialized (assuming only that they are finite and that the initial covariance estimate is not singular), and it is true for any fixed degree of regularization $\lambda\!>\!0$. Notably, the value of $\lambda$ is \textit{not} required to be inversely proportional to the largest (but a priori unknown) eigenvalue of some Hessian matrix, an assumption that is typically needed to prove the convergence of most \textit{gradient-based} methods. This stability with respect to hyperparameters is a well-known property of proximal algorithms, one that has been previously observed beyond the setting of variational inference in this paper.

Finally we note that the bounds in eqs.~(\ref{eq:epsilon_converge_2}--\ref{eq:Delta_converge_2}) can be tightened with more elaborate bookkeeping and also extended to updates that use varying levels of regularization $\{\lambda_t\}_{t=0}^\infty$ at different iterations of the algorithm. At various points in what follows, we indicate how to strengthen the results of the theorem along these lines. Throughout this section, we use the matrix norm $\norm{\cdot}$ to denote the spectral norm, and we use the notation $\nu_{\text{min}}(J)$  and $\nu_{\text{max}}(J)$ to denote the minimum and maximum eigenvalues of a matrix~$J$.


\subsection{Infinite batch limit}
\label{ssec-infinite-batch}

The first step of the proof is analyze how the statistics computed at each iteration of \Cref{alg:LS_GSM_VI} simplify in the infinite batch limit ($B\!\rightarrow\!\infty$).  Let $q_t$ denote the Gaussian variational approximation at the $t^{\rm th}$ iteration of the algorithm, let $z_b\sim\N(\mu_t,\Sigma_t)$ denote the $b^{\rm th}$ sample from this distribution, and let $g_b = \nabla\log p(z_b)$ denote the corresponding score of the target distribution~$p$ at this sample. Recall that step~5 of \Cref{alg:LS_GSM_VI} computes the following batch statistics:
\begin{align}
    \overline{z}_{B} &= \frac{1}{B} \sum_{b=1}^B z_b,  \quad\quad 
    C_{B} = \frac{1}{B} \sum_{b=1}^B (z_b-\overline{z}_B) (z_b-\overline{z}_B)^\top, \label{eq:x_batch} \\[0.5ex]
    \overline{g}_{B} &= \frac{1}{B} \sum_{b=1}^B g_b, \quad\quad 
    \Gamma_{B} = \frac{1}{B} \sum_{b=1}^B (g_b-\overline{g}_B) (g_b-\overline{g}_B)^\top, \label{eq:g_batch}
\end{align}
Here we use the subscript on these averages to explicitly indicate the batch size. (Also, to avoid an excess of indices, we do not explicitly indicate the iteration $t$ of the algorithm.) These statistics simplify considerably when the target distribution is multivariate Gaussian and the number of batch samples goes to infinity. In particular, we obtain the following result. \\

\begin{flemma}[\textbf{Infinite batch limit}]
\label{lemma-limiting-statistics}
Suppose $p\!=\!\N(\mup,\Sigmap)$. Then with probability 1, as the number of
    batch samples goes to infinity $(B\!\rightarrow\!\infty)$,
the statistics in eqs.~(\ref{eq:x_batch}--\ref{eq:g_batch}) tend to
\begin{align}
    \lim_{B \rightarrow \infty} \overline{z}_B &= \mu_t, \label{eq:xB_limit} \\
    \lim_{B \rightarrow \infty} C_B &= \Sigma_t, \label{eq:CB_limit} \\
    \lim_{B \rightarrow \infty} \overline{g}_B &= \Sigmap^{-1}(\mup - \mu_t), \label{eq:gB_limit}  \\
   \lim_{B \rightarrow \infty} \Gamma_B &= \Sigmap^{-1}\Sigma_t \Sigmap^{-1}. \label{eq:GammaB_limit}
\end{align}
\end{flemma}

\begin{proof}
The first two of these limits follow directly from the strong law of large numbers.
    In particular, for the sample mean in eq.~(\ref{eq:x_batch}), we have with probability 1 that
\begin{equation}
    \lim_{B \rightarrow \infty} \overline{z}_B
      =  \lim_{B\rightarrow\infty} \left[ \frac{1}{B} \sum_{b=1}^B  z_{b}\right]
      =  \int z \, q_t(dz)
      = \mu_t, \label{eq:calc_xB_limit}
 \end{equation}
thus yielding eq.~(\ref{eq:xB_limit}). Likewise for the sample covariance in eq.~(\ref{eq:x_batch}), we have with probability~1 that
 \begin{equation}
    \lim_{B \rightarrow \infty} C_B
      = \lim_{B\rightarrow\infty} \left[\frac{1}{B} \sum_{b=1}^B (z_b\! -\! \overline{z}_B) (z_b\! -\! \overline{z}_B)^\top\right]
     = \int (z\! -\! \mu_t) (z\! -\! \mu_t)^\top q_t(dz)
     = \Sigma_t, \label{eq:calc_CB_limit}
\end{equation}
thus yielding eq.~(\ref{eq:CB_limit}).
Next we consider the infinite batch limits for $\overline{g}_B$ and $\Gamma_B$, in
eq.~(\ref{eq:g_batch}), involving the scores of the target distribution.
Note that if this target distribution is multivariate Gaussian, with $p=\N(\mup,\Sigmap)$, then we have
\begin{equation}
g_b = \nabla\log p(z_b) = \Sigmap^{-1}(\mup\!-\!z_b),
\label{eq:gb_app}
\end{equation}
showing that the score $g_b$ is a linear function of $z_b$.
Thus the infinite batch limits $\overline{g}_B$ and~$\Gamma_B$ follow directly
    from those for~$\overline{z}_B$ and $C_B$.
In particular, combining eq.~(\ref{eq:gb_app}) with the calculation in eq.~(\ref{eq:calc_xB_limit}), we see that
\begin{equation}
\lim_{B\rightarrow\infty} \overline{g}_B
     = \lim_{B\rightarrow\infty}\left[ \frac{1}{B} \sum_{b=1}^B g_b\right]
      = \lim_{B\rightarrow\infty}\Big[\Sigmap^{-1}(\mup-\overline{z}_B)\Big]
      = \Sigmap^{-1}(\mup-\mu_t)
 \end{equation}
for the mean of the scores in this limit, thus yielding eq.~(\ref{eq:gB_limit}).
Likewise, by the same reasoning, we see that
\begin{equation}
    \lim_{B\rightarrow\infty} {\Gamma}_B
     = \lim_{B\rightarrow\infty}\left[ \frac{1}{B} \sum_{b=1}^B (g_b-\overline{g}_B) (g_b-\overline{g}_B)^\top\right]
     = \lim_{B\rightarrow\infty}\Sigmap^{-1}C_B\Sigmap^{-1}
      = \Sigmap^{-1}\Sigma_t\Sigmap^{-1}\label{eq:calc_Gamma_B_limit}
 \end{equation}
for the covariance of the scores in this limit, thus yielding eq.~(\ref{eq:GammaB_limit}). This proves the lemma.
\end{proof}


\subsection{Recursions for $\varepsilon_t$ and $J_t$}
\label{ssec-recursions-infinite}

Next we use~\Cref{lemma-limiting-statistics} to derive recursions for the normalized error $\varepsilon_t$ in eq.~(\ref{eq:epsilon_t_2}) and the normalized covariance $J_t$ in eq.~(\ref{eq:J_t_2}). Both follow directly from our previous results. \\

\begin{fproposition}[\textbf{Recursion for $\varepsilon_t$}]
\label{prop-recursion-mean}
Suppose $p\!=\!\N(\mup,\Sigmap)$, and let $B\!\rightarrow\!\infty$ in \Cref{alg:LS_GSM_VI}. Then with probability 1, the normalized error at the $(t\!+\!1)^{\rm th}$ iteration of satisfies
 \begin{equation}
 \varepsilon_{t+1} = \left[I-\frac{\lambda_t}{1+\lambda_t} J_{t+1}\right]\varepsilon_t.
 \label{eq:epsilon_recurse_2}
\end{equation}
\end{fproposition}

\begin{proof}
Consider the update for the variational mean in step 7 of \Cref{alg:LS_GSM_VI}.
We begin by computing the infinite batch limit of this update.
Using the limits for $\overline{z}_B$ and $\overline{g}_B$ from~\Cref{lemma-limiting-statistics}, we see that
\begin{align}
\mu_{t+1}
  &= {\lim_{B\rightarrow\infty}}\left[\left(\frac{1}{1\!+\!\lambda_t}\right)\mu_t + \left(\frac{\lambda_t}{1\!+\!\lambda_t}\right)\big(\Sigma_{t+1}{\overline{g}_B} + {\overline{z}_B}\big)\right], \\[1ex]
  &= \left(\frac{1}{1\!+\!\lambda_t}\right)\mu_t +
      \left(\frac{\lambda_t}{1\!+\!\lambda_t}\right)\Big(\Sigma_{t+1}\Sigmap^{-1}(\mup\!-\!\mu_t) + \mu_t\Big),  \\[1ex]
  &= \mu_t +  \frac{\lambda_t}{1\!+\!\lambda_t}\Sigma_{t+1}\Sigmap^{-1}(\mup\!-\!\mu_t). \label{eq:mu_update_infinite_batch}
\end{align}
The proposition then follows by substituting eq.~(\ref{eq:mu_update_infinite_batch}) into the definition of the normalized error in eq.~(\ref{eq:epsilon_t_2}):
\begin{align}
\varepsilon_{t+1}
  &= \Sigmap^{-\frac{1}{2}}(\mu_{t+1}\!-\!\mup), \\[1ex]
  &= \Sigmap^{-\frac{1}{2}}\left[ \mu_t +
    \frac{\lambda_t}{1\!+\!\lambda_t}\Sigma_{t+1}\Sigmap^{-1}(\mup\!-\!\mu_t)- \mup\right], \\
  &= \left[ I -
    \frac{\lambda_t}{1\!+\!\lambda_t}\Sigmap^{-\frac{1}{2}}\Sigma_{t+1}\Sigmap^{-\frac{1}{2}}\right]\Sigmap^{-\frac{1}{2}}(\mu_t\!-\!\mup), \\[1ex]
  &= \left[ I - \frac{\lambda_t}{1\!+\!\lambda_t}J_{t+1}\right]\varepsilon_t.
\end{align}
This proves the proposition, and we note that this recursion takes the same form as eq.~(\ref{eq:epsilon_update}), in the proof sketch of \Cref{thm:converge}, if a fixed level of regularization is used at each iteration.
\end{proof}

\begin{fproposition}[\textbf{Recursion for $J_t$}]
\label{prop-recursion-cov}
Suppose $p\!=\!\N(\mup,\Sigmap)$, and let $B\!\rightarrow\!\infty$ in \Cref{alg:LS_GSM_VI}. Then with probability 1, the normalized covariance at the $(t\!+\!1)^{\rm th}$ iteration of satisfies
 \begin{equation}
\lambda_t J_{t+1}\left(J_t + \frac{1}{1\!+\!\lambda_t}\varepsilon_t\varepsilon_t^\top\right)J_{t+1} + J_{t+1} = (1\!+\!\lambda_t)J_t
 \label{eq:J_recurse_2}
\end{equation}
\end{fproposition}

\begin{proof}
Consider the quadratic matrix equation, from step 6 of \Cref{alg:LS_GSM_VI}, that is satisfied by the variational covariance after $t\!+\!1$ updates:
\begin{equation}
\Sigma_{t+1} U_B \Sigma_{t+1} + \Sigma_{t+1} = V_B.
\label{eq:quad-matrix}
\end{equation}
We begin by computing the infinite batch limit of the matrices, $U_B$ and $V_B$, that appear in this equation. Starting from eq.~(\ref{eq:V1}) for~$V_B$, and using the limits for $\overline{z}_B$ and $C_B$ from~\Cref{lemma-limiting-statistics}, we see that
\begin{align}
\lim_{B\rightarrow\infty} V_B
  &= \lim_{B\rightarrow\infty}\left[\Sigma_t + \lambda_t C_B + \frac{\lambda_t}{1\!+\!\lambda_t}
        (\mu_t-\overline{z}_B)(\mu_t-\overline{z}_B)^\top\right], \\[1ex]
   &= (1\!+\!\lambda_t)\Sigma_t, \\[1ex]
   &= \Sigmap^{\frac{1}{2}}\big[(1\!+\!\lambda_t) J_t\big] \Sigmap^{\frac{1}{2}},
\end{align}
where in the last line we have used eq.~(\ref{eq:J_t_2}) to re-express the right side in terms of $J_t$. Likewise, starting from eq.~(\ref{eq:U1}) for~$U_B$, and using the limits for $\overline{g}_B$ and $\Gamma_B$ from~\Cref{lemma-limiting-statistics}, we see that
\begin{align}
\lim_{B\rightarrow\infty} U_B
  &=  \lim_{B\rightarrow\infty} \left[\lambda_t \Gamma_B + \frac{\lambda_t}{1\!+\!\lambda_t} \, \bar{g}_B \, \bar{g}_B^\top\right] \\[1ex]
  &= \lambda_t \Sigmap^{-1}\Sigma_t\Sigmap^{-1} + \frac{\lambda_t}{1\!+\!\lambda_t}
    \Sigmap^{-1}(\mu\!-\!\mu_t)(\mu\!-\!\mu_t)^\top\Sigmap^{-1} \\[1ex]
  &= \lambda_t \Sigmap^{-1}\Sigma_t\Sigmap^{-1} + \frac{\lambda_t}{1\!+\!\lambda_t}
    \Sigmap^{-1}(\mup\!-\!\mu_t)(\mup\!-\!\mu_t)^\top\Sigmap^{-1} \\[1ex]
  &= \lambda_t\Sigmap^{-\frac{1}{2}}\left( J_t +
    \frac{1}{1\!+\!\lambda_t}\varepsilon_t\varepsilon_t^\top\right)\Sigmap^{-\frac{1}{2}},
\end{align}
where again in the last line we have used eqs.~(\ref{eq:epsilon_t_2}) and~(\ref{eq:J_t_2}) to re-express the right side in terms of~$\varepsilon_t$ and $J_t$.
Next we substitute these limits for $U_B$ and $V_B$ into the quadratic matrix equation in eq.~(\ref{eq:quad-matrix}). It follows that
\begin{equation}
\lambda_t \Sigma_{t+1} \Sigmap^{-\frac{1}{2}}\left(J_t +
    \frac{1}{1\!+\!\lambda_t}\varepsilon_t\varepsilon_t^\top\right)\Sigmap^{-\frac{1}{2}}
    \Sigma_{t+1} + \Sigma_{t+1} =  \Sigmap^{\frac{1}{2}}\big[(1\!+\!\lambda_t) J_t\big] \Sigmap^{\frac{1}{2}}.
\label{eq:pre_J_recurse}
\end{equation}
Finally, we obtain the recursion in eq.~(\ref{eq:J_recurse_2}) by left and right
    multiplying eq.~(\ref{eq:pre_J_recurse}) by $\Sigmap^{-\frac{1}{2}}$ and
    again making the substitution $J_{t+1}=\Sigmap^{-\frac{1}{2}}\Sigma_{t+1}\Sigmap^{-\frac{1}{2}}$ from eq.~(\ref{eq:J_t_2}).
\end{proof}

The proof of convergence in future sections relies on various relaxations to derive the simple error bounds in eqs.~(\ref{eq:epsilon_converge_2}--\ref{eq:Delta_converge_2}). Before proceeding, it is therefore worth noting the following property of \Cref{alg:LS_GSM_VI} that is not apparent from these bounds. \\

\begin{fcorollary}[\textbf{One-step convergence}]
Suppose $p\!=\!\N(\mup,\Sigmap)$, and consider the limit of infinite batch size
    ($B\!\rightarrow\!\infty$) in \Cref{alg:LS_GSM_VI} followed by the
    \textit{additional} limit of no regularization ($\lambda_0\!\rightarrow\!\infty$). In this combined limit, the algorithm converges with probability 1 in one step: i.e., $\lim_{\lambda_0\rightarrow \infty}\lim_{B\rightarrow \infty}\|\varepsilon_1\| = \lim_{\lambda_0\rightarrow \infty}\lim_{B\rightarrow \infty}\|\Delta_1\|=0$.
\label{cor:one-step}
\end{fcorollary}

\begin{proof}
Consider the recursion for $J_1$ given by eq.~(\ref{eq:J_recurse_2}) in the additional limit $\lambda_0\rightarrow\infty$. In this limit one can ignore the terms that are not of leading order in $\lambda_0$, and the recursion simplifes to $J_1 J_0 J_1 \!=\! J_0$. This equation has only one positive-definite solution given by $J_1\!=\!I$. Next consider the recursion for $\varepsilon_1$ given by eq.~(\ref{eq:epsilon_recurse_2}) in the additional limit $\lambda_0\rightarrow\infty$. In this limit this recursion simplifies to
$\varepsilon_1 = (I\!-\!J_1)\varepsilon_0$, showing that $\varepsilon_1\!=\! 0$. It follows that $\Sigma_1\!=\!\Sigma$ and $\mu_1\!=\!\mu$, and future updates have no effect.
\end{proof}


\subsection{Sandwiching inequality}
\label{ssec:sandwich}

To complete the proof of convergence for \Cref{thm:converge}, we must show that $\|\varepsilon_t\|\!\rightarrow\!0$ and $\|J_t\!-\! I\|\!\rightarrow\!0$ as $t\!\rightarrow\!\infty$. We showed in Propositions \ref{prop-recursion-mean} and~\ref{prop-recursion-cov} that $\varepsilon_t$ and $J_t$ satisfy simple recursions. However, it is not immediately obvious how to translate these recursions for $\varepsilon_t$ and $J_t$ into recursions for $\|\varepsilon_t\|$ and $\|J_t\!-\! I\|$. To do so requires additional machinery.

One crucial piece of machinery is the \textit{sandwiching inequality} that we prove in this section. In addition to the normalized covariance matrices $\{J_t\}_{t=0}^\infty$, we introduce two sequences of \textit{auxiliary} matrices, $\{H_t\}_{t=1}^\infty$ and $\{K_t\}_{t=1}^\infty$ satisfying
\begin{equation}
0\prec H_{t+1}\preceq J_{t+1}\preceq K_{t+1}
\label{eq:HJK}
\end{equation}
for all $t\!\geq\!0$; this is what we call the sandwiching inequality. These auxiliary matrices are defined by the recursions
\begin{align}
\label{eq:H_recurse}
\lambda_t H_{t+1} \left(J_t + \frac{1}{1\!+\!\lambda_t} \norm{\varepsilon_t}^2 I \right) H_{t+1} + H_{t+1} =
    (1\! +\! \lambda_t) J_t, \\
\label{eq:K_recurse}
    \lambda_t K_{t+1} J_t K_{t+1} + K_{t+1} = (1\! +\! \lambda_t) J_t.
\end{align}
We invite the reader to scrutinize the differences between these recursions for $H_{t+1}$ and~$K_{t+1}$ and the one for $J_{t+1}$ eq.~(\ref{eq:J_recurse_2}). Note that in eq.~(\ref{eq:K_recurse}), defining $K_{t+1}$, we have dropped the term in eq.~(\ref{eq:J_recurse_2}) involving the outer-product $\varepsilon_t\varepsilon_t^\top$, while in eq.~(\ref{eq:H_recurse}), defining $H_{t+1}$, we have replaced this term by a scalar multiple of the identity matrix. As we show later, these auxiliary recursions are easier to analyze because the matrices $H_{t+1}$ and $K_{t+1}$ (unlike $J_{t+1}$) share the same eigenvectors as~$J_t$. Later we will exploit this fact to bound their eigenvalues as well as the errors $\|J_{t+1}\!-\! I\|$.

In this section we show that the recursions for $H_{t+1}$ and $K_{t+1}$ in eqs.~(\ref{eq:H_recurse}--\ref{eq:K_recurse}) imply the sandwiching inequality in eq.~(\ref{eq:HJK}). As we shall see, the sandwiching inequality follows mainly from the monotonicity property of these quadratic matrix equations proven in \Cref{lemma-ABCXY}. \\

\begin{fproposition}[\textbf{Sandwiching inequality}]
Let $\Sigma_0\!\succ\! 0$ and $\lambda_t\!>\!0$ for all $t\!\geq\! 0$. Also,~let $\{\varepsilon_t\}_{t=1}^\infty$, $\{J_t\}_{t=1}^\infty$, $\{H_t\}_{t=1}^\infty$, and $\{K_t\}_{t=1}^\infty$ be defined, respectively, by the recursions in eqs.~(\ref{eq:epsilon_recurse_2}), (\ref{eq:J_recurse_2}), and (\ref{eq:H_recurse}--\ref{eq:K_recurse}). Then for all $t\!\geq\! 0$ we have
\begin{equation}
\label{eq:sandwich-ineq}
0\prec H_{t+1}\preceq J_{t+1}\preceq K_{t+1}.
\end{equation}
\end{fproposition}

\begin{proof}
We prove the orderings in the proposition from left to right. Since $\Sigma_0\!\succ\! 0$, it follows from eq.~(\ref{eq:J_t_2}) that $J_0\!\succ\! 0$, and \Cref{lemma-qsol-psd} ensures for the recursion in eq.~(\ref{eq:J_recurse_2}) that $J_{t+1}\!\succ\! 0$ for all $t\!\geq\! 0$.
Likewise, since $J_t\succ 0$ for all $t\geq 0$, \Cref{lemma-qsol-psd} ensures for the recursion in eq.~(\ref{eq:H_recurse}) that $H_{t+1}\!\succ\! 0$ for all $t\!\geq\!0$. This proves the first ordering in the proposition. To prove the remaining orderings, we note that for all vectors $\varepsilon_t$,
\begin{equation}
\lambda_t J_t\, \preceq\, \lambda_t \left(J_t + \frac{1}{1\!+\!\lambda_t}\varepsilon_t\varepsilon_t^\top\right)\, \preceq\,
  \lambda_t \left(J_t + \frac{1}{1\!+\!\lambda_t}\|\varepsilon_t\|^2 I \right).
\label{eq:orderings}
\end{equation}
We now apply \Cref{lemma-ABCXY} to the quadratic matrix equations that define
the recursions for $H_{t+1}$, $J_{t+1}$, and $K_{t+1}$.
From the first ordering in eq.~(\ref{eq:orderings}), and for the recursions for $J_{t+1}$ and $K_{t+1}$ in eqs.~(\ref{eq:J_recurse_2}) and (\ref{eq:K_recurse}), \Cref{lemma-ABCXY} ensures that $J_{t+1}\! \preceq\! K_{t+1}$. Likewise, from the second ordering in eq.~(\ref{eq:orderings}), and for the recursions for $J_{t+1}$ and $H_{t+1}$ in eqs.~(\ref{eq:J_recurse_2}) and (\ref{eq:H_recurse}), \Cref{lemma-ABCXY} ensures that $H_{t+1}\! \preceq\! J_{t+1}$.
\end{proof}


\subsection{Eigenvalue bounds}
\label{ssec:eig-bounds}

The sandwiching inequality in the previous section provides a powerful tool for analyzing the eigenvalues of the normalized covariance matrices $\{J_t\}_{t=1}^{\infty}$. As shown in the following lemma, much of this power lies in the fact that the matrices $J_t$, $H_{t+1}$, and~$K_{t+1}$ are jointly diagonalizable. \\

\begin{flemma}[\textbf{Joint diagonalizability}]
\label{lemma:joint}
Let $\lambda_t\!>\!0$ for all $t\!\geq\! 0$, and let $\{\varepsilon_t\}_{t=1}^\infty$, $\{J_t\}_{t=1}^\infty$, $\{K_t\}_{t=1}^\infty$, and $\{H_t\}_{t=1}^\infty$ be defined, respectively, by the recursions in eqs.~(\ref{eq:epsilon_recurse_2}), (\ref{eq:J_recurse_2}), and (\ref{eq:H_recurse}--\ref{eq:K_recurse}). Then for all $t\!\geq\! 0$ we have the following:
\begin{itemize}
\item[(i)] $H_{t+1}$ and $K_{t+1}$ share the same eigenvectors as $J_t$.
\item[(ii)] Each eigenvalue $\nu_J$ of $J_t$ determines a corresponding eigenvalue $\nu_H$ of $H_{t+1}$ and a corresponding eigenvalue $\nu_K$ of~$K_{t+1}$ via the \textit{positive} roots of the quadratic equations
\begin{align}
\lambda_t  \left(\nu_J + \frac{\|\varepsilon_t\|^2}{1\!+\!\lambda_t}\right) \nu_H^2 + \nu_H &= (1\!+\!\lambda_t) \nu_J, \label{eq:eigJH} \\
\lambda_t  \nu_J \nu_K^2 + \nu_K &= (1\!+\!\lambda_t) \nu_J. \label{eq:eigJK}
\end{align}
\end{itemize}
\end{flemma}

\begin{proof}
Write $J_t = Q \Lambda_J Q^\top$, where $Q$ is the orthogonal matrix storing the eigenvectors of $J_t$ and $\Lambda_J$ is the \textit{diagonal} matrix storing its eigenvalues. Now define the matrices
\begin{align}
\Lambda_H &= Q^\top H_{t+1} Q, \\
\Lambda_K &= Q^\top K_{t+1} Q.
\end{align}
We will prove that $J_t$, $H_{t+1}$, and $K_{t+1}$ share the same eigenvectors as $J_t$ by showing that the matrices $\Lambda_H$ and $\Lambda_K$ are also diagonal. We start by multiplying eqs.~(\ref{eq:H_recurse}--\ref{eq:K_recurse}) on the left by $Q^\top$ and on the right by $Q$. In this way we find
\begin{align}
\label{eq:H_diag_recurse}
\lambda_t \Lambda_H \left(\Lambda_J + \frac{1}{1\!+\!\lambda_t} \norm{\varepsilon_t}^2 I \right) \Lambda_H + \Lambda_H = (1\! +\! \lambda_t) \Lambda_J, \\
\label{eq:K_diag_recurse}
    \lambda_t \Lambda_K \Lambda_J \Lambda_K + \Lambda_K = (1\! +\! \lambda_t) \Lambda_J.
\end{align}
Since $\Lambda_J$ is diagonal, we see from eqs.~(\ref{eq:H_diag_recurse}--\ref{eq:K_diag_recurse})
that $\Lambda_H$ and $\Lambda_K$ also have purely diagonal solutions; this proves the first claim of the lemma. We obtain the scalar equations in eqs.~(\ref{eq:eigJH}--\ref{eq:eigJK}) by focusing on the corresponding diagonal elements (i.e., eigenvalues) of the matrices $\Lambda_H$, $\Lambda_J$, and $\Lambda_K$ in eqs.~(\ref{eq:H_diag_recurse}--\ref{eq:K_diag_recurse}); this proves the second claim of the lemma.
\end{proof}

To prove the convergence of Algorithm~\ref{alg:LS_GSM_VI}, we will also need upper and lower bounds on eigenvalues of the normalized covariance matrices. The next lemma provides these bounds. \\

\begin{flemma}[\textbf{Bounds on eigenvalues of $J_{t+1}$}]
Let $\lambda_t\!>\!0$ for all $t\!\geq\! 0$, and let $\{\varepsilon_t\}_{t=1}^\infty$, $\{J_t\}_{t=1}^\infty$, $\{K_t\}_{t=1}^\infty$, and $\{H_t\}_{t=1}^\infty$ be defined, respectively, by the recursions in eqs.~(\ref{eq:epsilon_recurse_2}), (\ref{eq:J_recurse_2}), and (\ref{eq:H_recurse}--\ref{eq:K_recurse}). Then for all $t\!\geq 0$, the largest and smallest eigenvalues of $J_{t+1}$ satisfy
\begin{align}
\label{eq:maxJ}
\nu_{\text{max}}(J_{t+1}) & \,\leq\, 
\sqrt{\frac{1+\lambda_t}{\lambda_t}}, \\[0.5ex]
\label{eq:minJ}
\nu_{\text{min}}(J_{t+1}) & \, \geq\, 
\min\left(
  \nu_{\text{min}}(J_t),\, \frac{1+\lambda_t}{1+\lambda_t+\|\varepsilon_t\|^2}\right).
\end{align}
\label{lemma:eigsJ}
\end{flemma}

\begin{proof}
We will prove these bounds using the sandwiching inequality. We start by proving an upper bound on $\nu_{\text{max}}(K_{t+1})$. Recall from \Cref{lemma:joint} that each eigenvalue $\nu_K$ of $K_{t+1}$ is determined by a corresponding eigenvalue $\nu_J$ of $J_t$ via the positive root of the quadratic equation in eq.~(\ref{eq:eigJK}).
Rewriting this equation, we see that
\begin{equation}
\nu_K^2\, =\, \frac{1\!+\!\lambda_t}{\lambda_t} - \frac{\nu_K}{\lambda_t\nu_J}\, \leq\,  \frac{1\!+\!\lambda_t}{\lambda_t},
\end{equation}
showing that every eigenvalue of $K_{t+1}$ must be less than $\sqrt{\frac{1+\lambda_t}{\lambda_t}}$. Now from the sandwiching inequality, we know that $J_{t+1}\!\preceq\! K_{t+1}$, from which it follows that $\nu_{\text{max}}(J_{t+1}) \leq \nu_{\text{max}}(K_{t+1})$. Combining these observations, we have shown
\begin{equation}
\nu_{\text{max}}(J_{t+1})\, \leq\, \nu_{\text{max}}(K_{t+1})\, \leq\, \sqrt{\frac{1\!+\!\lambda_t}{\lambda_t}},
\end{equation}
which proves the first claim of the lemma. Next we prove a lower bound on $\nu_{\text{min}}(H_{t+1})$. Again, recall from \Cref{lemma:joint} that each eigenvalue $\nu_H$ of $H_{t+1}$ is determined by a corresponding eigenvalue $\nu_J$ of $J_t$ via the positive root of the quadratic equation in eq.~(\ref{eq:eigJH}).  We restate this equation here for convenience:
\begin{displaymath}
\lambda_t  \left(\nu_J + \frac{\|\varepsilon_t\|^2}{1\!+\!\lambda_t}\right) \nu_H^2 + \nu_H = (1\!+\!\lambda_t) \nu_J
\end{displaymath}
We now exploit two key properties of this equation, both of which are proven in \Cref{lemma-f-mono}. Specifically, \Cref{lemma-f-mono} states that if $\nu_H$ is computed from the positive root of this equation, then $\nu_H$ is
a \textit{monotonically increasing function} of $\nu_J$, and it also satisfies the \textit{lower bound}
\begin{equation}
\nu_H \geq \min\left(\nu_J, \frac{1\!+\!\lambda_t}{1\!+\!\lambda_t\!+\!\|\varepsilon_t\|^2}\right).
\end{equation}
We can combine these properties to derive a lower bound on the smallest eigenvalue of $H_{t+1}$; namely, it must be the case that
\begin{equation}
\nu_{\text{min}}(H_{t+1}) \geq \min\left(\nu_\text{min}(J_t), \frac{1\!+\!\lambda_t}{1\!+\!\lambda_t\!+\!\|\varepsilon_t\|^2}\right).
\label{eq:eig-min-H}
\end{equation}
Now again from the sandwiching inequality, we know that $J_{t+1}\!\succeq\! H_{t+1}$, from which it follows that $\nu_{\text{min}}(J_{t+1}) \geq \nu_{\text{min}}(H_{t+1})$. Combining this observation with eq.~(\ref{eq:eig-min-H}), we see that
\begin{equation}
\nu_{\text{min}}(J_{t+1}) \geq \nu_{\text{min}}(H_{t+1}) \geq \min\left(\nu_\text{min}(J_t), \frac{1\!+\!\lambda_t}{1\!+\!\lambda_t\!+\!\|\varepsilon_t\|^2}\right),
\label{eq:eig-min-J}
\end{equation}
which proves the second claim of the lemma.
\end{proof}


\subsection{Recursions for $\|\varepsilon_t\|$ and $\|\Delta_t\|$}
\label{ssec:errors-recurse}

In this section, we analyze how the errors $\|\varepsilon_t\|$ and~$\|\Delta_t\|$ evolve from one iteration of \Cref{alg:LS_GSM_VI} to the next. These \textit{per-iteration} results are the cornerstone of the proof of convergence in the infinite batch limit. \\

\begin{fproposition}[\textbf{Decay of $\|\varepsilon_t\|$}]
\label{prop-mean-contraction}
Suppose that $p = \N(\mup,\Sigmap)$.
Then for \Cref{alg:LS_GSM_VI} in the limit of infinite batch size ($B\!\rightarrow\!\infty$), the normalized errors in eq.~(\ref{eq:epsilon_t_2}) of the variational mean \textit{strictly decrease} from one iteration to the next: i.e., $\|\varepsilon_{t+1}\| < \|\varepsilon_t\|$. More precisely, they satisfy
\begin{equation}
\norm{\varepsilon_{t+1}}
    \leq \left(1 - \frac{\lambda_t}{1\! +\! \lambda_t}
    \nu_\text{min}(J_{t+1}) \right) \norm{\varepsilon_t},
\label{eq:prop-mean-contraction}
\end{equation}
where the multiplier in parentheses on the right side is strictly less than one.
\end{fproposition}

\begin{proof}
Recall from \Cref{prop-recursion-mean} that the normalized errors in the variational mean satisfy the recursion
\begin{equation}
   \varepsilon_{t+1} = \left[I - \frac{\lambda_t}{1\!+\!\lambda_t} J_t\right] \varepsilon_t.
\end{equation}
Taking norms and applying the sub-multiplicative property of the spectral norm, we have
 \begin{equation}
  \label{eq-epsilon-subm}
      \norm{\varepsilon_{t+1}} \leq \left\| I - \frac{\lambda_t}{1\!+\!\lambda_t} J_{t+1}\right\|\norm{\varepsilon_t}.
 \end{equation}
Consider the matrix norm that appears on the right side of eq.~(\ref{eq-epsilon-subm}). By \Cref{lemma:eigsJ}, and specifically eq.~(\ref{eq:maxJ}) which gives the ordering $J_{t+1}\preceq\sqrt{\frac{1+\lambda_t}{\lambda_t}}I$, it follows that
\begin{equation}
I - \frac{\lambda_t}{1\!+\!\lambda_t} J_{t+1} \succeq  \left(1 - \sqrt{\frac{\lambda_t}{1\!+\!\lambda_t}}\,\right)I \succ 0.
\end{equation}
Thus the spectral norm of this matrix is strictly greater than zero and determined by the minimum eigenvalue of $J_{t+1}$.  In particular, we have
\begin{equation}
 \left\|I - \frac{\lambda_t}{1\!+\!\lambda_t} J_t\right\| = 1-\frac{\lambda_t}{1\!+\!\lambda_t}\nu_{\text{min}}(J_{t+1}),
 \label{eq:minEigJdecay}
\end{equation}
and the proposition is proved by substituting eq.~(\ref{eq:minEigJdecay}) into eq.~(\ref{eq-epsilon-subm}).
\end{proof}


\begin{fproposition}[\textbf{Decay of $\|\Delta_t\|$}]
\label{prop-cov-contraction}
Suppose that $p = \N(\mup,\Sigmap)$.
Then for \Cref{alg:LS_GSM_VI} in the limit of infinite batch size ($B\!\rightarrow\!\infty$), the normalized errors in eq.~(\ref{eq:Delta_t_2}) of the variational covariance satisfy
\begin{equation}
\norm{\Delta_{t+1}}\,
    \leq\, \norm{\varepsilon_t}^2 + \frac{1}{1\!+\!\lambda_t\nu_{\text{min}}(J_t)}\norm{\Delta_t}.
\label{eq:prop-cov-contraction}
\end{equation}
\end{fproposition}

\begin{proof}
We start by applying the triangle inequality and the sandwiching inequality:
\begin{align}
\|\Delta_{t+1}\|
  &= \|J_{t+1}\!-\!I\|, \\
  &\leq \|J_{t+1}\!-\!K_{t+1}\|\, +\, \|K_{t+1}\!-\!I\|, \\
  &\leq \|H_{t+1}\!-\!K_{t+1}\|\, +\, \|K_{t+1}\!-\!I\|. \label{eq:tri_sand}
\end{align}
Already from these inequalities we can see the main outlines of the result in eq.~(\ref{eq:prop-cov-contraction}). Clearly, the first term in eq.~(\ref{eq:tri_sand}) must vanish when $\|\varepsilon_t\|\!=\!0$ because the auxiliary matrices~$H_{t+1}$ and~$K_{t+1}$, defined in eqs.~(\ref{eq:H_recurse}--\ref{eq:K_recurse}), are equal when $\varepsilon_t\!=\!0$. Likewise, the second term in eq.~(\ref{eq:tri_sand}) must vanish when $\|\Delta_t\|\!=\!0$, or equivalently when $J_t\!=\!I$, because in this case eq.~(\ref{eq:K_recurse}) is also solved by $K_{t+1}\!=\!I$.

First we consider the left term in eq.~(\ref{eq:tri_sand}). Recall from \Cref{lemma:joint} that the matrices $H_{t+1}$ and~$K_{t+1}$ share the same eigenvectors; thus the spectral norm $\|H_{t+1}\!-\!K_{t+1}\|$ is equal to the largest gap between their corresponding eigenvalues. Also recall from eqs.~(\ref{eq:eigJH}--\ref{eq:eigJK}) of \Cref{lemma:joint} that these corresponding eigenvalues $\nu_H$ and $\nu_K$ are determined by the positive roots of the quadratic equations
\begin{align}
\lambda_t  \left(\nu_J + \frac{\|\varepsilon_t\|^2}{1\!+\!\lambda_t}\right) \nu_H^2 + \nu_H &= (1\!+\!\lambda_t) \nu_J, \label{eq:nuJnuH} \\
\lambda_t  \nu_J \nu_K^2 + \nu_K &= (1\!+\!\lambda_t) \nu_J,
\label{eq:nuJnuK}
\end{align}
where $\nu_J$ is their (jointly) corresponding eigenvalue of $J_{t}$. Since these two equations agree when $\|\varepsilon_t\|^2\!=\!0$, it is clear that $|\nu_H\!-\!\nu_K|\rightarrow 0$ as $\|\varepsilon_t\|\rightarrow 0$. More precisely, as we show in \Cref{lemma-g-mono} of section \ref{ssec:supporting-lemmas}, it is the case that
\begin{equation}
|\nu_H\!-\!\nu_K| \leq \|\varepsilon_t\|^2.
\end{equation}
(Specifically, this is property (v) of \Cref{lemma-g-mono}.) It follows in turn from this property that
\begin{equation}
\|H_{t+1}\!-\!K_{t+1}\|\, \leq\, \|\varepsilon_t\|^2.
\label{eq:HKnorm}
\end{equation}
We have thus bounded the left term in eq.~(\ref{eq:tri_sand}) by a quantity that, via \Cref{prop-mean-contraction}, is decaying geometrically to zero with the number of iterations of the algorithm.

Next we focus on the right term in eq.~(\ref{eq:tri_sand}). The spectral norm $\|K_{t+1}\!-\!I\|$ is equal to the largest gap between any eigenvalue of $K_{t+1}$ and the value of 1 (i.e., the value of all eigenvalues of $I$). Recall from eq.~(\ref{eq:eigJK}) of \Cref{lemma:joint} that each eigenvalue~$\nu_J$ of $J_{t}$ determines a corresponding eigenvalue $\nu_K$ of $K_{t+1}$ via the positive root of the quadratic equation
\begin{equation}
\lambda_t\nu_J\nu_K^2 + \nu_K = (1\!+\!\lambda_t)\nu_J.
\label{eq:nuJnuK2}
\end{equation}
This correspondence has an important \textit{contracting} property that eigenvalues of $J_t$ not equal to one are mapped to eigenvalues of $K_{t+1}$ that are closer to one. In particular, as we show in \Cref{lemma-f-mono} of section \ref{ssec:supporting-lemmas}, it is the case that
\begin{equation}
|\nu_K\!-\!1| \leq \frac{1}{1\!+\!\lambda_t\nu_J} |\nu_J\!-\!1|.
\end{equation}
(Specifically, this is property (vii) of \Cref{lemma-f-mono}.) It follows in turn from this property that
\begin{equation}
\|K_{t+1}\!-\!I\|\, \leq\, \frac{1}{1\!+\!\lambda_t\nu_{\text{min}}(J_t)}\|J_t\!-\!I\|.
\label{eq:KIJI}
\end{equation}
Finally, the proposition is proved by substituting eq.~(\ref{eq:HKnorm}) and eq.~(\ref{eq:KIJI}) into eq.~(\ref{eq:tri_sand}).
\end{proof}


The results of \Cref{prop-mean-contraction} and \Cref{prop-cov-contraction} could be used to further analyze the convergence of \Cref{alg:LS_GSM_VI} when different levels of regularization~$\lambda_t$ are used at each iteration. By specializing to a fixed level of regularization, however, we obtain the especially interpretable results of eqs.~(\ref{eq:epsilon_contract}--\ref{eq:Delta_contract}) in the proof sketch of \Cref{thm:converge}. To prove these results, we need one further lemma. \\

\begin{flemma}[\textbf{Bound on $\nu_\text{min}(J_t)$}]
Suppose that $p = {\N}(\mup,\Sigmap)$ in \Cref{alg:LS_GSM_VI},
    and let $\alpha\!>\!0$ denote the minimum eigenvalue of the matrix
    $\mat{\Sigmap}^{-\frac{1}{2}}\mat{\Sigma}_0\mat{\Sigmap}^{-\frac{1}{2}}$. Then in the limit of infinite batch size ($B\!\rightarrow\!\infty$), and for any fixed level of regularization $\lambda\!>\!0$, we have for all $t\geq 0$ that
\begin{equation}
\nu_{\text{min}}(J_t) \geq \min\left(\alpha,\frac{1\!+\!\lambda}{1\!+\!\lambda\!+\|\varepsilon_0\|^2}\right).
\label{eq:minEigJt}
\end{equation}
\end{flemma}

\begin{proof}
We prove the result by induction.
Note that
    $\nu_{\text{min}}(J_0)\!=\!\nu_{\text{min}}\Big(\mat{\Sigmap}^{-\frac{1}{2}}\mat{\Sigma}_0\mat{\Sigmap}^{-\frac{1}{2}}\Big)\!=\!\alpha$,
    so that eq.~(\ref{eq:minEigJt}) holds for $t\!=\!0$. Now assume that the result holds for some iteration $t\!>\!0$. Then
\begin{align}
\nu_{\text{min}}(J_{t+1})
   &\geq \min\left(\nu_{\text{min}}(J_{t}), \frac{1\!+\!\lambda}{1\!+\!\lambda\!+\|\varepsilon_t\|^2}\right), \\
   &\geq \min\left(\min\left(\alpha,\frac{1\!+\!\lambda}{1\!+\!\lambda\!+\|\varepsilon_0\|^2}\right),
     \frac{1\!+\!\lambda}{1\!+\!\lambda\!+\|\varepsilon_t\|^2}\right), \\
   &= \min\left(\alpha,\frac{1\!+\!\lambda}{1\!+\!\lambda\!+\|\varepsilon_0\|^2}\right),
 \end{align}
where the first inequality is given by eq.~(\ref{eq:minJ}) of~\Cref{lemma:eigsJ}, the second inequality follows from the inductive hypothesis, and the final equality holds because $\|\varepsilon_t\|\!<\!\|\varepsilon_0\|$ from \Cref{prop-mean-contraction}.
\end{proof}

Note how the bound in eq.~(\ref{eq:minEigJt}) depends on $\alpha$ and $\|\varepsilon_0\|$, both of which reflect the quality of initialization. In particular, when $\alpha\ll 1$, the initial covariance is close to singular, and when $\|\varepsilon_0\|$ is large, the initial mean is a poor estimate. Both these qualities of initialization play a role in the next result. \\


\begin{fcorollary}[\textbf{Rates of decay for $\|\varepsilon_t\|$ and $\|\Delta_t\|$}]
Suppose that $p = \N(\mup,\Sigmap)$ and let $\alpha\!>\!0$ denote the minimum
    eigenvalue of the matrix $\mat{\Sigmap}^{-\frac{1}{2}}\mat{\Sigma}_0\mat{\Sigmap}^{-\frac{1}{2}}$. Also, for any fixed level of regularization $\lambda\!>\!0$, define
\begin{align}
\beta &= \min\left(\alpha,\frac{1\!+\!\lambda}{1\!+\!\lambda\!+\!\|\varepsilon_0\|^2}\right), \\
\delta &= \frac{\lambda\beta}{1\!+\!\lambda},
\end{align}
where $\beta\in(0,1]$ measures the quality of initialization and $\delta\in(0,1)$ measures a rate of decay. Then in the limit of infinite batch size ($B\!\rightarrow\!\infty$), the normalized errors in eqs.~(\ref{eq:epsilon_t_2}--\ref{eq:Delta_t_2}) satisfy
\begin{align}
\|\varepsilon_{t+1}\|^2 &\leq (1\!-\!\delta)^2 \|\vec{\varepsilon}_t\|^2,
\label{eq:epsilon_contract_2} \\
\|\Delta_{t+1}\| &\leq (1\!-\!\delta) \|\Delta_t\| +  \|\varepsilon_t\|^2.
\label{eq:Delta_contract_2}
\end{align}
\label{cor:contract}
\vspace{-1ex}
\end{fcorollary}

\begin{proof}
The results follow from the previous ones in this section. In particular, from \Cref{prop-mean-contraction} and the previous lemma, we see that
\begin{equation}
\norm{\varepsilon_{t+1}}
    \leq \left(1 - \frac{\lambda}{1\! +\! \lambda}
    \nu_\text{min}(J_{t+1}) \right) \norm{\varepsilon_t}
    \leq \left(1 - \frac{\lambda\beta}{1\!+\!\lambda}\right)\|\varepsilon_t\|
     = (1\!-\!\delta) \|\varepsilon_t\|.
\end{equation}
Likewise, from \Cref{prop-cov-contraction} and the previous lemma, we see that
\begin{align}
\norm{\Delta_{t+1}}
   &\leq \norm{\varepsilon_t}^2 + \frac{1}{1\!+\!\lambda\nu_{\text{min}}(J_t)}\norm{\Delta_t}, \\[1ex]
   &\leq \norm{\varepsilon_t}^2 + \frac{1}{1\!+\!\lambda\beta}\norm{\Delta_t}, \\[1ex]
   &= \norm{\varepsilon_t}^2 + \left(1-\frac{\lambda\beta}{1\!+\!\lambda\beta}\right)\norm{\Delta_t}, \\[1ex]
   &\leq \norm{\varepsilon_t}^2 + \left(1-\frac{\lambda\beta}{1\!+\!\lambda}\right)\norm{\Delta_t}, \\[1ex]
   &= \norm{\varepsilon_t}^2 + (1-\delta)\norm{\Delta_t}.
\end{align}
\end{proof}


\subsection{Induction}
\label{ssec:induction}

From the previous corollary we can at last give a simple proof of \Cref{thm:converge}. It should also be clear that tighter bounds can be derived, and differing levels of regularization accommodated, if we instead proceed from the more general bounds in Propositions~\ref{prop-mean-contraction} and~\ref{prop-cov-contraction}.

\begin{proof}[Proof of \Cref{thm:converge}]
We start from eqs.~(\ref{eq:epsilon_contract_2}--\ref{eq:Delta_contract_2}) of \Cref{cor:contract} and proceed by induction. At iteration $t\!=\!0$, we see from these equations that
\begin{align}
\|\varepsilon_{1}\| &\leq (1\!-\!\delta) \|\vec{\varepsilon}_0\|, \\
\|\Delta_{1}\| &\leq (1\!-\!\delta) \|\Delta_0\| +  \|\varepsilon_0\|^2.
\end{align}
The above agree with eqs.~(\ref{eq:epsilon_converge}--\ref{eq:Delta_converge}) at iteration $t\!=\!0$ and therefore establish the base case of the induction. Next we assume the inductive hypothesis that eqs.~(\ref{eq:epsilon_converge}--\ref{eq:Delta_converge}) are true at some iteration $t\!-\!1$. Then again, appealing to eqs.~(\ref{eq:epsilon_contract_2}--\ref{eq:Delta_contract_2}) of \Cref{cor:contract}, we see that
\begin{align}
\|\varepsilon_{t}\|
  &\leq (1\!-\!\delta) \|\vec{\varepsilon}_{t-1}\|, \\
  &\leq (1\!-\!\delta) (1\!-\!\delta)^{t-1} \|\vec{\varepsilon}_0\|, \\
  &= (1\!-\!\delta)^{t} \|\vec{\varepsilon}_0\|, \\[2ex]
\|\Delta_t\|
  &\leq (1\!-\!\delta) \|\Delta_{t-1}\| +  \|\varepsilon_{t-1}\|^2, \\
  &\leq (1\!-\!\delta) \Big[(1\!-\!\delta)^{t-1}\|\Delta_0\| + (t\!-\!1)(1\!-\!\delta)^{t-2}\|\varepsilon_0\|^2\Big]
       +  (1\!-\!\delta)^{2(t-1)}\|\varepsilon_0\|^2, \\
  &= (1\!-\!\delta)^t \|\Delta_0\| + \Big[(t\!-\!1)(1\!-\!\delta)^{t-1} + (1\!-\!\delta)^{2t-2}\Big]\|\varepsilon_0\|^2, \\
  &\leq (1\!-\!\delta)^t \|\Delta_0\| + \Big[(t\!-\!1)(1\!-\!\delta)^{t-1} + (1\!-\!\delta)^{t-1}\Big]\|
\varepsilon_0\|^2, \\
  &= (1\!-\!\delta)^t \|\Delta_0\| + t(1\!-\!\delta)^{t-1}\| \varepsilon_0\|^2.
\end{align}
This proves the theorem.
\end{proof}


\subsection{Supporting lemmas}
\label{ssec:supporting-lemmas}

In this section we collect a number of lemmas whose results are needed throughout this appendix but whose proofs digress from the main flow of the argument. \\

\begin{figure}[t]
\centerline{\includegraphics[width=0.66\textwidth]{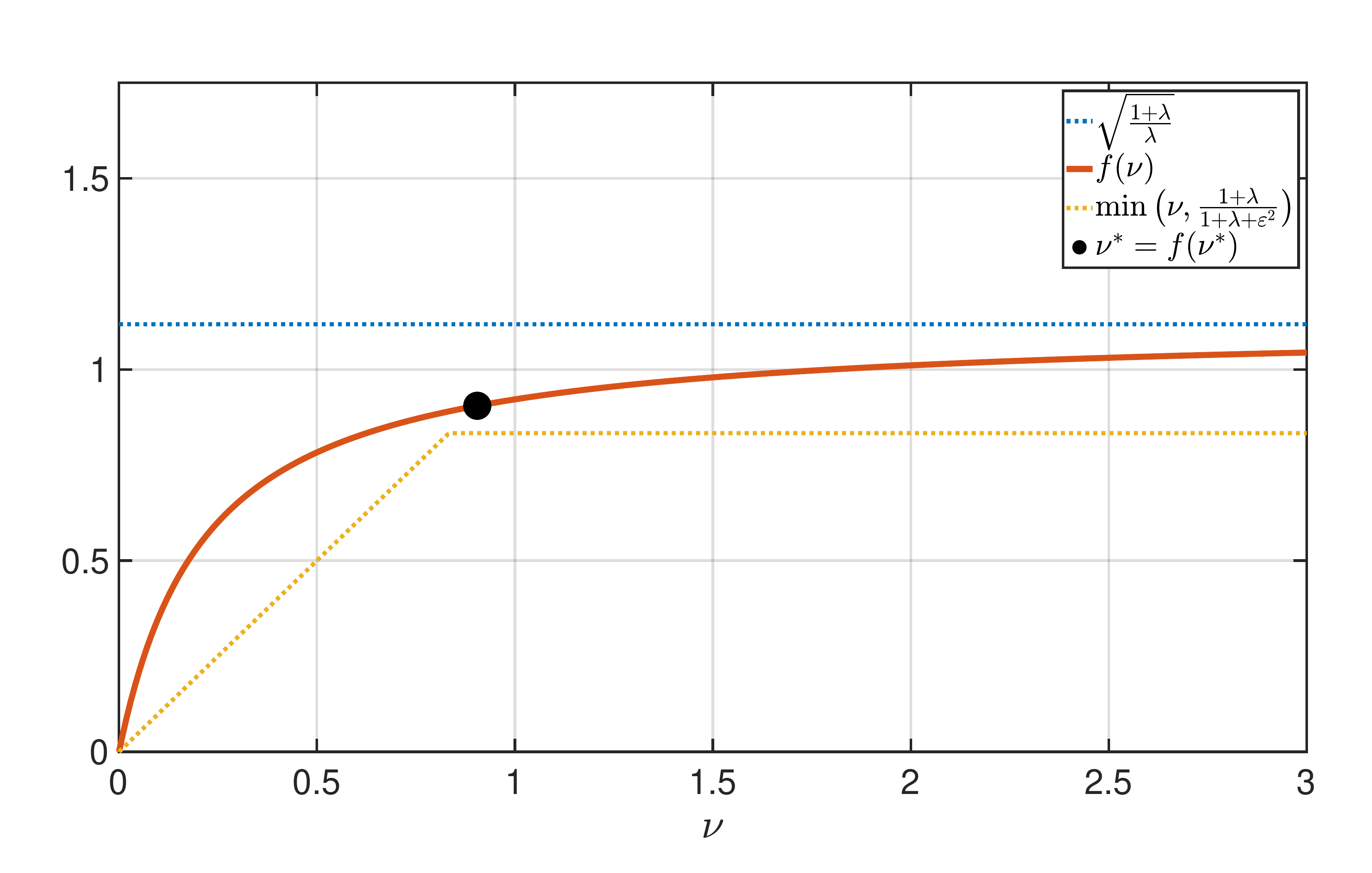}}
\caption{Plot of the function $f$ in eq.~(\ref{eq:f_nu}), as well as its fixed point and upper and lower bounds from \Cref{lemma-f-mono}, with $\lambda\!=\!4$ and $\varepsilon^2\!=\!1$.}
\label{fig:plot_f}
\end{figure}

\begin{flemma}
\label{lemma-f-mono}
Let $\lambda\!>\!0$ and $\varepsilon^2\geq0$, and let $f:\mathbb{R}_+\rightarrow\mathbb{R}_+$ be the function defined implicitly as follows: if $\nu\!>\!0$ and $\xi\!=\!f(\nu)$, then $\xi$ is equal to the \textit{positive} root of the quadratic equation
\begin{equation}
\lambda\left(\nu + \frac{\varepsilon^2}{1\!+\!\lambda}\right) \xi^2  + \xi - (1\!+\!\lambda)\nu\, =\, 0.
\label{eq:f_implicit}
\end{equation}
Then $f$ has the following properties:
\begin{itemize}[noitemsep]
\item[(i)] $f$ is monotonically increasing on $(0,\infty)$.
\item[(ii)] $f(\nu)\!<\!\sqrt{\frac{1+\lambda}{\lambda}}$ for all $\nu\!>\! 0$.
\item[(iii)] $f$ has a unique fixed point $\nu^* = f(\nu^*)$.\\[-2.5ex]
\item[(iv)] $f(\nu)\!\geq\!\nu^*$ for all $\nu\!\geq\!\nu^*$.
\item[(v)] $f(\nu)\!>\!\nu$ for all $\nu\in(0,\nu^*)$.
\item[(vi)] $f(\nu)\! \geq\! \min\left(\nu, \frac{1+\lambda}{1 + \lambda + \varepsilon^2}\right)$ for all $\nu\!>\!0$.
\item[(vii)] If $\varepsilon^2\!=\!0$, then $|\nu\!-\!1| \geq (1\!+\!\lambda\nu)|f(\nu)\!-\!1|$ for all $\nu\!>\!0$.
\end{itemize}
\end{flemma}
\vspace{5pt}
%
Before proving the lemma, we note that it is straightforward to solve the quadratic equation in eq.~(\ref{eq:f_implicit}). Doing so, we find
\begin{equation}
f(\nu) = \frac{-1+\sqrt{1+4\lambda(1+\lambda)\nu^2 + 4\lambda\varepsilon^2\nu}}
  {2\lambda\left(\nu + \frac{\varepsilon^2}{1+\lambda}\right)}.
\label{eq:f_nu}
\end{equation}
In most aspects, this explicit form for $f$ is less useful than the implicit one given in the statement of the lemma. However, eq.~(\ref{eq:f_nu}) is useful for visualizing properties (i)-(vi), and Fig.~\ref{fig:plot_f} shows a plot of $f(\nu)$ with $\lambda\!=\!4$ and $\varepsilon^2\!=\!1$. We now prove the lemma.

\begin{proof}
Let $\nu\!>\!0$. To prove property (i) that $f$ is monotonically increasing, it suffices to show $f'(\nu)\!>\!0$. Differentiating eq.~(\ref{eq:f_implicit}) with respect to $\nu$, we find that
\begin{equation}
\lambda \xi^2 + 2\lambda\left(\nu + \frac{\varepsilon^2}{1\!+\!\lambda}\right) \xi f'(\nu) + f'(\nu) - (1\!+\!\lambda) = 0,
\end{equation}
where $\xi\!=\!f(\nu)$. To proceed, we re-arrange terms to isolate $f'(\nu)$ on the left side and use eq.~(\ref{eq:f_implicit}) to remove quadratic powers of $\xi$. In this way, we find:
\begin{align}
\left[1 + 2\lambda\left(\nu + \frac{\varepsilon^2}{1\!+\!\lambda}\right) \xi\right]f'(\nu)
  &= 1 + \lambda - \lambda \xi^2, \\[-1ex]
  &= 1 + \lambda - \frac{(1+\lambda)\nu-\xi}{\nu + \frac{\varepsilon^2}{1\!+\!\lambda}}, \\[-1ex]
  &= \frac{\xi+\varepsilon^2}{\nu + \frac{\varepsilon^2}{1\!+\!\lambda}}.
\end{align}
Note that the term in brackets on the left side is strictly positive, as is the term on the right side. It follows that $f'(\nu)\!>\!0$, thus proving property (i). Moreover, since $f$ is monotonically increasing, it follows from eq.~(\ref{eq:f_nu}) that
\begin{equation}
f(\nu)\, <\, \lim_{\omega\rightarrow\infty} f(\omega)\, =\, \sqrt{\frac{1+\lambda}{\lambda}},
\end{equation}
thus proving property (ii). To prove property (iii), we solve for fixed points of~$f$. Let $\nu^*\!>\!0$ denote a fixed point satisfying $\nu^*\! =\! f(\nu^*)$. Then upon setting $\nu\!=\!\nu^*$ in eq.~(\ref{eq:f_implicit}), we must find that $\xi\!=\!\nu^*$ is a solution of the resulting equation, or
\begin{equation}
\lambda\left(\nu^* + \frac{\varepsilon^2}{1\!+\!\lambda}\right) {\nu^*}^2  + \nu^* - (1\!+\!\lambda)\nu^*\, =\, 0.
\label{eq:fixed_point}
\end{equation}
Eq.~(\ref{eq:fixed_point}) has one root at zero, one negative root, and one positive root, but only the last of these can be a fixed point of $f$, which is defined over $\mathbb{R}_+$. This fixed point corresponds to the positive root of the quadratic equation:
\begin{equation}
\left(\nu^* + \frac{\varepsilon^2}{1\!+\!\lambda}\right) {\nu^*}\, =\, 1.
\label{eq:fixed_point_quad}
\end{equation}
This proves property (iii). Property (iv) follows easily from properties (i) and (iii): if $\nu\!\geq\!\nu^*$, then $f(\nu)\!\geq\! f(\nu^*)\! = \nu^*$, where the inequality holds because $f$ is monotonically increasing and the equality holds because $\nu^*$ is a fixed point of $f$. To prove property (v), suppose that $\nu\in(0,\nu^*)$. Then from eq.~(\ref{eq:fixed_point_quad}), it follows that
\begin{equation}
\left(\nu + \frac{\varepsilon^2}{1\!+\!\lambda}\right) {\nu}\, <\, 1.
\label{eq:nu_less_than_fixed_point}
\end{equation}
Now let $\xi\!=\!f(\nu)$. Then from eq.~(\ref{eq:f_implicit}) and eq.~(\ref{eq:nu_less_than_fixed_point}), it follows that
\begin{align}
0 &= \nu \cdot 0 \\[1ex]
   &= \nu\left[\lambda\left(\nu + \frac{\varepsilon^2}{1\!+\!\lambda}\right) \xi^2  + \xi - (1\!+\!\lambda)\nu\right], \\
  &= \lambda\nu\left(\nu + \frac{\varepsilon^2}{1\!+\!\lambda}\right) \xi^2\, +\, \nu\xi\, -\, (1\!+\!\lambda)\nu^2, \\[1ex]
  &<  \lambda\xi^2\, +\, \nu\xi\, -\, (1\!+\!\lambda)\nu^2, \\[2ex]
  &= (\xi-\nu)(\xi + (1\!+\!\lambda)\nu). \label{eq:factor-ineq}
\end{align}
Since the right factor in eq.~(\ref{eq:factor-ineq}) is positive, the inequality as a whole can only be satisfied if $\xi\!>\!\nu$, or equivalently if $f(\nu)\!>\nu$, thus proving property~(v). To prove property (vi), we observe from eq.~(\ref{eq:fixed_point_quad}) that $\nu^*\!\leq\!1$, and from this \textit{upper} bound on $\nu^*$, we re-use eq.~(\ref{eq:fixed_point_quad}) to derive the \textit{lower} bound
\begin{equation}
\nu^* = \frac{1}{\nu^* + \frac{\varepsilon^2}{1+\lambda}} \geq
   \frac{1}{1 + \frac{\varepsilon^2}{1+\lambda}} =  \frac{1\!+\!\lambda}{1\!+\!\lambda\!+\!\varepsilon^2}.
 \label{eq:fixed_point_lower_bound}
\end{equation}
With this lower bound, we show next that property (vi) follows from properties (iv) and (v). In particular, if $\nu\in(0,\nu^*)$, then from property (v) we have $f(\nu)\!>\!\nu$; on the other hand, if $\nu\!\geq\!\nu^*$, then from property (iv) and the lower bound in eq.~(\ref{eq:fixed_point_lower_bound}), we have $f(\nu)\!\geq\!\nu^*\!\geq\!\frac{1+\lambda}{1+\lambda\!+\!\varepsilon^2}$. But either $\nu\in(0,\nu^*)$ or $\nu\!\geq\!\nu^*$, and hence for all $\nu\!>\!0$ we have
\begin{equation}
f(\nu) \geq \min\left(\nu, \frac{1\!+\!\lambda}{1\! +\! \lambda\! +\! \varepsilon^2}\right),
\end{equation}
which is exactly property (vi). Fig.~(\ref{fig:plot_f}) plots the lower and upper bounds on $f$ from properties (ii) and (vi), as well as the fixed point $\nu^*\! =\! f(\nu^*)$. Property (vii) considers the special case when $\varepsilon^2\!=\!0$. In this case, we can also rewrite eq.~(\ref{eq:f_implicit}) as
\begin{equation}
\nu-1\ =\ \lambda\nu \xi^2 + \xi - \lambda\nu - 1\ =\
  (1 +\lambda\nu+\lambda\nu \xi)(\xi - 1),
\end{equation}
and taking the absolute values of both sides, we find that
\begin{equation}
|\nu-1|\ =\  (1 +\lambda\nu+\lambda\nu \xi)|\xi - 1|\ \geq (1+\lambda\nu)|\xi-1|
\end{equation}
for all $\nu>0$, thus proving property (vii). The meaning of this property becomes more evident upon examining the function's fixed point: note from eq.~(\ref{eq:fixed_point_quad}) that $\nu^*\!=\!1$ when $\varepsilon^2\!=\!0$. Thus property (vii) can alternatively be written as
\begin{equation}
|f(\nu)-\nu^*| \leq \frac{1}{1+\lambda\nu}|\nu-\nu^*|,
\end{equation}
showing that the function converges to its fixed point when it is applied in an iterative fashion.
\end{proof}


\begin{flemma}
\label{lemma-g-mono}
Let $\lambda,\nu\!>\!0$, and let $g:[0,\infty)\!\rightarrow\!\mathbb{R}_+$ be the function defined implicitly as follows: if $\xi\!=\!g(\varepsilon^2)$, then $\xi$ is equal to the \textit{positive} root of the quadratic equation
\begin{equation}
\lambda\left(\nu + \frac{\varepsilon^2}{1\!+\!\lambda}\right) \xi^2  + \xi - (1\!+\!\lambda)\nu\, =\, 0.
\label{eq:g_implicit}
\end{equation}
Then $g$ has the following properties:
\begin{itemize}[noitemsep]
\item[(i)] $g$ is monotonically decreasing on $[0,\infty)$.
\item[(ii)] $g(0) < \sqrt{\frac{1+\lambda}{\lambda}}$.
\item[(iii)] $g'(0) > -1$.
\item[(iv)] $g$ is convex on $[0,\infty)$.
\item[(v)] $|g(\varepsilon^2)\!-\!g(0)| \leq \varepsilon^2$.
\end{itemize}
\end{flemma}

\begin{figure}[t]
\centerline{\includegraphics[width=0.66\textwidth]{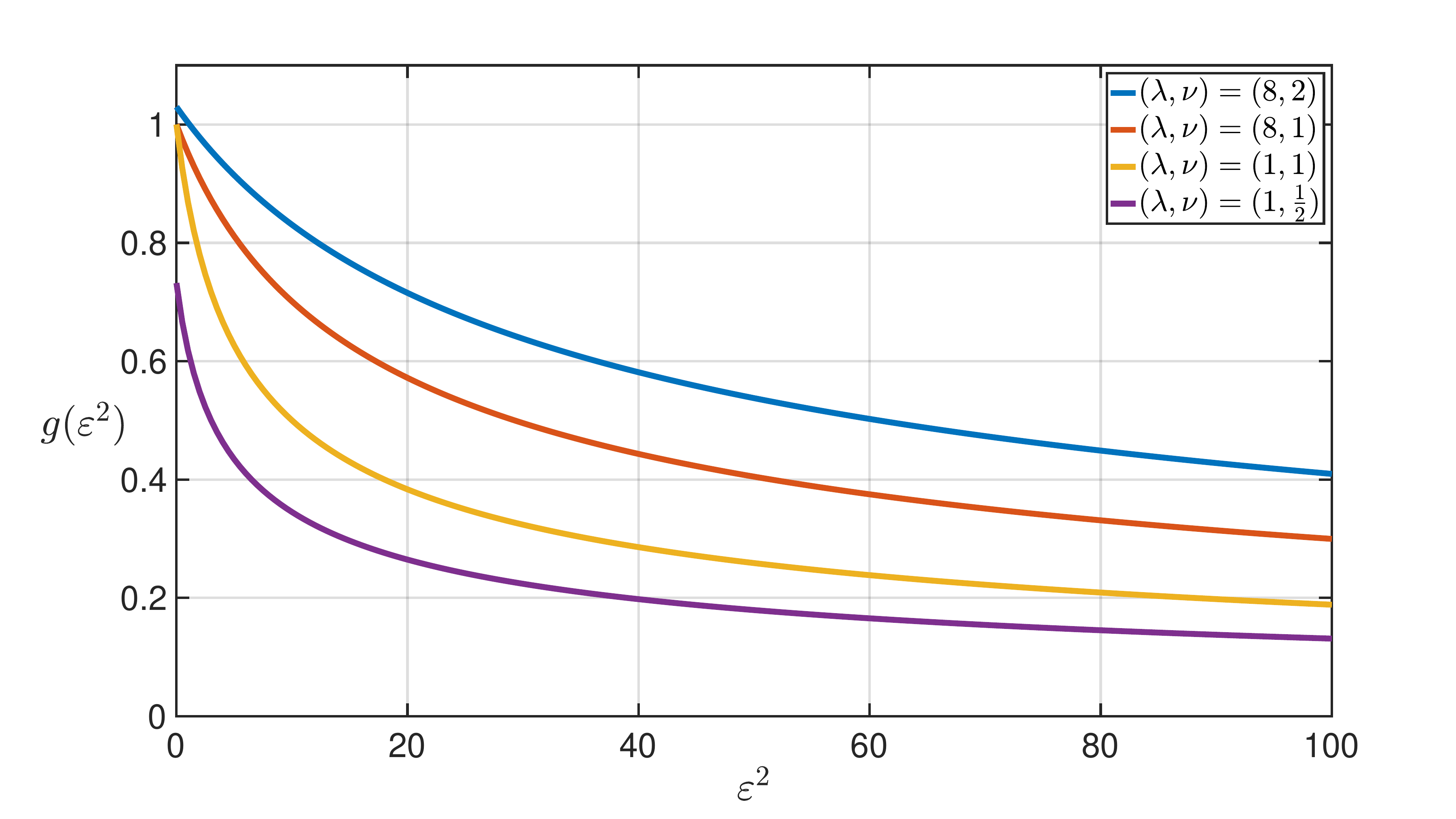}}
\caption{Plot of the function $g$ in \Cref{lemma-g-mono} and eq.~(\ref{eq:g_eps}) for several different values of $\lambda$ and $\nu$.}
\label{fig:plot_g}
\end{figure}

\noindent
Before proving the lemma, we note that it is straightforward to solve the quadratic equation in eq.~(\ref{eq:g_implicit}). Doing so, we find
\begin{equation}
g(\varepsilon^2) = \frac{-1+\sqrt{1+4\lambda(1+\lambda)\nu^2 + 4\lambda\varepsilon^2\nu}}
  {2\lambda\left(\nu + \frac{\varepsilon^2}{1+\lambda}\right)}.
\label{eq:g_eps}
\end{equation}
This explicit formula for $g$ is not needed for the proof of the lemma. However, eq.~(\ref{eq:g_eps}) is useful for visualizing properties (i)-(ii), and Fig.~\ref{fig:plot_g} shows several plots of $g(\varepsilon^2)$ for different values of $\lambda$ and $\nu$. We now prove the lemma.

\begin{proof}
To prove property (i) that $g$ is monotonically increasing, it suffices to show $g'(\varepsilon^2)\!<\!0$. Differentiating eq.~(\ref{eq:g_implicit}) with respect to $\varepsilon^2$, we find that
\begin{equation}
\frac{\lambda}{1\!+\!\lambda} \xi^2 + 2\lambda\left(\nu + \frac{\varepsilon^2}{1\!+\!\lambda}\right) \xi g'(\varepsilon^2) + g'(\varepsilon^2) = 0
\label{eq:dg}
\end{equation}
where $\xi\!=\!g(\varepsilon^2)$, and solving for $g'(\varepsilon)$, we find that
\begin{equation}
g'(\varepsilon^2) = - \frac{\lambda\xi^2}{(1\!+\!\lambda)(1\! +\! 2\lambda\nu\xi) + 2\lambda\varepsilon^2\xi} < 0,
\label{eq:dg_neg}
\end{equation}
which proves property (i). To prove property (ii), let $\xi_0\!=\!g(0)$ denote the positive root of eq.~(\ref{eq:g_implicit}) when $\varepsilon^2\!=\!0$. Then this root satisfies
\begin{equation}
\xi_0^2 = \frac{1\!+\!\lambda}{\lambda} - \frac{\xi_0}{\lambda\nu} < \frac{1\!+\!\lambda}{\lambda},
\label{eq:xi0}
\end{equation}
from which the result follows. Moreover, it follows from eqs.~(\ref{eq:dg_neg}--\ref{eq:xi0}) that
\begin{equation}
g'(0) = -\frac{\lambda \xi_0^2}{(1\!+\!\lambda)(1+2\lambda\nu \xi_0)} >
   -\frac{\lambda \xi_0^2}{1\!+\!\lambda} > -\frac{\lambda}{1\!+\!\lambda}\frac{1\!+\!\lambda}{\lambda} = -1,
 \label{eq:dg_bound}
\end{equation}
thus proving property (iii).
To prove property (iv) that $g$ is convex, it suffices to show $g''(\varepsilon^2)>0$. Differentiating eq.~(\ref{eq:dg}) with respect to $\varepsilon^2$, we find that
\begin{equation}
\frac{4\lambda\xi}{1\!+\!\lambda}\, g'(\varepsilon^2) + 2\lambda\left(\nu + \frac{\varepsilon^2}{1\!+\!\lambda}\right) \Big(\xi g''(\varepsilon^2) + g'(\varepsilon^2)^2\Big) + g''(\varepsilon^2) = 0.
\end{equation}
To proceed, we re-arrange terms to isolate $g''(\varepsilon^2)$ on the left side and use eq.~(\ref{eq:dg}) to re-express the term on the right. In this way, we find:
\begin{align}
\left[1 + 2\lambda\left(\nu + \frac{\varepsilon^2}{1\!+\!\lambda}\right)\xi\right]g''(\varepsilon^2)
  &= -\frac{4\lambda\xi}{1\!+\!\lambda} g'(\varepsilon^2) - 2\lambda\left(\nu + \frac{\varepsilon^2}{1\!+\!\lambda}\right)g'(\varepsilon^2)^2, \\[1.5ex]
  &= -\frac{g'(\varepsilon^2)}{\xi}
    \left[\frac{4\lambda\xi^2}{1\!+\!\lambda} + 2\lambda\left(\nu + \frac{\varepsilon^2}{1\!+\!\lambda}\right)\xi g'(\varepsilon^2)\right], \\[1.5ex]
  &= -\frac{g'(\varepsilon^2)}{\xi}
    \left[\frac{4\lambda\xi^2}{1\!+\!\lambda}  - \frac{\lambda\xi^2}{1\!+\!\lambda} -  g'(\varepsilon^2)\right], \\[1.5ex]
  &= -\frac{g'(\varepsilon^2)}{\xi}
    \left[\frac{3\lambda\xi^2}{1\!+\!\lambda}  -  g'(\varepsilon^2)\right].
\end{align}
Note that the term in brackets on the left side is strictly positive, and because $g$ is monotonically decreasing, with $g'(\varepsilon^2)<0$, so is the term on the right. It follows that $g''(\varepsilon^2)>0$, thus proving property (iv). Finally, to prove property (v), we combine the results that $g$ is monotonically decreasing, that its derivative at zero is greater than -1, and that it is convex:
\begin{equation}
|g(\varepsilon^2)-g(0)|\,
  =\, g(0) - g(\varepsilon^2)\,
  \leq\, g(0) - (g(0)+g'(0)\varepsilon^2)
  = -g'(0)\varepsilon^2\,
  \leq\, \varepsilon^2.
\end{equation}
\end{proof}





\section{Additional experiments and details}
\label{appendix-sec-experiments}

\subsection{Implementation of baselines}
\label{appendix-sec-baselines}

In \Cref{alg:advi}, we describe the version of ADVI implemented in the experiments.
In particular, we use ADAM as the optimizer for updating the variational parameters.
We also implemented an alternate version of ADVI using the score-based divergence and the Fisher
divergence in place of the (negative) ELBO loss.
In \Cref{alg:gsm}, we also describe the implementation of the GSM algorithm~\citep{Modi2023}.

\begin{algorithm*}[t]
\small
\caption{Implementation of ADVI}
\label{alg:advi}
\begin{algorithmic}[1]
\STATE \textbf{Input:}
Iterations $T$, batch size $B$,
unnormalized target $\tilde p$,
learning rate $\lambda_t > 0$,
initial variational mean $\mu_0 \in \reals^D$,
initial variational covariance $\Sigma_0 \in \SS_{++}^{D}$

\FOR{$t = 0,\ldots,T-1$}
    \STATE Sample $z_1,\ldots,z_B \sim q_t = \N(\mu_t, \Sigma_t)$

\STATE Compute stochastic estimate of the (negative) ELBO
\vspace{-10pt}
\begin{align*}
\mathcal{L}^{(t)}_{\text{ELBO}}(z_{1:B}) = -\sum_{b=1}^B \log(\tilde p(z_b) - \log q_t(z_b))
\end{align*}
\vspace{-10pt}

    \STATE Update variational parameters $w_t := (\mu_t,\Sigma_t)$ with gradient
\vspace{-5pt}
    \[
    w_{t+1} = w_t - \lambda_t \nabla_w \mathcal{L}^{(t)}_{\text{ELBO}}(z_{1:B})
    \hspace{25px}
\COMMENT{\emph{Our implementation uses the ADAM update.}}
        \]
\vspace{-15pt}
\ENDFOR
\STATE \textbf{Output:} variational parameters $\mu_T, \Sigma_T$
\end{algorithmic}
\end{algorithm*}

\begin{algorithm*}[t]
\small
\caption{Implementation of GSM}
\label{alg:gsm}
\begin{algorithmic}[1]
\STATE \textbf{Input:}
 Iterations $T$, batch size $B$,
 unnormalized target $\tilde p$,
    initial variational mean $\mu_0 \in \reals^D$,
    initial variational covariance $\Sigma_0 \in \SS_{++}^D$

\FOR{$t = 0,\ldots,T-1$}
    \STATE Sample $z_1,\ldots,z_B \sim q_t = \N(\mu_t, \Sigma_t)$

    \FOR{$b = 1,\ldots,B$}

        \STATE Compute the score of the sample
        $s_b = \nabla_z \log (\tilde p(z_b))$

        \STATE Calculate intermediate quantities
    \vspace{-5pt}
        \begin{align*}
        \epsilon_b =  \Sigma_t s_b - \mu_t + z_b, \qquad
\text{and~}
        \text{solve}\,\, \rho(1\!+\!\rho) =  s_b^\top \Sigma_t s_b   +
            \big[(\mu_t\!-z_{b})^\top\!s_b\big]^2\,  \mathrm{for}\, \rho>0
        \end{align*}
    \vspace{-10pt}

        \STATE Estimate the update for mean and covariance
        \vspace{-5pt}
        \begin{align*}
            \delta\mu_b &= \tfrac{1}{1+\rho}\left[ I - \tfrac{(\mu_t-z_b) s_b^\top}{1+\rho+
            (\mu_t-z_b)^\top s_b}\right] \epsilon_b
        \\
            \delta\Sigma_b &= ({\mu}_t-z_b)({\mu}_t-z_b)^\top -
            (\tilde\mu_b-
            z_b)(\tilde\mu_b-z_b)^\top,
            \qquad
            \text{where~}
            \tilde{\mu}_b  = \mu_t + \delta\mu_b
        \end{align*}
        \vspace{-15pt}

        \ENDFOR

    \STATE Update variational mean and covariance
    \vspace{-10pt}
    \begin{align*}
        \mu_{t+1} &= \mu_t + \tfrac{1}{B} \sum_{b=1}^B \delta \mu_b, \qquad
        \Sigma_{t+1} = \Sigma_t + \tfrac{1}{B} \sum_{b=1}^B \delta \Sigma_b
    \end{align*}
    \vspace{-10pt}

    \ENDFOR
    \STATE \textbf{Output:} variational parameters $\mu_T, \Sigma_T$
\end{algorithmic}
\end{algorithm*}

\subsection{Wallclock timings}
\label{ssec-wallclock}

In the main paper, we report the number of gradient evaluations as a measure of the cost of the
algorithm.
While the complete cost is not captured by  the number of gradient evaluations alone,
here we show that the computational cost of the algorithms are dominated by
gradient evaluations, and so number of gradient evaluations is a good proxy of the computational cost.
We additionally note that all work with full covariance matrices make a basic assumption that $\O(D^2)$
is not prohibitive because there are $\O(D^2)$ parameters in the model itself.
While the BaM update (when $B \geq D$) takes $\O(D^3)$ computation per iteration,
in this setting,
$\O(D^3)$ is not generally regarded as prohibitive in models where there are $\O(D^2)$
parameters to estimate.

\begin{figure}
\centering
\includegraphics[scale=0.41]{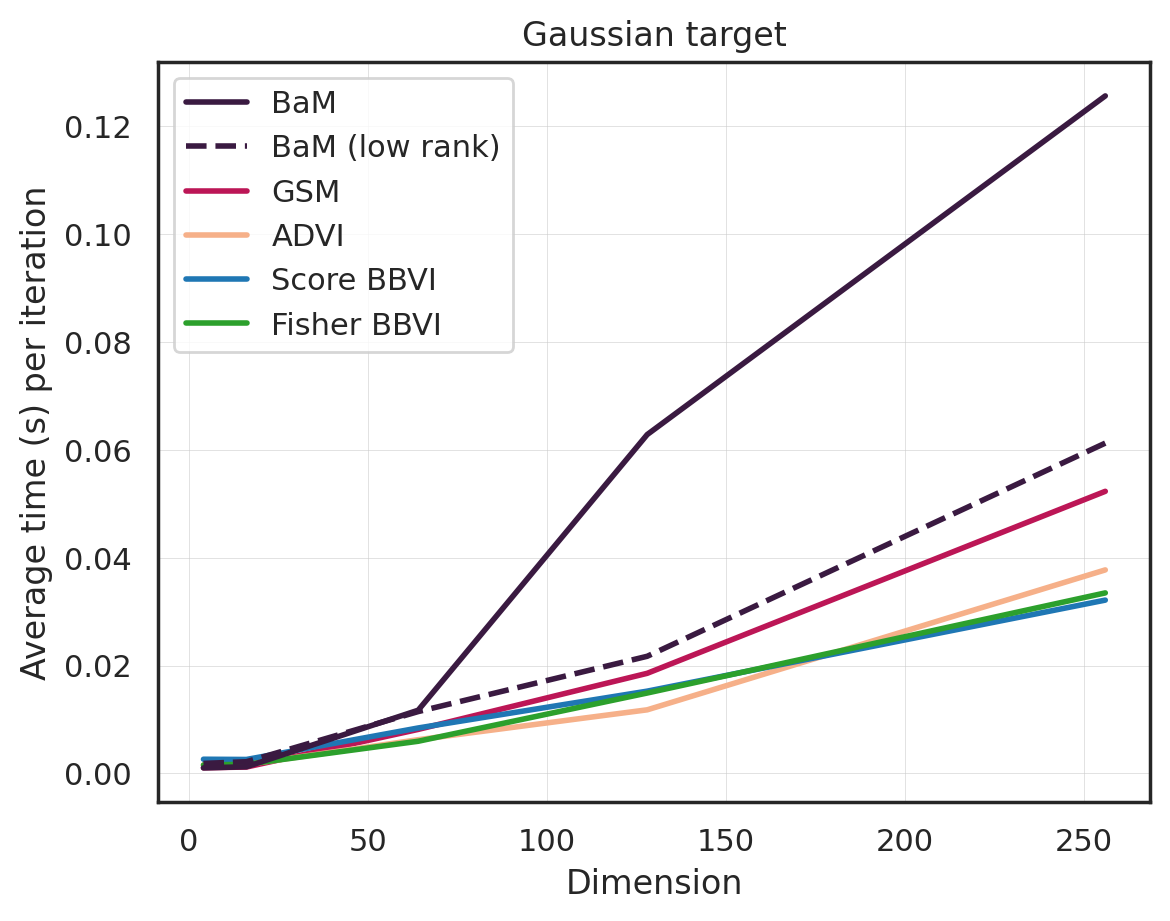}
\caption{Wallclock timings for the Gaussian targets example.}
\label{fig:timings}
\end{figure}

In \Cref{fig:timings}, we plot the wallclock timings for  Gaussian targets of increasing
dimension, where  $D = 4, 16, 64, 128, 256$.
We observe that for dimensions 64 and under, all methods have similar timings;
for the larger dimensions, we observe that the low-rank BaM solver has a similar timing.
All experiments in the paper fit into the lower-dimensional regime or the low-rank regime,
with the exception of the deep generative models application, which includes
larger batch sizes.
Thus, for the lower-dimensional regime and the low-rank examples, we report the number of gradient evaluations
as the primary measure of cost; the cost per iteration for the mini-batch regime is
$\O(D^2 B + B^3)$.
For the deep generative model example, we additionally report in \Cref{fig:image:time}
the wallclock timings.
We note that the wallclock timings themselves are heavily dependent on
implementation and JIT-compilation details and hardware.

\subsection{Gaussian target}
\label{ssec-gaussian-expts}

Each target distribution was generated randomly; here the covariance
was constructed by generating a $D\times D$ matrix $A$ and computing $\Sigma_{*} = AA^\top$.

\begin{figure*}[t]
    \centering
    \includegraphics[scale=0.48]{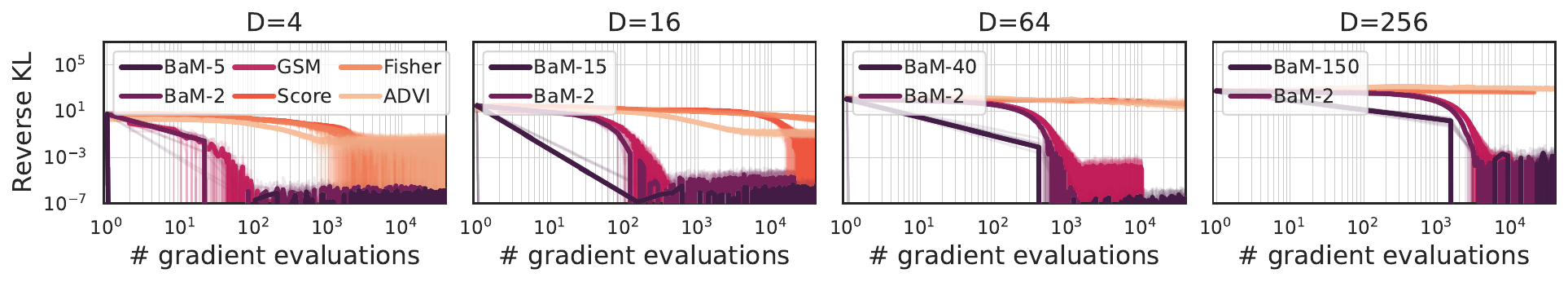}
    \caption{Gaussian targets of increasing dimension.
    Solid curves indicate the mean over 10 runs (transparent curves).
    {ADVI, Score, Fisher,} and GSM use batch size of 2. The batch size for BaM is given in the legend.
}
    \label{fig-gaussian-target-rkl}
\end{figure*}

For all experiments, the algorithms were initialized with  {$\mu_0 \sim \text{uniform}[0,0.1]$}
and $\Sigma_0 = I$.
In \Cref{fig-gaussian-target-rkl}, we report the results for the reverse KL divergence.
We observe largely the same conclusions as with the forward KL divergence presented in \Cref{sec-experiments}.

\begin{figure}[t]
    \centering
    \includegraphics[scale=0.41]{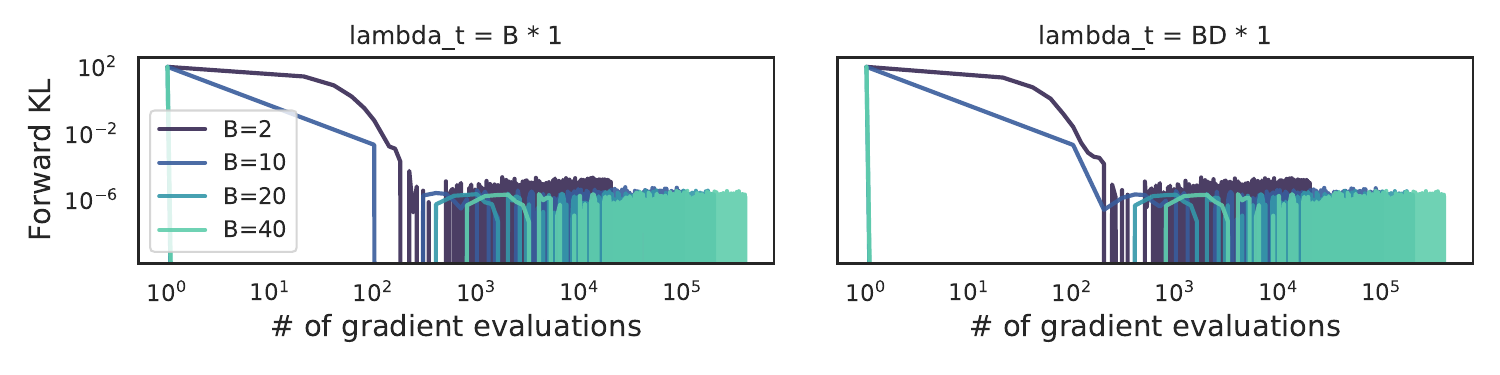}
    \includegraphics[scale=0.41]{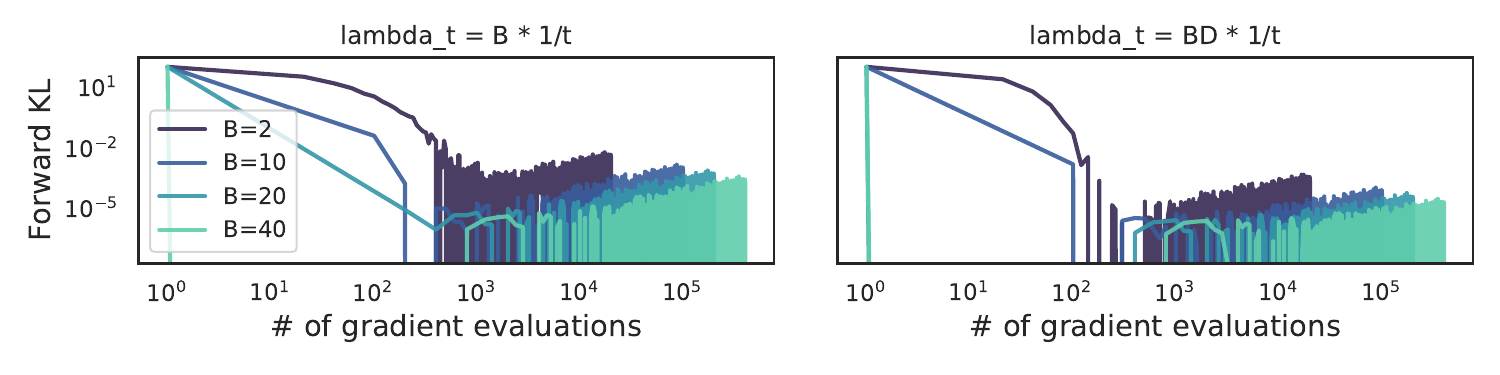}
    \caption{Gaussian target, $D=16$}
    \label{fig-gaussian-lambda}
\end{figure}

In addition,
we evaluated BaM with a number of different schedules for the learning rates:
\mbox{$\lambda_t = B, BD, \frac{B}{t+1}, \frac{BD}{t+1}$}.
We show one such example for $D=16$
in \Cref{fig-gaussian-lambda}, where each figure represents a particular choice of $\lambda_t$,
and where each line is the mean over 10 runs.
For the constant learning rate,
the lines for $B=20,40$ are on top of each other.
Here we observe that the constant learning rates perform the best for Gaussian targets.
For the gradient-based methods (ADVI, Score, Fisher), the learning rate was set  by choosing the
best value over a grid search. For ADVI and Fisher, the selected learning rate was  0.01.
For Score, a different learning rate was selected for each dimension $D=4,16,64,256$:
$[0.01,0.005,0.001,0.001]$.


\subsection{Non-Gaussian target}
\label{ssec-nongaussian-expts}

\begin{figure*}[t]
    \centering
    \includegraphics[scale=0.40]{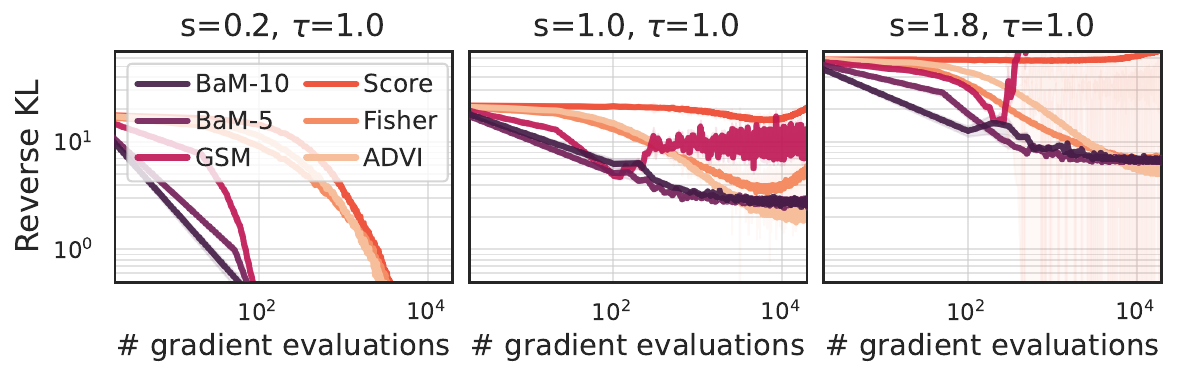}
    \includegraphics[scale=0.40]{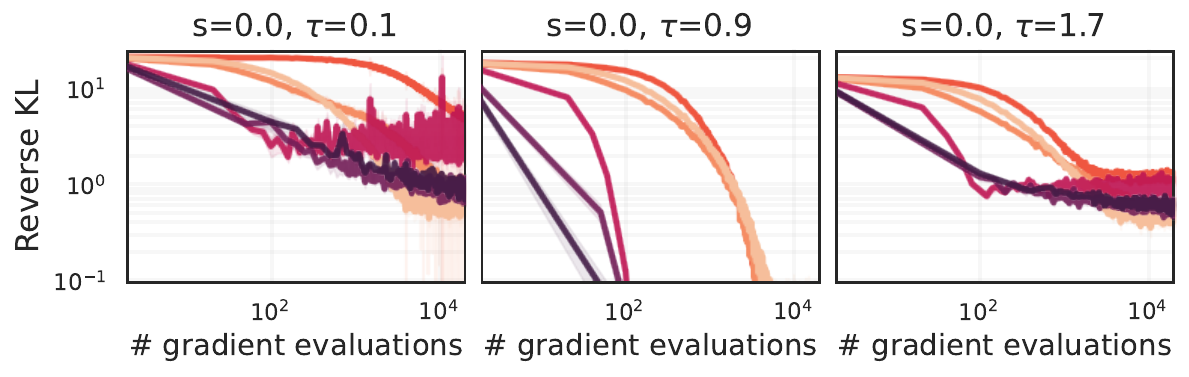}
\caption{Non-Gaussian targets constructed using the sinh-arcsinh
distribution, varying the skew $s$ and the tail weight $t$.
ADVI and GSM use a batch size of $B\!=\! 5$.
    }
\label{fig-nongaussian-target2}
\end{figure*}

Here we again consider the sinh-arcsinh distribution with~\mbox{$D=10$}, where we
vary the skew and tails.
{We present the reverse KL results in \Cref{fig-nongaussian-target2}.}

\begin{figure}[p]
    \centering
    \begin{subfigure}[b]{0.49\linewidth}
    \centering
    \includegraphics[scale=0.31]{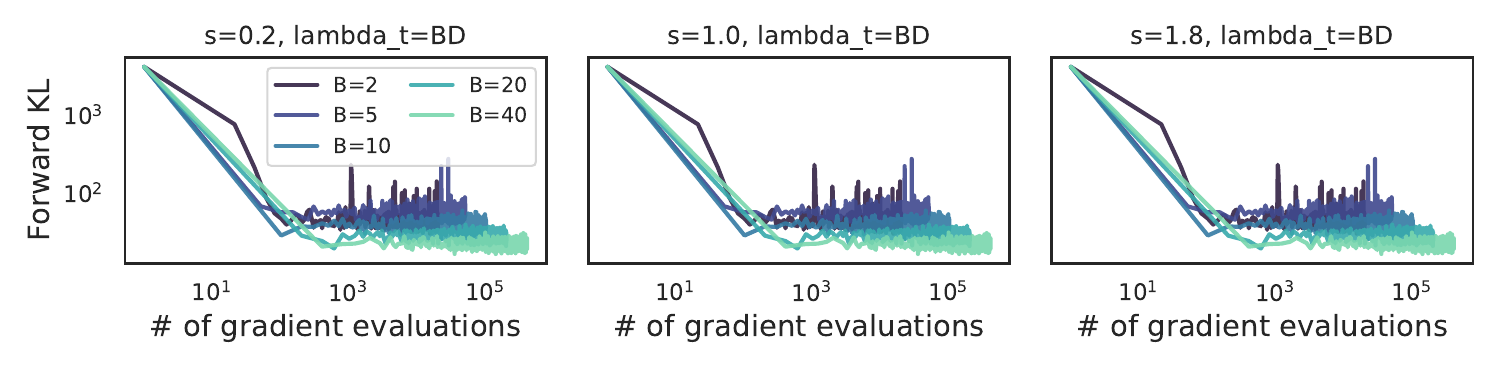}
    \includegraphics[scale=0.31]{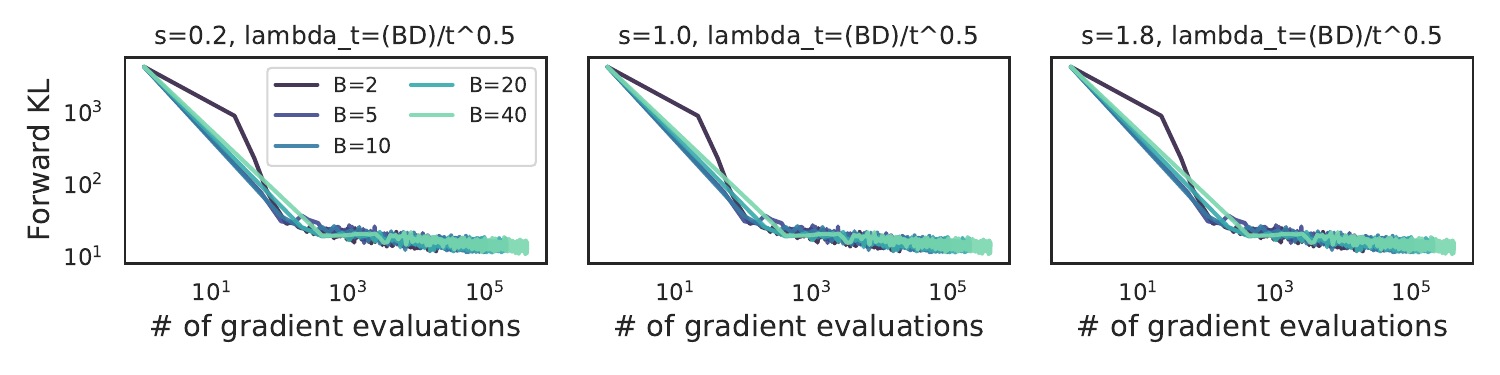}
    \includegraphics[scale=0.31]{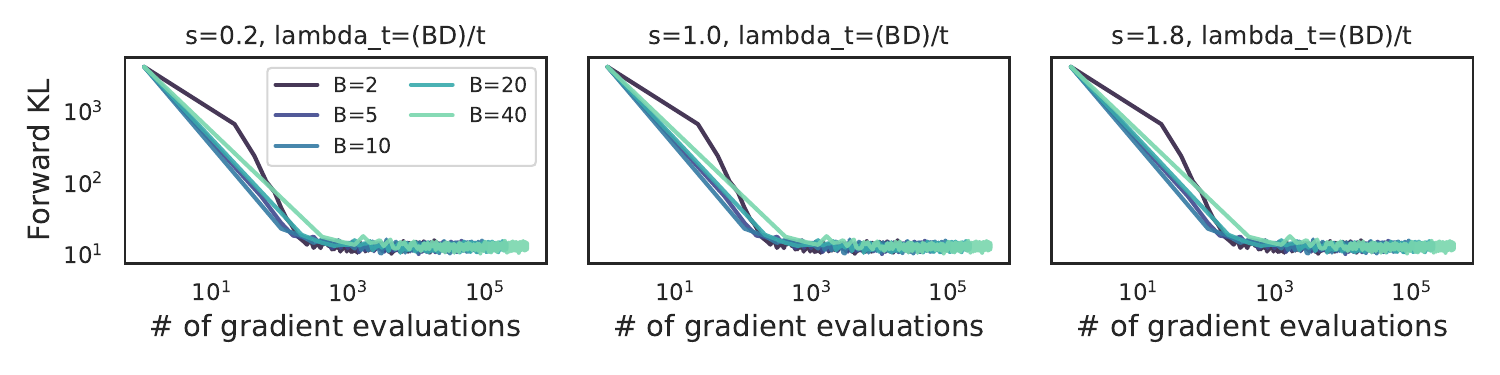}
    \caption{Forward KL: varying $\lambda_t$ and skew}
    \end{subfigure}
    \begin{subfigure}[b]{0.49\linewidth}
    \centering
    \includegraphics[scale=0.31]{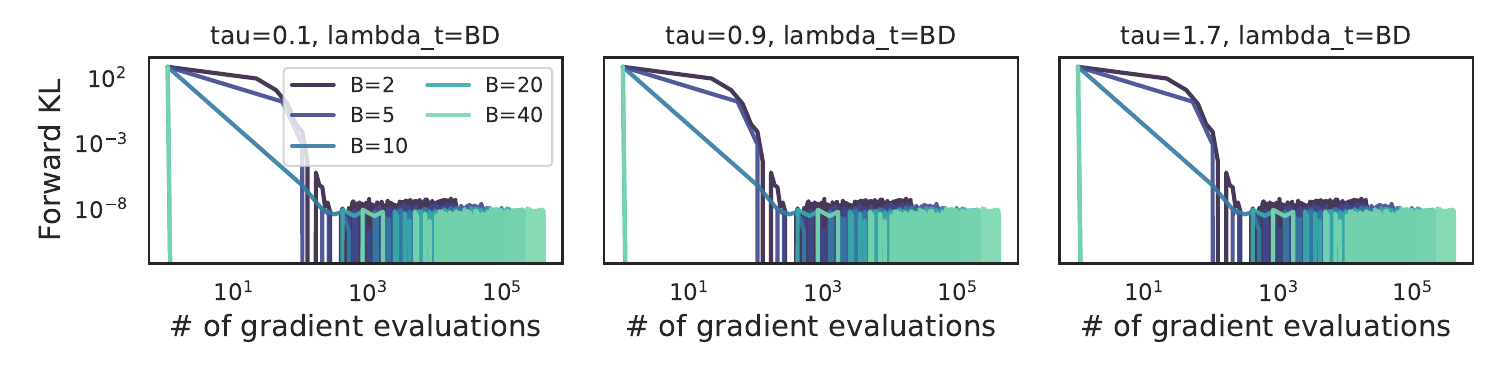}
    \includegraphics[scale=0.31]{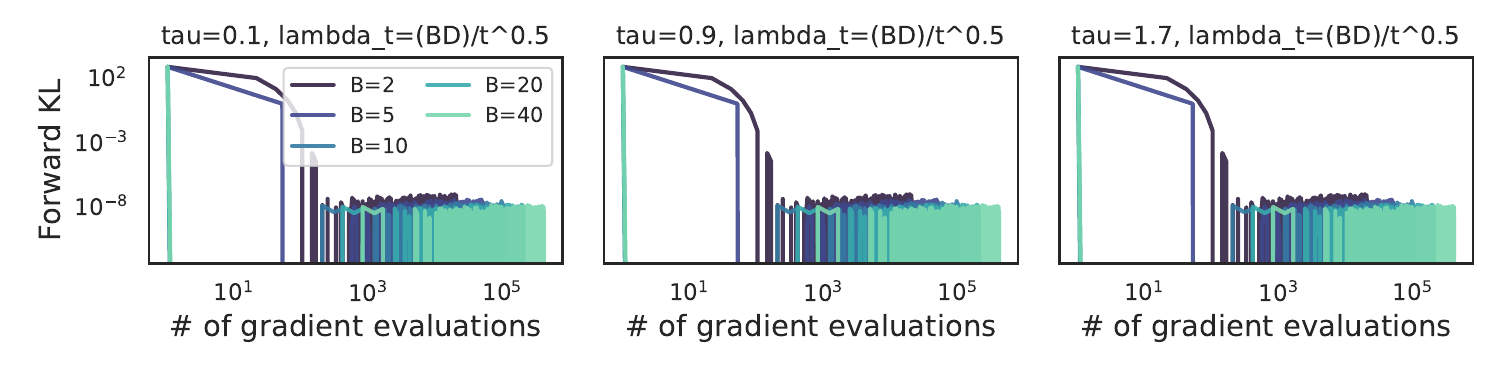}
    \includegraphics[scale=0.31]{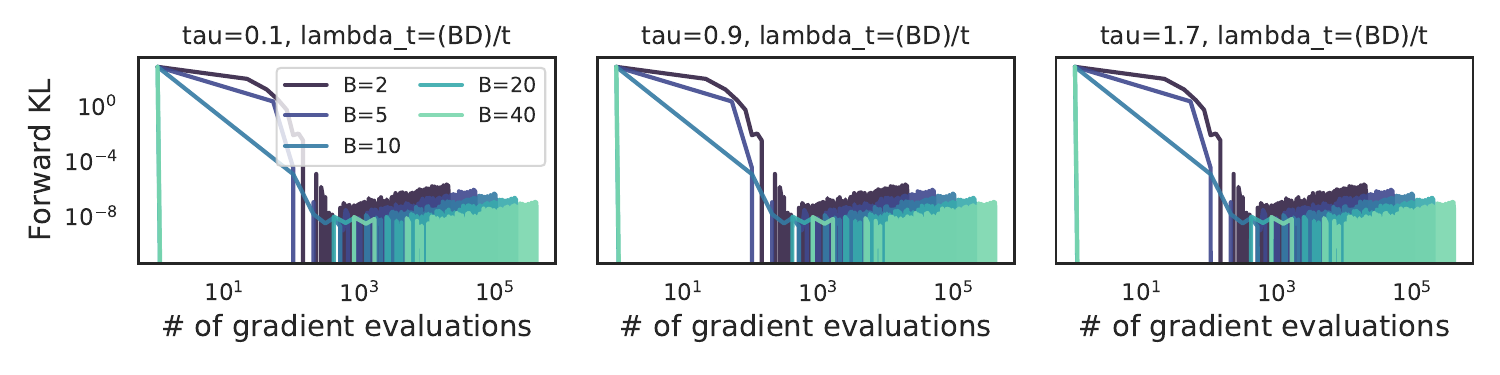}
    \caption{Forward KL: varying $\lambda_t$ and tails}
    \end{subfigure}
    \begin{subfigure}[b]{0.49\linewidth}
    \centering
    \includegraphics[scale=0.31]{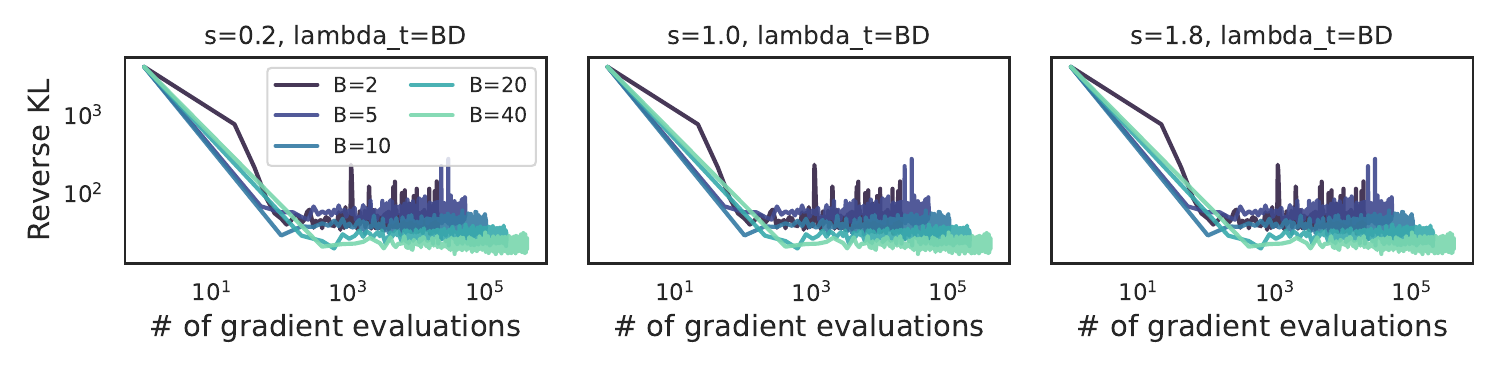}
    \includegraphics[scale=0.31]{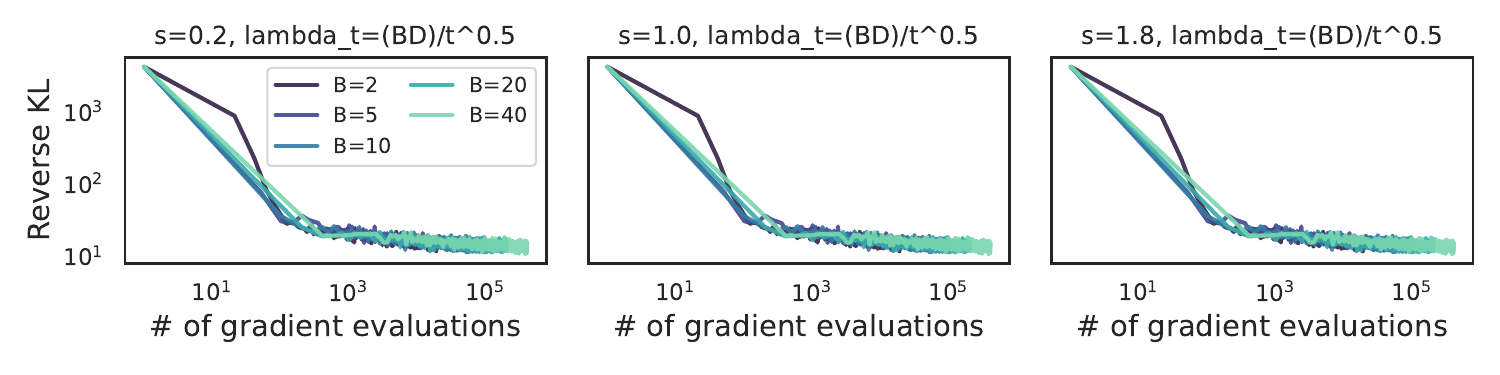}
    \includegraphics[scale=0.31]{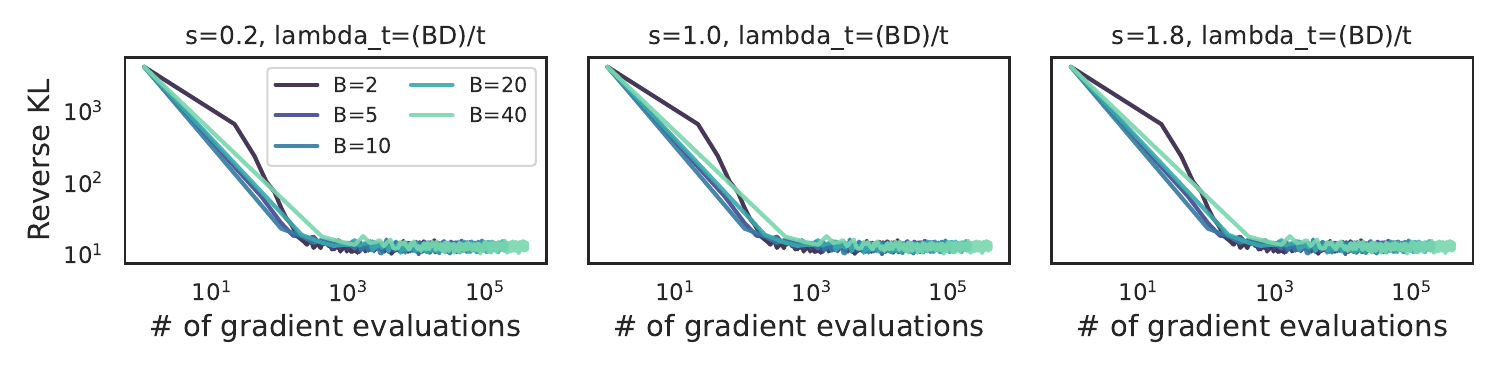}
    \caption{Reverse KL: varying $\lambda_t$ and skew}
    \end{subfigure}
    \begin{subfigure}[b]{0.49\linewidth}
    \centering
    \includegraphics[scale=0.31]{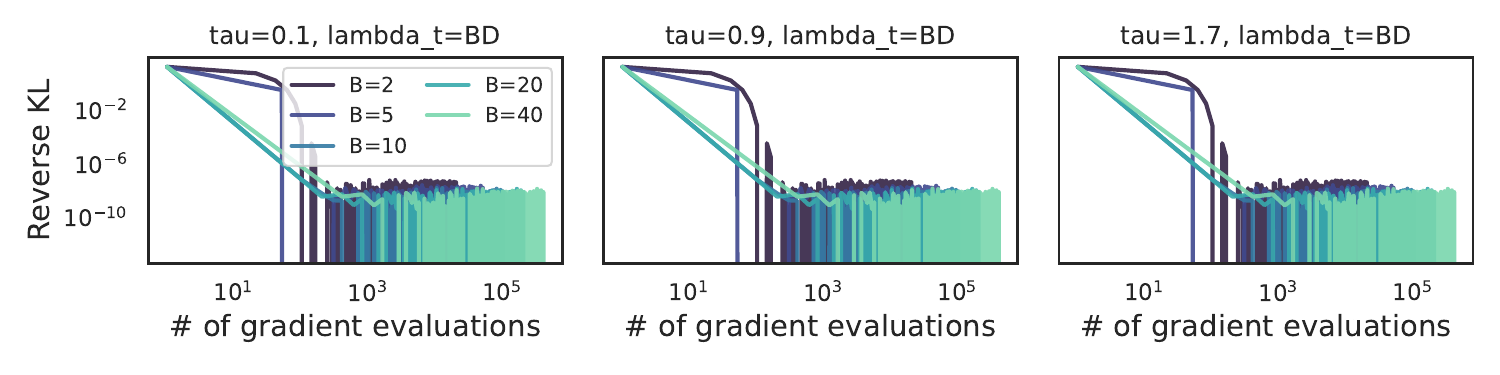}
    \includegraphics[scale=0.31]{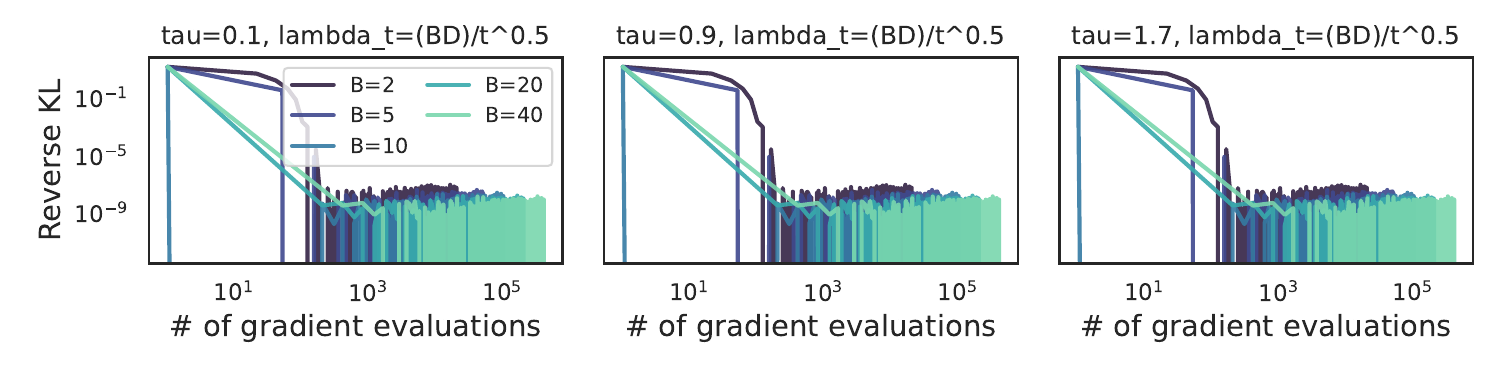}
    \includegraphics[scale=0.31]{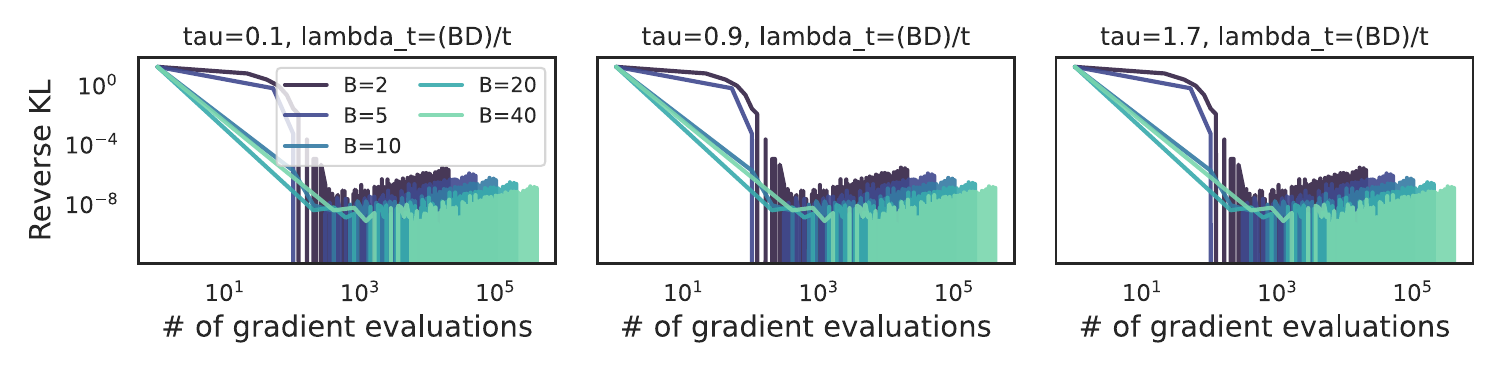}
    \caption{Reverse KL: varying $\lambda_t$ and tails}
    \end{subfigure}
\caption{Non-Gaussian target, $D=10$.
    Panels (a) and (b) show the forward KL, and
    panels (c) and (d) show the reverse KL.
    }
\label{fig-sinh1}
\end{figure}

All algorithms were initialized with a random initial mean $\mu_0$ and $\Sigma_0 = I$.
In \Cref{fig-sinh1},
we present several alternative plots showing the forward
and reverse KL divergence when varying the learning rate.
We investigate the performance for different schedules corresponding to
$\lambda_t = BD, \frac{BD}{\sqrt{t+1}}, \frac{BD}{(t+1)}$,
and we varied the batch size $B=2,5,10,20,40$.
Unlike for Gaussian targets, we found that constant $\lambda_t$ did not perform as well
as those with a varying schedule.
In particular, we found that $\lambda_t = \frac{BD}{t+1}$ typically converges
faster than the other schedule.

For the gradient-based methods (ADVI, Score, Fisher), a grid search was run over the
learning rate for ADAM. The final selected learning rates were 0.02 for ADVI
 and 0.05 for Fisher.
For Score, more tuning was required:
for the targets with fixed tails $\tau=1$ and varying skew $s=0.2,1,1.8$, the learning rates
$[0.01, 0.001, 0.001]$ were used,
and for the targets with fixed skew $s=0$ and varying tails $\tau=0.1,0.9,1.7$,
the learning rates
$[0.001, 0.01, 0.01]$, respectively.
We note that for the score-based divergence,
several of the highly skewed targets led to divergence (with respect to the grid search)
on most of the random seeds that were run.

\subsection{Posteriordb models}
\label{sec-posterior-db-details}

In Bayesian posterior inference applications,
it is common to measure the relative mean error and the relative
standard deviation error \citep{welandawe2022robust}:
\begin{align}
\text{relative~mean~error} =\norm{\frac{\mu-\hat{\mu}}{\sigma}}_2, \quad
\text{relative~SD~error} = \norm{\frac{\sigma-\hat{\sigma}}{\sigma}}_2,
\end{align}
where $\hat{\mu},\hat{\sigma}$ are computed from the variational distribution,
and $\mu,\sigma$ are the posterior mean and standard deviation.
We estimated the posterior mean and standard deviation
using  the reference samples from HMC.

\begin{figure*}[t]
    \centering
    \includegraphics[scale=0.43]{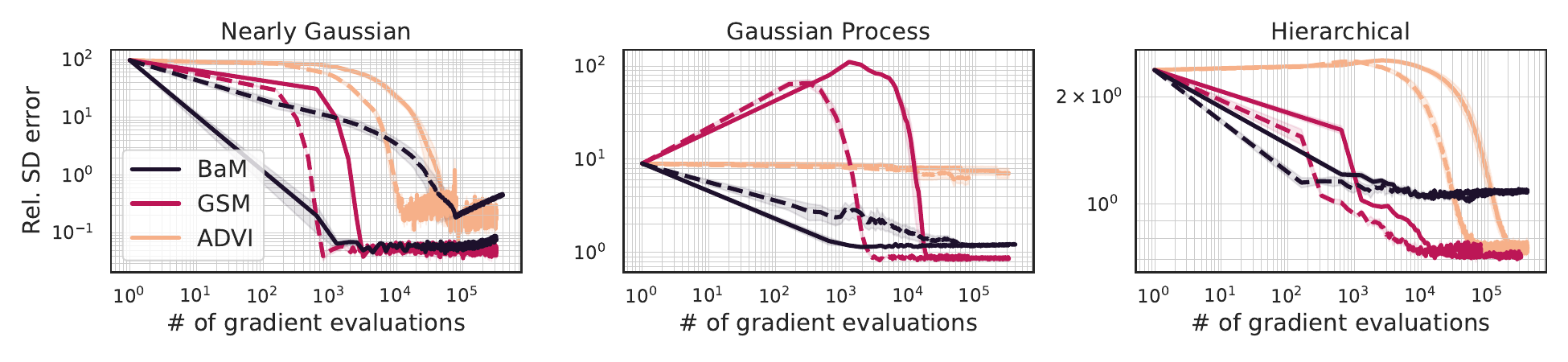}
\caption{Posterior inference in Bayesian models {measured by the relative standard deviation
    error.}
    The curves denote the mean over
    5 runs, and shaded regions denote their standard error.
    Solid curves ($B=32$)
    correspond to larger batch sizes than the dashed curves ($B=8$).
}
\label{fig-pdb-target2}
\end{figure*}

In the evaluation, all algorithms were initialized with
$\mu_0 \sim \text{uniform}[0,0.1]$ and $\Sigma_0 = I$.
The results for the relative mean error are presented in \Cref{sec-experiments}.
In \Cref{fig-pdb-target2}, we present the results for the relative SD error.
Here we typically observe the same trends as for the mean, except in the hierarchical example,
in which BaM learns the mean quickly but converges to a larger relative SD error.
However, the low error of GSM suggests that more robust tuning of the learning rate
may lead to better performance with BaM.

\subsection{Deep learning model}
\label{ssec-deep-learning}

\begin{figure*}[t]
    \centering
      \includegraphics[scale=0.53]{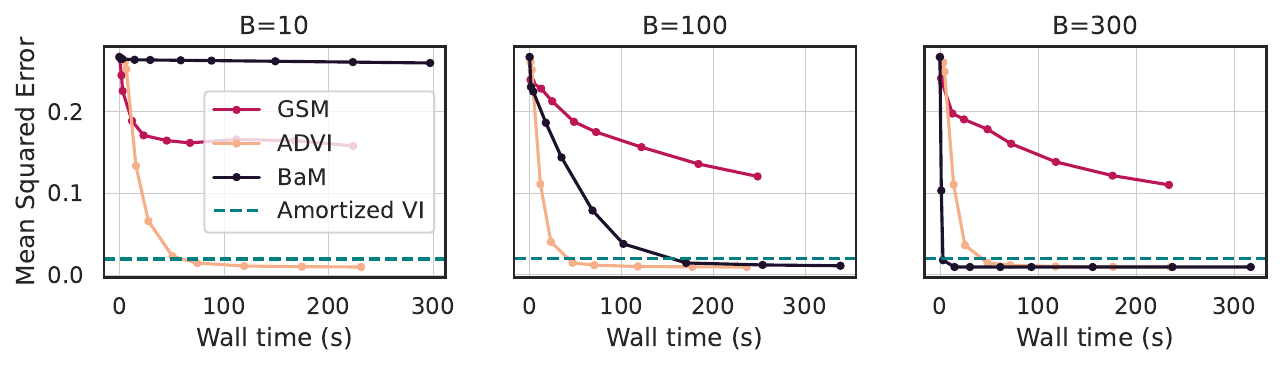}
    \caption{
    Image reconstruction error when the posterior mean of $z'$ is fed into the generative neural network.
    The $x$-axis denotes the wallclock time in seconds.
    }
  \label{fig:image:time}
\end{figure*}

In \Cref{fig:image:time}, we present the results from the main paper but with
wallclock times on the $x$-axis. We arrive at similar conclusions:
here BaM with $B=300$ converges the fastest compared to GSM and ADVI using any batch size.

\begin{figure}
    \centering
  \includegraphics[scale=0.40]{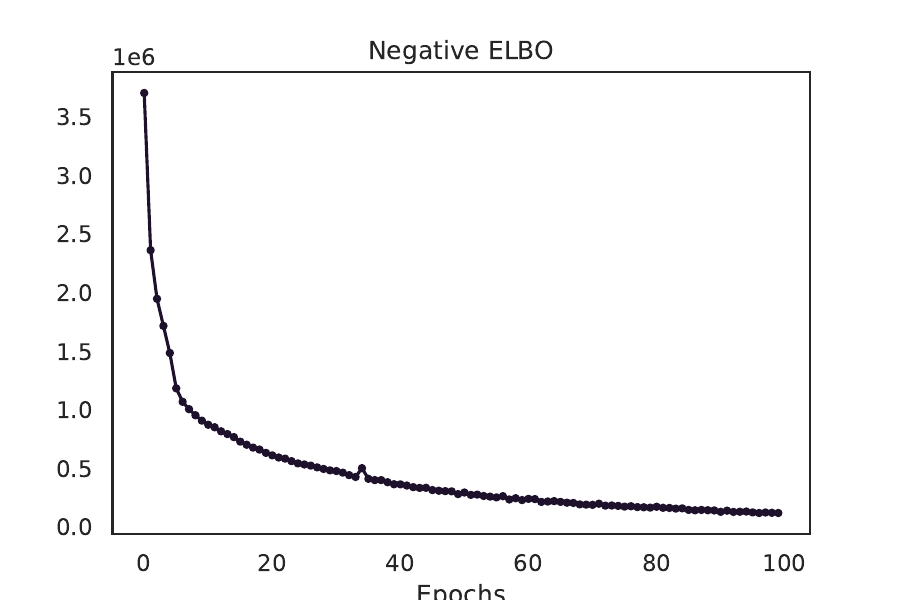}
  \caption{ELBO for variational autoencoder over 100 epochs}
  \label{fig-ELBO-VAE}
\end{figure}

We provide additional details for the experiment conducted in \Cref{sec-deep-learning}.
We first pre-train the neural network $\Omega(\cdot, \hat \theta)$ (the ``decoder'') using variational expectation-maximization.
That is, $\hat \theta$ maximizes the marginal likelihood $p(\{x_n\}_{n=1}^N \given \theta)$,
where $\{x_n\}_{n=1}^N$ denotes the training set.
The marginalization step is performed using an approximation
\[q(z_n | x_n) \approx p(z_n | x_n, \theta),\] obtained with amortized variational inference.
In details, we optimize the ELBO over the family of factorized Gaussians and
learn an inference neural network (the ``encoder'') that maps $x_n$ to the
parameters of $q(z_n \given x_n)$.
This procedure is standard for training a VAE
\citep{kingma2013auto,rezende2014stochastic,tomczak2022deep}.
For the decoder and the encoder, we use a convolution network with 5 layers.
The optimization is performed over 100 epochs, after which the ELBO converges
(\Cref{fig-ELBO-VAE}).

For the estimation of the posterior on a new observation, we draw an image $x'$
from the test set.
All VI algorithms are initialized at a standard Gaussian.
For ADVI and BaM, we conduct a pilot experiment of 100 iterations and select the
learning rate that achieves the lowest MSE for each batch size ($B=10, 100, 300$).
For ADVI, we consistently find the best learning rate to be $\ell = 0.02$
(after searching $\ell = 0.001, 0.01, 0.02, 0.05$).
For BaM, we find that different learning rates work better for different batch sizes:
\begin{itemize}[topsep=0pt,itemsep=0pt]
  \item $B=10$, $\lambda = 0.1$ selected from $\lambda = 0.01, 0.1, 0.2, 10$.
  \item $B=100$, $\lambda = 50$ selected from $\lambda = 2, 20, 50, 100, 200$.
  \item $B=300$, $\lambda = 7500$ selected from $\lambda = 1000, 5000, 7500, 10000$.
\end{itemize}
For $B=300$, all candidate learning rates achieve the minimal MSE
(since BaM converges in less than 100 iterations), and so we pick the one that
yields the fastest convergence.


\end{document}